\documentclass[preprint,review,12pt]{elsarticle}

\usepackage[english]{babel}
\usepackage{lineno}
\usepackage{amsmath,amsfonts,amssymb,amsthm,epsfig,url,epstopdf,array}
\usepackage{graphicx}
\usepackage[justification=centering]{caption}
\usepackage{multirow}
\usepackage{hyperref}
\usepackage{enumerate}
\usepackage{color}
\usepackage[pdftex,dvipsnames]{xcolor}
\usepackage{float}
\usepackage{subcaption}
\usepackage[flushleft]{threeparttable}
\usepackage{booktabs,caption}
\usepackage{xargs}
\usepackage{pgf}
\usepackage{tikz}
\usepackage{relsize}
\usetikzlibrary{arrows,automata,positioning}
\usepackage[flushleft]{threeparttable} 
\usepackage{algorithm}
\usepackage{algpseudocode}
\usepackage{scrextend}
\usepackage{lscape}
\usepackage{listings}
\usepackage{changepage}
\usepackage[many]{tcolorbox}
\usepackage{verbatim}

\hypersetup{pdfauthor={Name}}

\newcolumntype{L}{>{\arraybackslash}m{4cm}}

\newlength{\bubblesep}
\newlength{\bubblewidth}
\AtBeginDocument{\setlength{\bubblewidth}{.75\textwidth}}
\definecolor{bubblegray}{RGB}{220,220,220}
\definecolor{bubblepurple}{RGB}{104, 135, 190}

\newcommand{\bubble}[4]{%
  \tcbox[
    halign = left,
    arc=4.5mm,
    colback=#1,
    colframe=#1,
    tcbox width=auto limited, 
    width=\columnwidth - 2cm, 
    #2,
  ]{\color{#3}#4}%
}

\ExplSyntaxOn
\seq_new:N \l__ooker_bubbles_seq
\tl_new:N \l__ooker_bubbles_first_tl
\tl_new:N \l__ooker_bubbles_last_tl

\NewEnviron{rightbubbles}
 {
  \begin{flushright}
  \sffamily
  \seq_set_split:NnV \l__ooker_bubbles_seq { \par } \BODY
  \int_compare:nTF { \seq_count:N \l__ooker_bubbles_seq < 2 }
   {
    \bubble{bubblepurple}{rounded~corners}{white}{\BODY}\par
   }
   {
    \seq_pop_left:NN \l__ooker_bubbles_seq \l__ooker_bubbles_first_tl
    \seq_pop_right:NN \l__ooker_bubbles_seq \l__ooker_bubbles_last_tl
    \bubble{bubblegreen}{sharp~corners=southeast}{white}{\l__ooker_bubbles_first_tl}
    \par\nointerlineskip
    \addvspace{\bubblesep}
    \seq_map_inline:Nn \l__ooker_bubbles_seq
     {
      \bubble{bubblegreen}{sharp~corners=east}{white}{##1}
      \par\nointerlineskip
      \addvspace{\bubblesep}
     }
    \bubble{bubblegreen}{sharp~corners=northeast}{white}{\l__ooker_bubbles_last_tl}
    \par
   }
   \end{flushright}
 }
\NewEnviron{leftbubbles}
 {
  \begin{flushleft}
  \sffamily
  \seq_set_split:NnV \l__ooker_bubbles_seq { \par } \BODY
  \int_compare:nTF { \seq_count:N \l__ooker_bubbles_seq < 2 }
   {
    \bubble{bubblegray}{rounded~corners}{black}{\BODY}\par
   }
   {
    \seq_pop_left:NN \l__ooker_bubbles_seq \l__ooker_bubbles_first_tl
    \seq_pop_right:NN \l__ooker_bubbles_seq \l__ooker_bubbles_last_tl
    \bubble{bubblegray}{sharp~corners=southwest}{black}{\l__ooker_bubbles_first_tl}
    \par\nointerlineskip
    \addvspace{\bubblesep}
    \seq_map_inline:Nn \l__ooker_bubbles_seq
     {
      \bubble{bubblegray}{sharp~corners=west}{black}{##1}
      \par\nointerlineskip
      \addvspace{\bubblesep}
     }
    \bubble{bubblegray}{sharp~corners=northwest}{black}{\l__ooker_bubbles_last_tl}\par
   }
  \end{flushleft}
 }
\ExplSyntaxOff

\definecolor{codegreen}{rgb}{0,0.6,0}
\definecolor{codegray}{rgb}{0.5,0.5,0.5}
\definecolor{codepurple}{rgb}{0.58,0,0.82}
\definecolor{backcolour}{rgb}{0.95,0.95,0.92}

\lstdefinestyle{mystyle}{
  backgroundcolor=\color{backcolour},   commentstyle=\color{blue},
  keywordstyle=\color{codegreen},
  numberstyle=\tiny\color{codegray},
  stringstyle=\color{codepurple},
  basicstyle=\ttfamily\footnotesize,
  breakatwhitespace=false,         
  breaklines=true,                 
  captionpos=b,                    
  keepspaces=true,                 
  numbers=left,                    
  numbersep=5pt,                  
  showspaces=false,                
  showstringspaces=false,
  showtabs=false,                  
  tabsize=2
}

\journal{Information Sciences}

\begin{document}
\begin{frontmatter}

\title{Prolog-based agnostic explanation module for structured pattern classification}


\author[tilburg]{Gonzalo N\'apoles\corref{mycorrespondingauthor}}
\address[tilburg]{Department of Cognitive Science \& Artificial Intelligence, Tilburg University, The Netherlands.}

\cortext[mycorrespondingauthor]{Corresponding author}
\ead{g.r.napoles@uvt.nl}

\author[tilburg]{Fabian Hoitsma}

\author[tilburg]{Andreas Knoben}

\author[warsaw]{Agnieszka Jastrzebska}
\address[warsaw]{Faculty of Mathematics and Information Science, Warsaw University of Technology, Poland.}

\author[miami]{Maikel Leon Espinosa}
\address[miami]{Department of Business Technology, Miami Herbert Business School, University of Miami, USA.}

\begin{abstract}
This paper presents a Prolog-based reasoning module to generate counterfactual explanations given the predictions computed by a black-box classifier. Our approach comprises four well-defined stages that can be applied to any structured pattern classification problem. Firstly, we pre-process the given dataset by imputing missing values and normalizing the numerical features. Secondly, we transform numerical features into symbolic ones using fuzzy clustering such that extracted fuzzy clusters are mapped to an ordered set of predefined symbols. Thirdly, we encode instances as a Prolog rule using the nominal values, the predefined symbols, the decision classes, and the confidence values. Fourthly, we compute the overall confidence of each Prolog rule using fuzzy-rough set theory to handle the uncertainty caused by transforming numerical quantities into symbols. This step comes with an additional theoretical contribution to a new similarity function to compare the previously defined Prolog rules involving confidence values. Finally, we implement a chatbot as a proxy between humans and the Prolog-based reasoning module to resolve natural language queries and generate counterfactual explanations. During the numerical simulations using synthetic datasets, we study the performance of our system when using different fuzzy operators and similarity functions.
\end{abstract}

\begin{keyword}
Explainable artificial intelligence \sep counterfactual explanations \sep symbolic reasoning \sep fuzzy clustering \sep fuzzy-rough sets.
\end{keyword}

\end{frontmatter}


\section{Introduction}
\label{sec:introduction}

Within the Artificial Intelligence (AI) contributions in recent years, one can observe a growing number of approaches related to model explainability and interpretability \cite{gilpin2019explaining}. In simple terms, model explainability is a property that relies on the ability to describe how a given algorithm works, or more specifically, how a given algorithm decides for certain input data. Such explanations can be generated using the model's knowledge structures or an agnostic proxy method. In contrast, model interpretability comprises a spectrum given by three properties \cite{Lipton2018,Barredo2020}: \textit{transparency}, \textit{decomposability} and \textit{simulatability}, although the last one could be subject to debate. Unfortunately, state-of-the-art algorithms devoted to solving complex problems rarely have these valuable properties alone.

As many domains (e.g., healthcare, education, politics, and law) look to deploy deep learning systems, accountability \cite{Barredo2020}, and transparency \cite{Roscher2020} have become of paramount relevance. One of the reasons is that if we cannot deliver explainability properly, we will be limiting AI applications' potential scope. However, aside from the legal and professional considerations that need to be made, there is also an argument that improving explainability is essential, even in more prosaic business scenarios. Understanding how an algorithm works can help align better the activities of data scientists and analysts with the key questions and needs of their organization.

The literature reports interesting agnostic post-hoc methods to produce explanations from black boxes, but they involve some limitations. The first shortcoming is the insufficient focus on symbolic explanations. This results in a relatively inflexible approach to generating descriptions, where descriptions are almost entirely static (such as in the dictionary approaches mentioned by Stepin et al.~\cite{Stepin2021}). Moreover, there is a clear benefit in producing explanations using comprehensible symbols that humans can easily understand. The second neglected aspect is the automatic quality evaluation of the generated explanations. Relating generated descriptions with a robust mathematical approach is essential to quantify the explanations' confidence. This aspect is almost entirely neglected in the literature. Lastly, many existing methods are tailored to work only with a particular machine learning model. For example, Hatwell et al.~\cite{Hatwell2020} delivered a technique to be used with random forests, while Montavon et al.~\cite{Montavon2018} proposed an approach with a limited application to deep neural networks.

The issues mentioned above serve as the motivation behind our proposal. In particular, we see the problem of generating explanations from black boxes as a reasoning problem rather than a merely descriptive task. We will focus on producing counterfactual explanations and resolving natural language queries to accomplish our goal. The counterfactual explanation basis can be summarized with the following question: what small changes could be made to the feature vector to reach the desired outcome?

Therefore, the main contribution of this paper consists of a symbolic reasoning module to generate counterfactual explanations from the predictions computed by a black-box classifier. The module can also resolve what-if queries using ground-truth labels instead of the predicted ones. The proposed approach uses a potentially trainable knowledge base consisting of Prolog fuzzy rules mined from a set of labeled instances. The workflow of our proposal is briefly described as follows. After imputing the missing values and normalizing the numerical features, we build a set of fuzzy prototypes to describe each numerical feature during the granulation step. Such prototypes are obtained using a fuzzy clustering algorithm and subsequently associated with a predefined set of symbols that fulfill an order relation. Afterward, we create the Prolog knowledge base where problem instances are encoded as fuzzy symbolic rules such that each symbol is paired with its confidence value. To quantify the confidence of explanations, we compute the confidence of each rule using fuzzy-rough sets as defined by the membership value of each rule to the corresponding positive region. In that regard, we propose a new parametrized similarity function to compare Prolog fuzzy rules, focusing on the certainty of symbolic terms in the rule antecedent. Finally, we implement a chatbot as a proxy between humans and the Prolog-based reasoning module to resolve natural language queries and generate counterfactual explanations.

It should be stated that the trainable component mentioned above concerns the parameters attached to the fuzzy granulation process, such as the fuzzy implicators or the similarity function used to build the fuzzy-rough sets. However, this paper does not explore such training capabilities beyond a sensitivity analysis conducted with synthetic datasets.

The remainder of the paper is organized as follows. Section \ref{sec:literature} reports relevant research on the generation of agnostic explanations. Section \ref{sec:prolog} elaborates on the knowledge base construction steps related to extracting the fuzzy prototypes from numerical representations and creating the Prolog fuzzy rules. Section \ref{sec:uncertainty} explains how to compute the rule confidence using fuzzy-rough sets, while Section \ref{sec:queries} briefly goes over the technical details concerning the chatbot handling the natural language queries. Section \ref{sec:simulations} conducts a sensitivity analysis and presents a proof-of-concept with several use cases. Section \ref{sec:remarks} provides concluding remarks and comments on future research endeavors.

\section{Related work on agnostic explanations}
\label{sec:literature}

For a long time, the non-direct interpretation issue of popular machine learning models (such as random forests, gradient boosting machines, and neural networks) has been recognized and reported \cite{Murdoch2019}. The existing approaches that aim to explain the inner workings of black-box data processing mechanisms differ in several aspects. Firstly, we distinguish models that operate globally and locally. The former describes a machine learning model behavior as a whole. An example of such an approach is the accumulated local effects method \cite{Apley2020}. The latter group aims to explain individual predictions. This section focuses on methods falling into this category since the proposed model belongs to this group. 

An essential ability of local models is to describe the contribution of each feature to the decisions. We may distinguish several approaches focusing on features in the existing state-of-the-art solutions. Examples of such approaches are LIME (Local Interpretable Model-agnostic Explanations) \cite{Ribeiro2016} and SHAP (SHapley Additive exPlanations) \cite{lundberg2017unified}. LIME explains the predictions of any model by building a white-box local surrogate model. This interpretable surrogate model can be used to explain individual prediction. An important disadvantage reported in \cite{zhang2019why} is that the process of building LIME explanations is nondeterministic, which intuitively goes against the idea that explanatory models should be transparent \cite{Lundberg2020}. SHAP is a game-theoretic approach to explaining the predictions of a machine learning algorithm. This method connects optimal credit allocation with local explanations using the classical Shapley values from game theory. The authors in \cite{Kumar2020} show that mathematical problems arise when Shapley values are used for feature importance and that the mitigating solutions necessarily induce further issues. In addition, they argue that Shapley's values are not a natural solution to the human-centric goals of explainability. 

In the subject domain, a much-desired property is model-agnosticism, which means the method is suitable for explaining the predictions of any prediction-oriented model. On the opposite side of the spectrum, we have model-specific approaches. In the group of model-specific approaches, we shall mention the high share of approaches paired with neural models, which pose particular troubles when interpreting their behavior. We can mention, for example, LRP (Layer-wise Relevance Propagation technique) \cite{Bach2015,Montavon2018} and NAMs (Neural Additive Models) \cite{Agarwal2020}. LRP can explain the predictions of a complex neural network by linking the network's output with input features. The method propagates the prediction backward in the network using propagation rules to produce the explanation. NAMs is a method dedicated to deep neural network classifiers and is constructed as a linear combination of neural networks. NAM describes the relationship between the given input feature and the produced output. 

Generating counterfactual explanations is often solved with decision rules to provide an end-user with an explanation of the reasons that led to the final prediction. Examples of rule-generating approaches are explanations with high-precision rules called anchors, representing local, sufficient conditions for predictions \cite{Ribeiro0G18}, methods extracting decision trees from trained neural networks \cite{Boz2002}, Collection of High Importance Random Path Snippets (CHIRPS) \cite{Hatwell2020}, LORE (LOcal Rule-based Explanations) \cite{Guidotti2018}, and Multi-Objective Counterfactuals (MOC) method \cite{Dandl2020}. Let us briefly describe the inner working of the last three methods as they generate counterfactual explanations.

CHIRPS extracts a decision path from each tree in the forest that contributes to the majority classification and then uses frequent pattern mining to identify the most commonly occurring split conditions. Then, a conjunctive form rule is constructed where the antecedent terms are derived from the attributes that had the most influence on the classification. This rule is returned alongside estimates of the rule's precision and coverage of the training data, along with counterfactual details. LORE is model-agnostic and aims to find a local interpretable predictor on a synthetic neighborhood generated by a genetic algorithm. Then it derives from the logic of the local interpretable predictor a meaningful explanation consisting of a decision rule -- which explains the reasons for the decision -- and a set of counterfactual rules, suggesting the changes in the instance's features leading to a different outcome. MOC provides counterfactual explanations of how given features contributed to a decision for a single problem instance. The advantage of MOC is that it is model-agnostic and works for numerical and categorical input features. It relies on solving a multi-objective optimization problem -- it searches for the optimal set of counterfactual explanations in a predefined search space. An open issue for MOC is letting users select the counterfactuals that meet their particular trade-off between the objectives. Furthermore, we should mention the method by Wachter et~al.~\cite{Wachter2018}, which minimizes a loss function between a model prediction for the given counterfactual and the desired outcome. Unfortunately, this method is not scalable and does not handle categorical features.

Yet another strategy of generating counterfactual descriptions for a trained machine learning model that should be mentioned involves generating an opposite prediction and then describing the differences that had to take place to get the opposite prediction. Examples of this approach include Model Agnostic suPervised Local Explanations (MAPLE) \cite{Plumb2018}, DIverse Counterfactual Explanations (DICE) \cite{Mothilal2020}, and Feasible and Actionable Counterfactual Explanations (FACE) \cite{Poyiadzi2020}. MAPLE runs post-hoc and is local (concerns a single instance). It combines random forests with feature selection methods to return counterfactual explanations based on feature importance. DICE is based on an optimization mechanism with constraints. FACE focuses on returning ``reachable'' counterfactual explanations (the simplest paths leading to explanations are considered, and the whole model is a graph).

We should mention that Hoitsma et al.~\cite{Hoitsma2020} presented an approach for generating counterfactual explanations, which is conceptually related to the algorithm presented in this paper. The cited article discusses a method that produces counterfactual explanations from the predictions computed by a Fuzzy Cognitive Map-based classifier. This method outputs a set of Prolog rules where the feature vector is encoded as a symbolic vector denoting the rule antecedent, while the rule consequent encodes the classifier's predictions. However, the approach in \cite{Hoitsma2020} has several limitations. Firstly, symbolic descriptions are produced using a relatively inconvenient approach since the domain expert must fine-tune Gaussian membership functions manually. Secondly, the proposed module does not have an option to output sentences in natural language. Finally, it lacks a mechanism to evaluate the consistency of the generated fuzzy rules. These issues are resolved in the new algorithm presented in the following sections.

\section{Building the Prolog knowledge base}
\label{sec:prolog}

This section elaborates on the steps needed to build the Prolog knowledge base consisting of a set of fuzzy rules. Firstly, we transform numerical quantities into meaningful symbols using a fuzzy clustering algorithm. Relying on a symbolic paradigm is challenging for AI systems, as pointed out by d'Avila Garcez et al. \cite{Garcez2015}. Existing studies have demonstrated that such an approach can enhance the explanation capability of learning systems \cite{Garcez2001, Manhaeve2019}. Secondly, we create fuzzy Prolog rules using the derived symbolic terms and the decision classes determined by the classifier. Besides, each symbolic term is attached with a confidence value to assess the accuracy of the symbolic representation. It should be stated that the literature includes other approaches to Prolog knowledge base construction and uncertainty management, such as the probabilistic approach discussed by De Raedt and Kimmig \cite{Raedt2015}. However, probability theory, fuzzy sets, and fuzzy-rough sets tackle different types of uncertainties and replacing one with another often leads to misleading interpretations. In this paper, we use fuzzy set theory as a mathematical model for imprecision and rough set theory as a mathematical model for inconsistency. The former is used when obtaining the symbolic knowledge representations, while the latter is necessary when quantifying the consistency of each rule.
	
\subsection{Deriving the symbolic terms}
\label{sec:prolog:symbols}

Structured pattern classification problems regularly involve both discrete and continuous features. While the values of discrete features can be included in a Prolog knowledge base, performing symbolic reasoning with constant features might be far from ideal. This section shows the use of fuzzy logic theory to transform numerical features into symbolic knowledge representation and quantify the uncertainty of such a transformation.

Let $\mathbf{F} = \mathbf{N} \cup \mathbf{D}$ the set of problem features describing a specific pattern classification problem where $\mathbf{N}$ is the set of numerical features while $\mathbf{D}$ is the set of discrete features. As mentioned, the values of $f_i \in \mathbf{D}$ are symbols, and their confidence degrees are assumed to be one.

Aiming at obtaining meaningful symbols from numerical quantities, we build $c$ fuzzy sets for each feature $f_i \in \mathbf{N}$. Each fuzzy set is associated with a linguistic term (e.g., \textit{low}, \textit{medium}, and \textit{high}) that provides meaning to the fuzzy set. Therefore, in our approach, meaning is closely related to the magnitude of numerical representations on a normalized scale. The number of fuzzy sets defines the granularity level of explanations: the more fuzzy sets, the more detailed the explanations.

In this paper, the fuzzy sets for each numerical feature are determined by running the fuzzy $c$-means algorithm \cite{Bezdek1984} on a two-dimensional space where each data point $x_i$ is represented by a symmetric tuple $(x_i,x_i)$. The algorithm will discover $c$ fuzzy sets along the identity line. The fuzzy $c$-means algorithm creates a membership matrix $\mathbf{U}_{k \times c}$ where $k$ is the number of data points to be processed (i.e., the number of instances in the dataset). Hence, $\mu_{ij} \in \mathbf{U}$ represents the degree to which the $i$-th data point belongs to the $j$-th fuzzy cluster. Furthermore, the algorithm returns a matrix of prototypes $\mathbf{Z}_{1 \times c}$ denoting the cluster centers. The fuzziness of this clustering algorithm is controlled by a~fuzzification coefficient $m \in[1,\infty]$, where larger values indicate more fuzziness. Equations \eqref{eqn:fcmeans_memfun} and \eqref{eqn:fcmeans_cj} display how to compute the membership values and the prototypes:

\begin{equation}
\mu_{ij} = \frac{1}{\sum\limits_{l=1}^c \Big( \frac{\Vert x_i - z_j \Vert}{\Vert x_i - z_l \Vert} \Big)^{2/(m-1)} },
\label{eqn:fcmeans_memfun}
\end{equation}
\begin{equation}
z_j = \frac{\sum\limits_{i=1}^k \mu_{ij}^m \cdot x_i}{\sum\limits_{i=1}^k \mu_{ij}^m}.
\label{eqn:fcmeans_cj}
\end{equation}

After computing the fuzzy sets, we must associate each data point with a linguistic term. Let $\mathbf{L}=\{l_i\}_{i=1}^c$ denote a partially ordered set consisting of $c$ linguistic terms such that $l_i$ precedes $l_{i+1}$ in the ordering (e.g., \textit{low} precedes \textit{medium}). If we order the prototypes in $\mathbf{Z}$ in ascending order according to their norm, we can establish a direct mapping between the symbolic terms in $\mathbf{S}$ and the prototypes in $\mathbf{Z}$. Finally, we can represent the numerical quantity $x_i$ with the fuzzy set reporting the largest membership value. This can be done with a function $\mu_{l_j}(x_i)$ that produces its maximal value when $x_i$ reports the largest membership value to the $j$-th fuzzy set.

The temporal complexity of this step is $O(|\mathbf{N}|)O(|\mathbf{F}|kc)$, where $|\mathbf{N}|$ gives the number of numerical features, $|\mathbf{F}|$ stands for the number of features, $c$~is the number of linguistic terms used to describe the numerical features, whereas $k$ is the number of instances. In short, such temporal complexity comes from applying the fuzzy $c$-means algorithm to each numerical feature in the dataset during the granulation step. Notice that this process can be parallelized concerning the numerical features, thus significantly speeding up the knowledge base construction process.

\subsection{Deriving the fuzzy Prolog rules}
\label{sec:prolog:rules}

The next step is building the fuzzy rules to be translated into a Prolog knowledge base. The fuzzy rules are symbolic because numerical values will be encoded as symbols using the procedure described in the previous sub-section. Such rules will include two types of confidence values. The confidence value associated with each linguistic term denotes how accurately the symbolic value represents the numerical ones. The confidence value associated with each rule indicates the extent to which that rule does not conflict with others. In the next section, we will explain how to compute the confidence values of rules using fuzzy-rough sets.

Let $\mathbf{S}$ be the unordered version of $\mathbf{L}$ where sub-indices no longer indicate that a linguistic term precedes another. The fuzzy rules have the structure: ``IF $f_1 \in \mathbf{N}$ is $s_1 \in \mathbf{S}$ with certainty $\mu_{s_1}(x_1)$ AND $\ldots$ AND $f_i \in \mathbf{N}$ is $s_i \in \mathbf{S}$ with certainty $\mu_{s_i}(x_i)$ AND $\ldots$ AND $f_n \in \mathbf{N}$ is $s_n \in \mathbf{S}$ with certainty $\mu_{s_n}(x_n)$ THEN $y_j$ with certainty $P(y_j|x_1,x_2,\ldots,x_n)$". In these rules, $x_i$ denotes the value of the $i$-th feature while $s_i$ represents the corresponding symbolic term, whereas $P(y_j|x_1,x_2,\ldots,x_n)$ is the confidence to which the classifier labeled the original instance with the decision class $y_j \in \mathbf{Y}$ given an instance with feature values $x_1,x_2,\ldots,x_n$. In the case of discrete features, we assume that $s_i=x_i$ and $\mu_{s_i}(x_i)=1$ to make the notation consistent.

The next step concerns the creation of a Prolog knowledge base to resolve natural language queries. In particular, we are interested in producing explanations for counterfactual queries with the format: ``Which symbolic values should the features in $\mathbf{Q} \subset \mathbf{F}$ have taken to produce $y_j \in \mathbf{Y}$ instead of $y_i \in \mathbf{Y}$, given the symbolic values of features in $\mathbf{G} \subset \mathbf{F}$, such that $\mathbf{Q} \neq \emptyset$, $\mathbf{G} \neq \emptyset$ and $\mathbf{Q} \cap \mathbf{G} = \emptyset$?''. The experts are expected to provide the set of features $\mathbf{Q} \subset \mathbf{F}$ to be investigated, the symbolic values of known features $\mathbf{G} \subset {F}$ and the alternative outcome $y_j \in \mathbf{Y}$. It should be mentioned that we can resolve these queries even when $\mathbf{Q} \cup \mathbf{G} \neq \mathbf{F}$. It is assumed that all numerical features have already been mapped to symbolic ones while retaining nominal ones defined in the original dataset.

Aiming at translating the fuzzy rules into Prolog ones, we rely on two predicates named \texttt{input} and \texttt{output}. The \texttt{input} predicate involves the symbolic terms describing a given instance and its confidence values, whereas the \texttt{output} predicate involves the decision class and the classifier's confidence. The Prolog rules have the format: \texttt{input}(id, [[$s_1$,$\mu_{s_1}(x_1)$],$\ldots$,[$s_i$,$\mu_{s_i}(x_i)$]$ $ $,\ldots$, [$s_n$,$\mu_{s_n}(x_n)$], $\mu_{R_*(\mathbf{\Theta}_j)}(x)$]) :- \texttt{output}([$y_j$,$P(y_j|x_1,x_2,\ldots,x_n)$]), where $\mu_{R_*(\mathbf{\Theta}_j)}(x)$ stands for the confidence of the whole rule (to be addressed in the following section). Observe that the antecedent and consequent in these Prolog rules are reversed when compared with the fuzzy rules defined above to match the inner semantics of counterfactual questions.

During querying, experts need to provide all the pieces of knowledge they have to resolve the query. This can be done by (i) asserting the dynamic predicate \texttt{output} with the desired outcome, (ii) instantiating some variables corresponding to features in $\mathbf{G} \subset \mathbf{F}$ when calling the \texttt{input} predicate, and (iii) asking for the symbolic values and confidences attached to features in $\mathbf{Q} \subset \mathbf{F}$. Moreover, it is advised to retract all asserted predicates before running any new query to avoid unexpected issues. The Prolog queries have the following format: \texttt{retractall}(\texttt{output}()), \texttt{assertz}(\texttt{output}([$y_j$,$p_0$])), \texttt{input} ([[$v_1$,$p_1$], $\ldots$, [$v_i$,$p_i$], $\ldots$, [$v_n$,$p_n$], $p_{n+1}$]), where $v_i$ is either the anonymous variable or a symbolic term used to instantiate the $i$-th symbolic variable, while $p_i$ is either the anonymous variable or a confidence value. In addition, we can add constraints to symbolic terms or confidence values associated with variables to be resolved. The same applies to the confidence values of rules. For example, we can ask for symbolic solutions excluding certain linguistic terms with a confidence value greater than 0.6, which appear in rules with a confidence degree greater than 0.8. Section \ref{sec:simulations} will introduce a case study illustrating the querying features.

It is worth mentioning that although the knowledge base is composed of fuzzy Prolog rules, the symbolic reasoning is crisp. Therefore, after resolving a query, confidence values will be used to assess the extent to which the user can trust the explanations generated by the module. 

\section{Uncertainty quantification of fuzzy rules}
\label{sec:uncertainty}

In the procedure described in the previous section, each symbolic term resulting from the fuzzy granulation process was attached with a confidence value. However, such a confidence value would not provide information on how a rule conflicts with others. In this section, we develop a rule confidence measure based on the fuzzy-rough set formalism \cite{Dubois1990, Inuiguchi2015} that considers the certainty of symbolic values involved in both the antecedent and the consequent of fuzzy Prolog rules.

Let $\mathbf{\Omega}$ be the set of fuzzy rules, $\mathbf{\Theta} \in \mathbf{\Omega}$ a fuzzy set with membership function $\mu_\mathbf{\Theta}(x)$ and $R \in \Psi(\mathbf{\Omega} \times \mathbf{\Omega})$ a fuzzy binary relation with membership function $\mu_R(y,x)$. The membership function $\mu_\mathbf{\Theta} : \mathbf{\Omega} \rightarrow [0,1]$ determines the degree to which $x \in \mathbf{\Omega}$ is a member of $\mathbf{\Theta}$, while $\mu_R: \mathbf{\Omega} \times \mathbf{\Omega} \rightarrow [0,1]$ is the degree to which $y$ is a member of $\mathbf{\Theta}$ taking into account the extent to which $x$ is a member of $\mathbf{\Theta}$. If no confusion arises, we will represent $R(x)$ with its membership function $\mu_{R(x)}(y)=\mu_R(y,x)$ to lighten the notation. Finally, let $\mathbf{\Theta}_j \in \mathbf{\Omega}$ denote the set of fuzzy rules associated with the $j$-th decision class as determined by the black-box classifier.

The first step to derive the rule certainty measure is to build a fuzzy set for every rule set $\mathbf{\Theta}_j$ resulting from partitioning $\mathbf{\Omega}$. Equation \eqref{eq:membership-decision} shows the corresponding membership function,

\begin{equation}
\label{eq:membership-decision}
\mu_{\mathbf{\Theta}_j}(x) = \left \{ \begin{matrix} P(y_j|x), & x \in \mathbf{\Theta}_j \\
		 0, & x \notin \mathbf{\Theta}_j \end{matrix} \right.
\end{equation}

\noindent where $P(y_j|x)$ denotes the confidence of decision class $y_j$ being correct given the instance encoded with the fuzzy rule $x \in \mathbf{\Omega}$. In a nutshell, $P(y_j|x)$ is the confidence attached to that decision class. If these probabilities are not available, then we can make $P(y_j|x)=1$ when $x \in \mathbf{\Theta}_j$.

The second step concerns the fuzzy binary relation $R \in \Psi(\mathbf{\Omega} \times \mathbf{\Omega})$ with membership function $\mu_R(y,x)$ such that $x,y \in \mathbf{\Omega}$. The latter function can be computed by combining the piece of knowledge $\mu_{\mathbf{\Theta}_j}(x)$ with the similarity value between fuzzy rules $x$ and $y$. Equation \eqref{eq:fuzzy-relation} formalizes how to compute the $\mu_R(y,x)$ function following this reasoning,

\begin{equation}
\label{eq:fuzzy-relation}
\mu_R(y,x) = \mu_{\mathbf{\Theta}_j}(x) e^{-\lambda d(x,y)}
\end{equation}

\noindent such that $\lambda > 0$ is a smoothing parameter and

\begin{equation}
\label{eq:distance-global}
d(x,y) = \sum_{i=1}^{|\mathbf{F}|} \sigma_i(x,y)
\end{equation}
		
\noindent where $\sigma_i(x,y)$ is a feature-wise distance function that quantifies the dissimilarity between two linguistic terms. Equation \eqref{eq:distance-local1} shows a simple feature-wise distance function used to compare nominal features, as turns out to be the case of symbolic terms describing a numerical quantity, 

\begin{equation}
\label{eq:distance-local1}
\sigma_i(x,y) = \left\{
\begin{array}{ll}
      0, & s(y_i) = s(x_i) \\
      1, & s(y_i) \neq s(x_i). \\
\end{array} 
\right. 
\end{equation}

The primary limitation of the feature-wise distance function depicted in Equation \eqref{eq:distance-local1} is that it does not consider the membership values associated with the linguistic terms. However, it might be that two numeric values are described by the same linguistic terms, having attached quite different membership values. Equation \eqref{eq:distance-local2} presents a fuzzy feature-wise distance function that tackles such a limitation, 

\begin{equation}
\label{eq:distance-local2}
\sigma_i(x,y) = \left\{
\begin{array}{cc}
      \Delta_i(x,y), & s(y_i) = s(x_i) \\
      1, & s(y_i) \neq s(x_i) \\
\end{array} 
\right. 
\end{equation}

\noindent such that $\Delta_i(x,y)=0.5 \cdot (1-\min \{\mu_{s(x_i)}(x_i),\mu_{s(y_i)}(y_i) \})$. It should be highlighted that Equation \eqref{eq:distance-local2} uses both the symbolic terms and their membership degrees when computing the distance between the fuzzy rules being compared. More explicitly, the dissimilarity is computed using the minimal membership degree attached to the symbolic terms in the antecedent of the rules. This value is subtracted since the dissimilarity between two symbolic values should increase as the minimal membership value decreases. Moreover, we enforce a constraint to ensure that the maximal value reported by our feature-wise distance function when the symbolic terms match is not greater than 0.5. Overall, we can assert that $d(x,y)$ will report differences even when all linguistic terms in the antecedent of rules $x$ and $y$ match, but their membership values are not maximal.

Finally, we can compute the confidence of rule $x \in \mathbf{\Omega}$ as the membership degree of $x$ to the fuzzy-rough positive region \cite{Inuiguchi2015} associated with the rule's decision class. Fuzzy-rough positive regions are equivalent to the fuzzy-rough lower approximations, which can be understood as the fuzzy set of rules that do not conflict with each other (i.e., rules having a reasonably similar antecedent have different consequents). To obtain the fuzzy-rough lower approximations, we can use the degree of $x$ being a member of $\mathbf{\Theta}_j$ defined by the fuzzy binary relation. This can be measured by the truth value of the statement ``$y \in R(x)$ implies $y \in \mathbf{\Theta}_j$" under fuzzy sets $R(x)$ and $\mathbf{\Theta}_j$. The authors in \cite{Inuiguchi2015} suggested using a necessity measure $\inf_{y \in \mathbf{\Omega}} \mathcal{I}(\mu_R(y,x), \mu_{\mathbf{\Theta}_j}(y))$ for such a quantification, where $\mathcal{I}:[0,1] \times [0,1] \rightarrow [0,1]$ is a fuzzy implication function. We will conduct a sensitivity analysis during numerical simulations involving several fuzzy implicators.  

Equation \eqref{eq:confidence} shows the membership function for the lower approximation set $R_*(\mathbf{\Theta}_j)$ associated with the $j$-th decision class,

\begin{equation}
\label{eq:confidence}
\mu_{R_*(\mathbf{\Theta}_j)}(x) = \min \left\lbrace \mu_{\mathbf{\Theta}_j}(x), \inf_{y \in \mathbf{\Omega}} \mathcal{I}(\mu_R(y,x), \mu_{\mathbf{\Theta}_j}(y)) \right\rbrace.
\end{equation}

It can be observed that this fuzzy derivation of the rough lower approximations does not assume that $\mu_R(x,x)=1, \forall x \in \mathbf{\Omega}$. Instead, we compute the minimum between $\mu_{\mathbf{\Theta}_j}(x)$ and $\inf_{y \in \mathbf{\Omega}} \mathcal{I}(\mu_R(y,x), \mu_{\mathbf{\Theta}_j}(y))$ to preserve the inclusiveness of $R_*(\mathbf{\Theta}_j)$ in the fuzzy set $\mathbf{\Theta}_j$. Similarly, we could compute the fuzzy-rough lower approximations and the positive and boundary regions; however, they are irrelevant to our study.

Roughly speaking, quantifying the certainty of fuzzy rules in the knowledge base has a temporal complexity of $O(|\mathbf{Y}|)O(|\mathbf{F}||\mathbf{\Omega}|^2)$. The whole process can be parallelized concerning the decision classes, which might be convenient when processing large datasets.

\section{Natural language Prolog queries}
\label{sec:queries}

In this section, we go over the technical details concerning the chatbot handling the natural language queries posed by users. Moreover, we formalize the structure of these questions and gather them into categories ranging from exploratory data analysis to symbolic explanations.

\subsection{Motivation}

A Prolog knowledge base can be queried directly using Prolog, but it might require some expertise from the user concerning declarative programming. Thus, we implemented a conversational agent (chatbot) intending to enable users without knowledge of the Prolog language and syntax to interact with the knowledge base by running natural language queries. By doing that, we used the Rasa platform\footnote{\url{https://rasa.com/}}, which provides an open-source tool that can be used to build conversational virtual assistants. It allows tight integration with the Python programming language, which is especially useful in our case, as the chatbot needs to interact with Prolog. The Python-Prolog interaction is enabled through the \verb|pyswip| module \footnote{\url{https://github.com/yuce/pyswip}}.

Besides running queries on the Prolog knowledge base to answer what-if and counterfactual questions, the conversational agent enables users to explore data, train and validate a classifier, and construct a knowledge base using natural language. Furthermore, the chatbot can answer questions about these steps and create visualizations. 

\subsection{Implementation}

The chatbot in Rasa consists of the following components: Natural Language Understanding (NLU) with intents and entities, rules, actions, and back-end Python script. The NLU component involves example phrases for each intent with annotated entities (if any). An intent represents what the user wants to achieve with the sent message, while an entity is a crucial piece of information within the sent message that needs to be considered in the chatbot's action. An entity can be used as a parameter for the action that follows. The example phrases are the training data for the NLU, which allows it to classify intents and extract entities from new user messages.

The actions are the predetermined responses the chatbot can return, given a specific intent. This can be a simple text or an image and can be extended through custom actions. These actions run using the functions in the back-end Python script. For example, load the data, perform exploratory data analysis and visualization, train the classifier, create the knowledge base and run queries. The rules component specifies what action should be performed according to the classified intent.

\subsection{Question categories}

This section will formalize the different queries supported by the conversational agent using the proposed reasoning module as the back end. The formalization will adopt the same notation as in the previous sections for mathematical rigor, although the query to be resolved should not be expected to contain any mathematical symbol. It shall be noted that the chatbot can handle different formulations for these queries.

\subsubsection{Loading the classification dataset}
The user can request the chatbot to load a certain dataset. The chatbot extracts the provided name of the dataset as an entity and searches for it in a folder with available datasets. The expected files should use the Attribute-Relation File Format (ARFF), an ASCII text file that describes a list of instances sharing a set of problem features.


\subsubsection{Conducting exploratory data analysis}

After loading the dataset, the user can explore the data by asking for visualizations and further information about the data. These can be roughly subdivided into the following categories.

Questions such as ``\textit{Let me see what the dataset contains}'' return the number of instances and the features in the dataset. Also, the correlation between two features can be plotted by asking ``\textit{Show how $f_i \in \mathbf{N}$ and $f_j \in \mathbf{N}$ are correlated}''. Moreover, the correlation matrix can be asked as follows: ``\textit{What does the correlation for the entire dataset look like?}''. The distribution of a given feature can be plotted in a histogram by asking a question such as ``\textit{How are the values for $f_i \in \mathbf{N}$ distributed?}''.

\subsubsection{Training the classification model}

When asked, the chatbot trains a random forest classifier after splitting the data into a training and test set. The chatbot returns the accuracy on the test set, together with a confusion matrix. The user can provide the parameters (n\_estimators, max\_depth) for training the classifier, or default values are used. An example input from the user could be: ``\textit{Train the model with a maximum depth of 7 and use 200 estimators.}'' After the training is complete, the user can ask how the data was split: ``\textit{How did you split the data for training and testing?}''

\subsubsection{Building the explanation module}

Once the classifier has been trained, the user can ask to construct the symbolic explanation module. This can be done by typing: ``\textit{Create the explanation module for this classifier}''. In that way, the symbolic terms for the numerical features and the fuzzy-rough sets, and the knowledge base in Prolog are constructed. The user can also ask for a plot of the membership of each instance to the fuzzy-rough regions: ``\textit{Plot the memberships of all the instances to the fuzzy-rough regions.}''

The user can request top rules in the Prolog knowledge base, for example, by asking: ``\textit{What are top rules in the knowledge base?}''. The top rules are those with the highest membership values to the positive fuzzy-rough region they belong to. Similarly, the difficulty level of the loaded classification problem can be requested by asking: ``\textit{Tell me the complexity of the classification problem please}''. The complexity of the loaded problem is calculated as one minus the average membership values of the instances to the positive fuzzy-rough region they belong to. Moreover, the user can ask for the explicit bias attached to a given protected feature, for example: ``\textit{Can you quantify the bias attached to $f_i \in \mathbf{F}$?}'' The bias is calculated as proposed by \citet{koumeri2021bias} using fuzzy-rough sets.

\subsubsection{Answering what-if questions}

The chatbot allows users to run queries on the Prolog knowledge base without being familiar with the declarative programming paradigm. These queries can be divided into what-if questions and counterfactual questions.
With what-if questions, the user provides symbolic values for selected input features and the decision class and asks for the symbolic values of unknown input features. The structure of these queries and the expected answers can be formalized, as illustrated below.

\begin{addmargin}[2em]{0em}
    \noindent \textit{User:} If $g_1 \in \mathbf{G}$ is $s_1 \in \mathbf{S}$, $\ldots$, $g_a \in \mathbf{G}$ is $s_a \in \mathbf{S}$ and the outcome is $y_{j} \in \mathbf{Y}$, then what is $q_1 \in \mathbf{Q}$, $\ldots$, $q_b \in \mathbf{Q}$?
    
    \noindent \textit{Bot:} I have run the query for you. These are the results: $q_1 \in \mathbf{Q}$ is $s_1 \in \mathbf{S}$ with a certainty of $\mu_{s_1}(x_1)$, $\ldots$, $q_b \in \mathbf{Q}$ is $s_b \in \mathbf{S}$ with a certainty of $\mu_{s_b}(x_b)$. The entire rule has a certainty of $\mu_{R_*(\mathbf{\Theta}_j)}(x)$.
\end{addmargin}

\subsubsection{Answering counterfactual questions}

Counterfactual queries are similar to what-if queries but involve an alternative decision. To do that, the user provides an instantiation for known features and requests symbolic values for certain features to obtain the desired outcome. Counterfactual queries posed by the user and the answers provided by the bot can be formalized as follows.

\begin{addmargin}[2em]{0em}
    \noindent \textit{User: What values should $q_1 \in \mathbf{Q}$, $\ldots$, $q_b \in \mathbf{Q}$ have taken, assuming that $g_1 \in \mathbf{G}$ is $s_1 \in \mathbf{S}$, $\ldots$, $g_a \in \mathbf{G}$ is $s_a \in \mathbf{S}$, for the outcome to be $y_{j}$ instead of $y_{i}$?}
    
    \noindent \textit{Bot: I have run the query for you. These are the results: $q_1 \in \mathbf{Q}$ should be $s_1 \in \mathbf{S}$ with a certainty of $\mu_{s_1}(x_1)$, $\ldots$, $q_b \in \mathbf{Q}$ should be $s_b \in \mathbf{S}$ with a certainty of $\mu_{s_b}(x_b)$. The entire rule has a certainty of $\mu_{R_*(\mathbf{\Theta}_j)}(x)$.}
\end{addmargin}

\vspace{3mm}

While the literature reports other conversational agents that generate explanations, their ability to resolve symbolic counterfactual queries is limited. For example, the agent in \cite{Grau2021} uses decision trees and subgroup discovery algorithms to answer what-if questions in a decision-making context. In contrast, the agent in \cite{chatbot2020} is primarily devoted to answering what-if questions and computing feature relevance scores.



\section{Numerical simulations}
\label{sec:simulations}

In this section, we will assess the proposed Prolog-based reasoning module for symbolic reasoning. Firstly, we will conduct a sensitivity analysis to study the impact of the granulation parameters on the confidence of the induced fuzzy rules. Secondly, we will present a proof-of-concept illustrating prominent use cases of the resulting conversational agent.

\subsection{Sensitivity analysis}
\label{sec:simulations:anaysis}

Aiming to conduct a sensitivity analysis, we will rely on ten structured pattern classification datasets with different properties (see Table \ref{tab:dataset_desc}). The number of instances in these datasets range from 57 to 900, features from 6 to 28, and decision classes from 2 to 11. Moreover, two highly imbalanced datasets are defined by the ratio between the minority and majority classes.

\begin{table}[!ht]
\centering
\footnotesize{
\caption{Properties of each dataset used in the numerical simulations.}
\label{tab:dataset_desc}
\begin{tabular}{|c|L|c|c|c|c|c|}
    \hline
    Index & Dataset & Instances & Numeric & Categorical & Classes & Imbalance\\ \hline
    1 & \multicolumn{1}{m{5cm}|}{statlog: concerns credit card applications in Australia} & 690 & 8 & 6 & 2 & no\\ \hline
    2 & \multicolumn{1}{m{5cm}|}{crx: concerns credit card applications} & 653 & 6 & 9 & 2 & no\\ \hline
    3 & \multicolumn{1}{m{5cm}|}{ecoli: contains protein localization sites} & 336 & 7 & 0 & 8 & 71:1\\ \hline
    4 & \multicolumn{1}{m{5cm}|}{flags: contains details of various nations and their flags} & 194 & 28 & 0 & 8 & 15:1 \\ \hline
    5 & \multicolumn{1}{m{5cm}|}{labor: includes agreements reached in Canada} & 57 & 8 & 8 & 2 & no \\ \hline
    6 & \multicolumn{1}{m{5cm}|}{diabetes: describes main characteristics of the Pima Indians Diabetes study} & 768 & 8 & 0 & 2 & no \\ \hline
    7 & \multicolumn{1}{m{5cm}|}{vehicle: made for the classification of a given silhouette as one of four types of vehicle} & 846 & 18 & 0 & 4 & no \\ \hline
    8 & \multicolumn{1}{m{5cm}|}{vertebral: contains patients diagnosed with disk hernia in France} & 310 & 6 & 0 & 3 & no \\ \hline
    9 & \multicolumn{1}{m{5cm}|}{vowel: relates to the recognition of the eleven steady state vowels of British English} & 990 & 10 & 3 & 11 & no \\ \hline
    10 & \multicolumn{1}{m{5cm}|}{wine: uses chemical analysis to determine the origin of wines} & 178 & 13 & 0 & 3 & no \\ \hline
\end{tabular}
}
\end{table}

The sensitivity analysis attempts to determine the impact of the choice of a fuzzy implicator on the granulation step (i.e., the process of extracting concepts from numerical data). A fuzzy implicator is a mapping used in our model to determine rule confidence (see Equation \eqref{eq:confidence}). There are certain formal assumptions that a fuzzy implicator must satisfy: for any $x,x_1,x_2,y_1,y_2 \in [0,1]$: if $x_1 \leq x_2$, then $\mathcal{I}(x_1,y) \geq \mathcal{I}(x_2,y)$ and if $y_1 \leq y_2$, then $\mathcal{I}(x,y_1) \leq \mathcal{I}(x,y_2)$. In addition, the following boundary conditions must be satisfied: $\mathcal{I}(0,0) = 1$,  $\mathcal{I}(1,0) = 0$, and $\mathcal{I}(1,1) = 1$. The literature concerning fuzzy operators discusses several popular fuzzy implicators \cite{Jayaram2009, Pagouropoulos2020}. In this paper, we will study the fuzzy implicators listed below. We selected them as candidate mappings because of their different behavior for various input values. While Łukasiewicz and Fodor are symmetric and exhibit abrupt changes in their shapes, G\"odel and Goguen are neither symmetric nor change as abruptly as the former operators.

\begin{itemize}
    \item Fodor implicator, given by the following equation:
    \begin{equation}
    \label{eq:fodor}
    \mathcal{I}_{FD}(x,y) = \begin{cases}1,& \mbox{if } x \leq y\\\max(1-x,y),&\mbox{otherwise} \end{cases}
    \end{equation}
    \item Goguen implicator, given by the following equation:
    \begin{equation}
    \label{eq:goguen}
    \mathcal{I}_{GG}(x,y) = \begin{cases}1,& \mbox{if } x \leq y\\\frac{y}{x},&\mbox{otherwise} \end{cases}
    \end{equation}
    \item G\"odel implicator, given by the following equation:
    \begin{equation}
    \label{eq:godel}
    \mathcal{I}_{GD}(x,y) = \begin{cases}1,& \mbox{if } x \leq y\\y,&\mbox{otherwise} \end{cases}
    \end{equation}
    \item Łukasiewicz implicator, given by the following equation:
    \begin{equation}
    \label{eq:luka}
    \mathcal{I}_{GD}(x,y) = \mathcal{I}_{LK}(x,y) = \min(1, 1-x+y)
    \end{equation}
\end{itemize}



\begin{figure*}[!htb]
\center
    \begin{subfigure}{0.49\textwidth}
	\center
	\includegraphics[width=\textwidth]{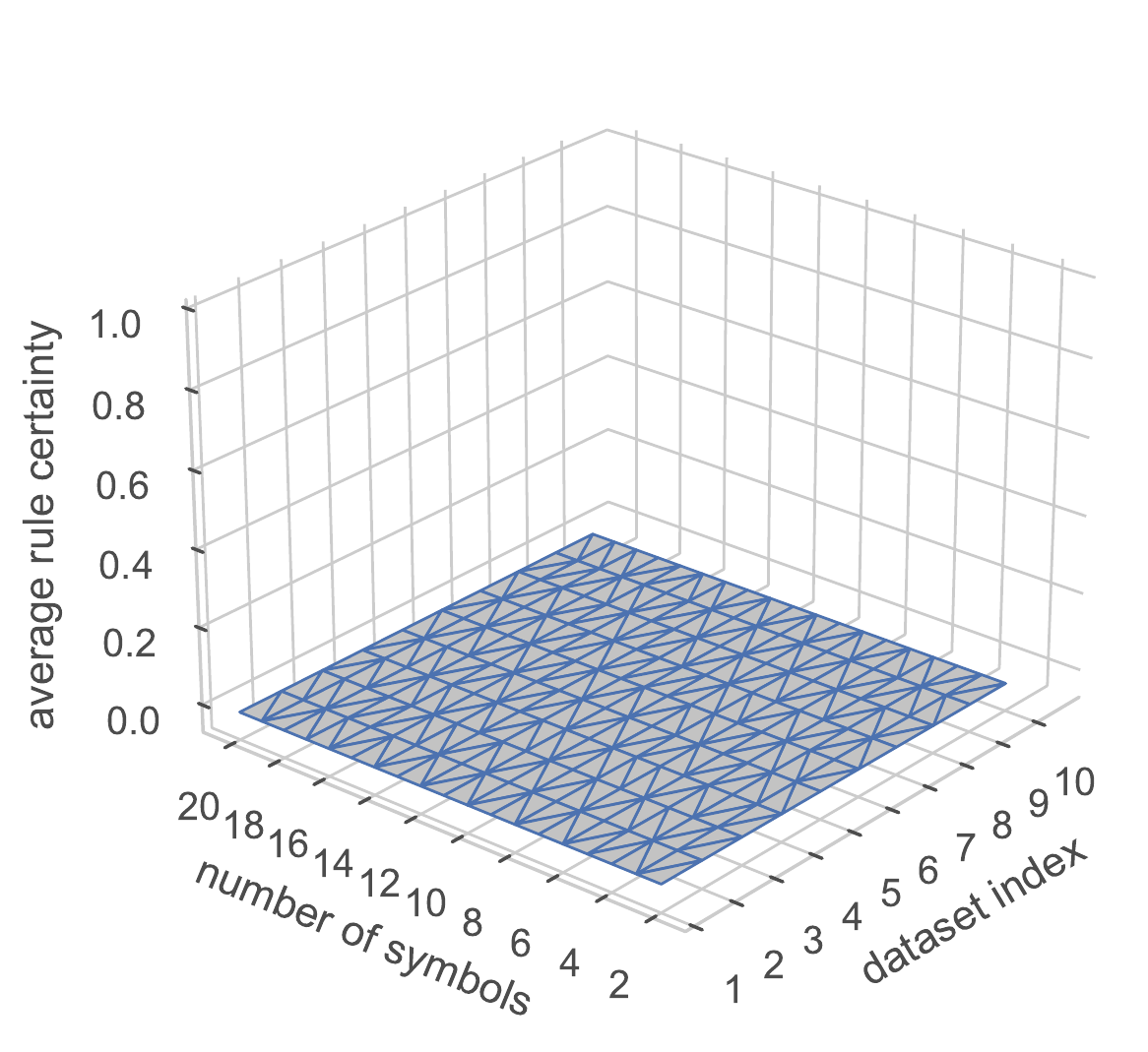}
	\caption{G\"odel and Equation \eqref{eq:distance-local1}}
	\end{subfigure}
	\begin{subfigure}{0.49\textwidth}
	\center
	\includegraphics[width=\textwidth]{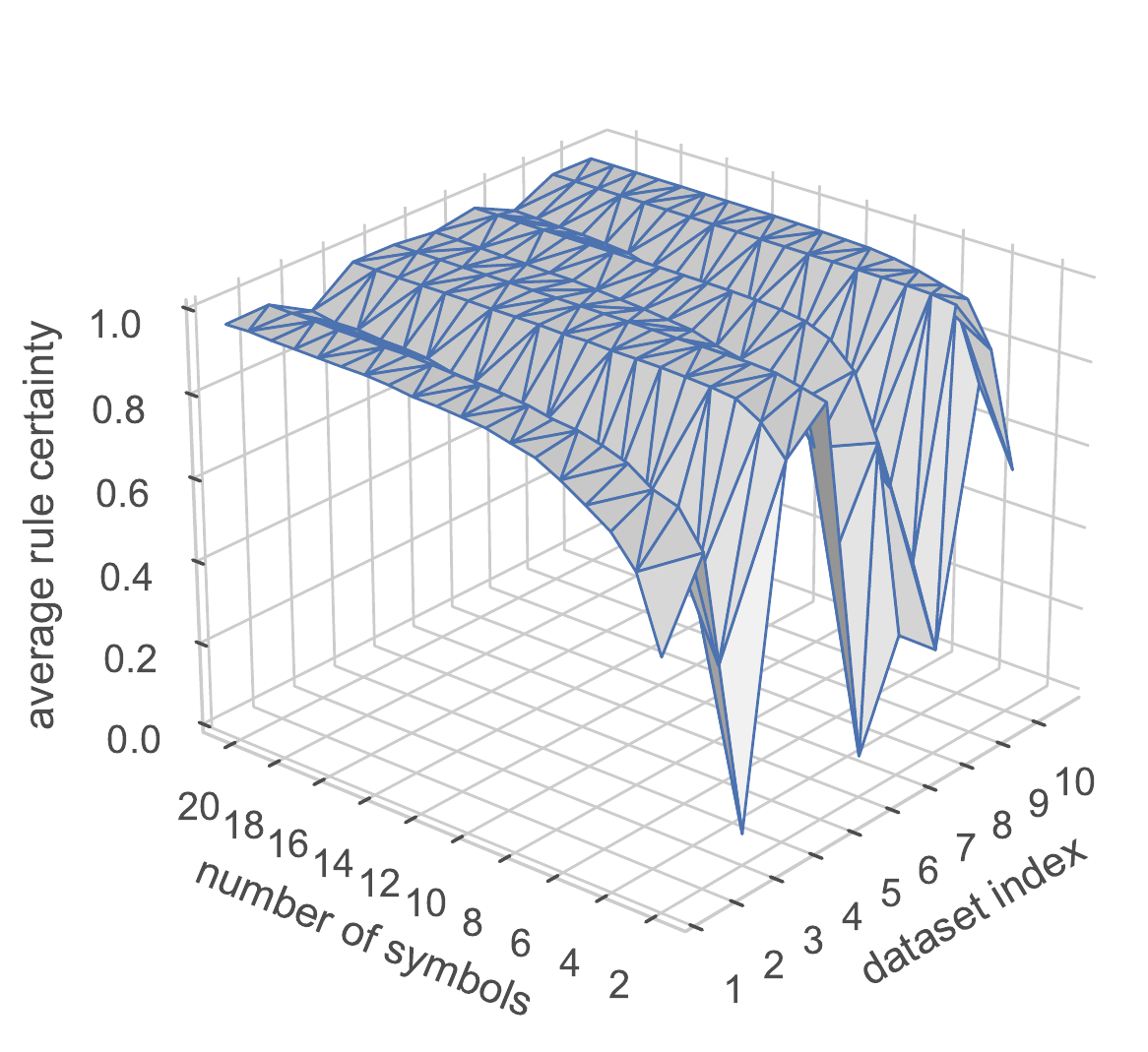}
	\caption{Łukasiewicz and Equation \eqref{eq:distance-local1}}
	\end{subfigure}
	\begin{subfigure}{0.49\textwidth}
	\center
	\includegraphics[width=\textwidth]{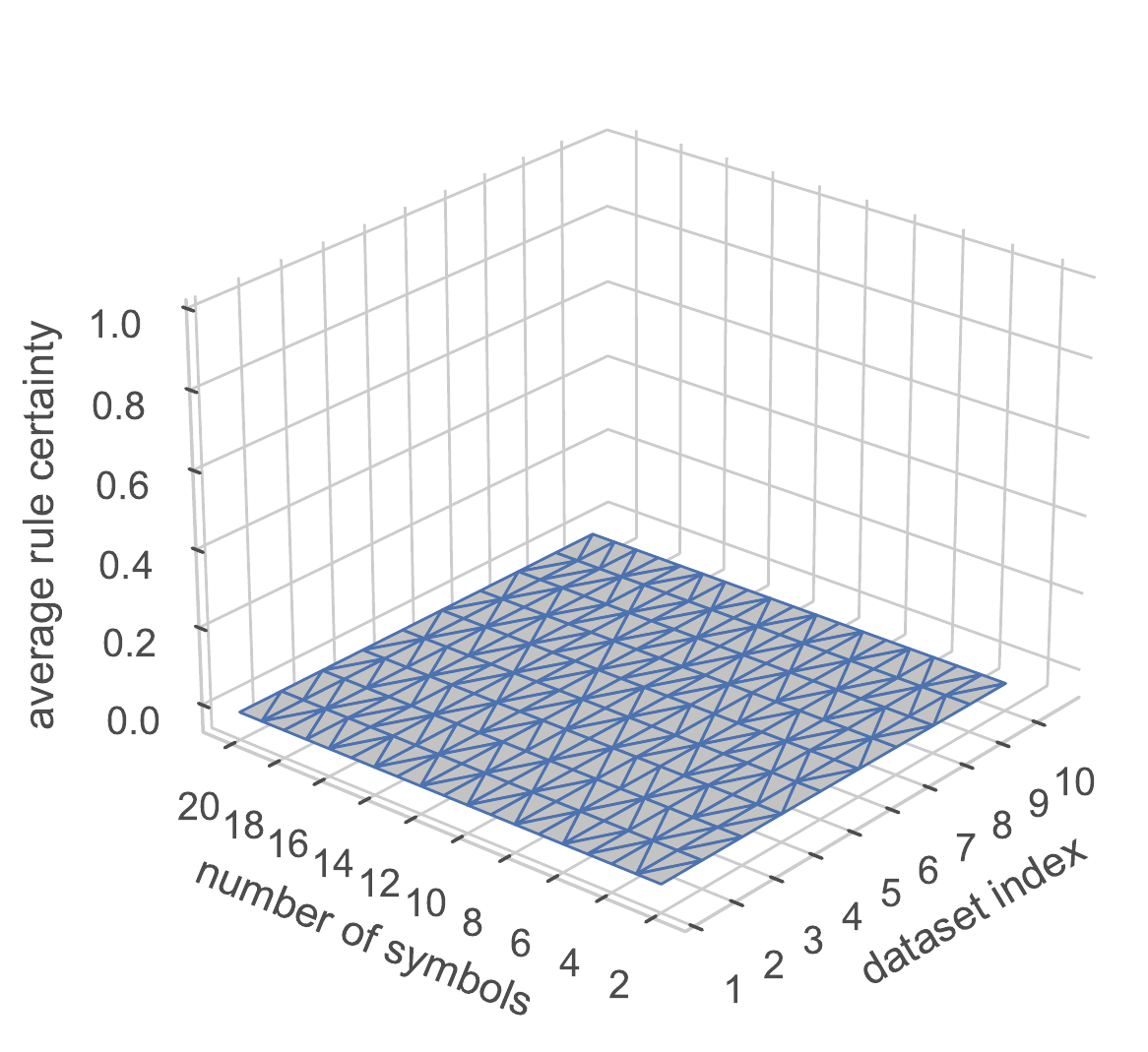}
	\caption{G\"odel and Equation \eqref{eq:distance-local2}}
	\end{subfigure}
	\begin{subfigure}{0.49\textwidth}
	\center
	\includegraphics[width=\textwidth]{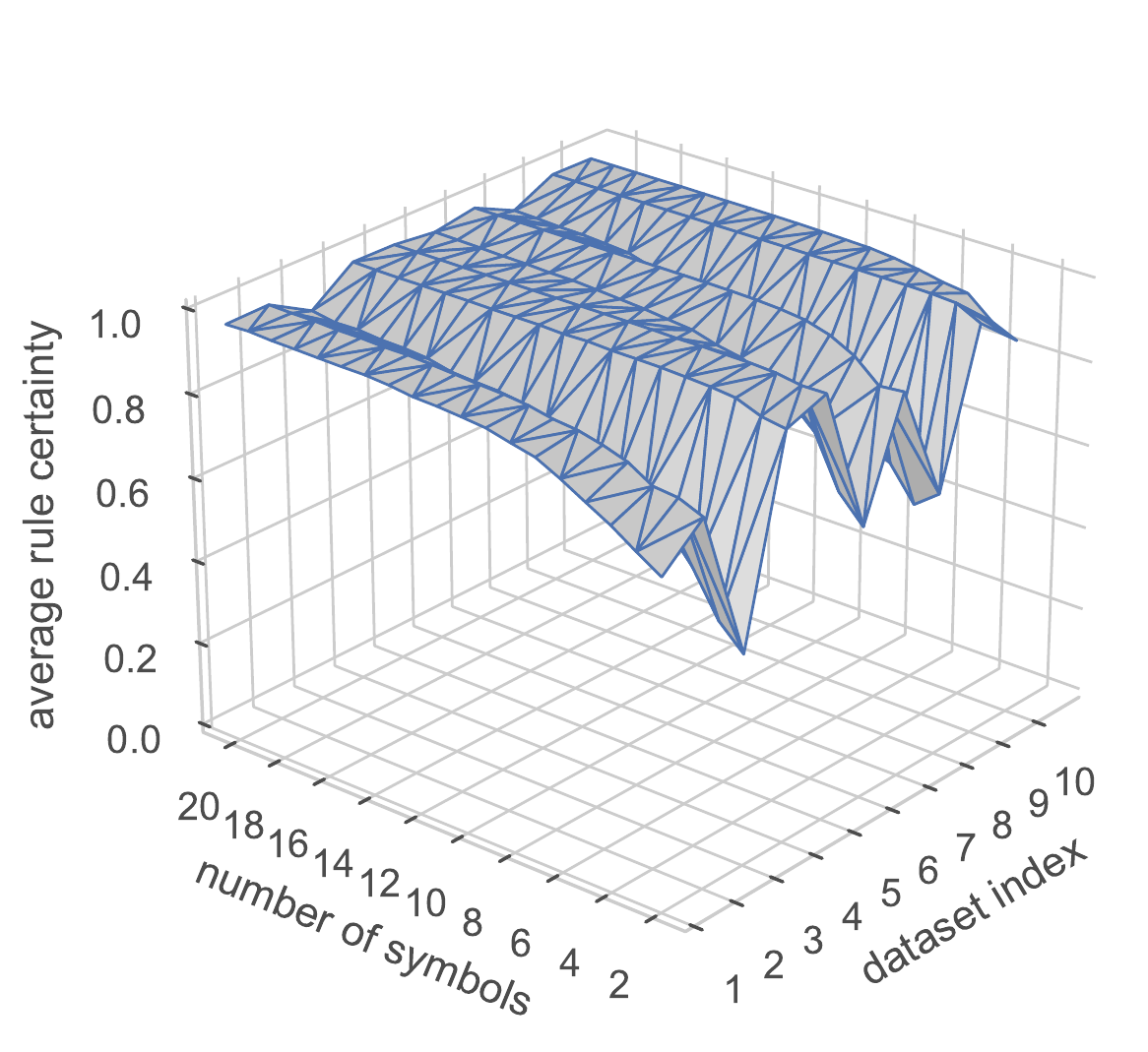}
	\caption{Łukasiewicz and Equation \eqref{eq:distance-local2}}
	\end{subfigure}	
	
	\captionsetup{justification=justified}
	\caption{Average rule confidence for each dataset when varying the number of symbols and the fuzzy implication function. G\"odel and Goguen report the same results while the same behavior is observed for Łukasiewicz and Fodor. Hence, only G\"odel and Łukasiewicz are visualized. In these simulations, Random Forest is the black box.}
\label{fig:terms}
\end{figure*}

Figure \ref{fig:terms} shows the average rule confidence for each dataset when varying the number of symbolic terms for different fuzzy implicators in Equation \eqref{eq:confidence} and feature-wise distance functions in Equation \eqref{eq:distance-global}. It can be noticed that G\"odel produces shallow confidence values while Łukasiewicz reports the highest confidence values. Note that only the fuzzy implicators Łukasiewicz and G\"odel are reported in Figure \ref{fig:terms}. The fuzzy implicator Goguen produces the exact same results as G\"odel, while Fodor produces the same results as Łukasiewicz. The results suggest that more symbolic terms do not necessarily lead to higher confidence values. Instead, more symbols will harm interpretability due to the high granularity of the symbolic explanations. Finally, we can conclude that the proposed feature-wise distance function leads to higher confidence values than the baseline function when using less symbolic terms. We used a Random Forest classifier in these simulations as the black box. Additional classifiers such as Logistic Regression, Na\"ive Bayes, Support Vector Machines, Light Gradient Boosting Machines, and $k$-Nearest Neighbors are explored in \ref{sec:appendix}, which simultaneously shows that the explanation module is model-agnostic.

\begin{figure*}[!ht]
\center

    \begin{subfigure}{0.49\textwidth}
	\center
	\includegraphics[width=\textwidth]{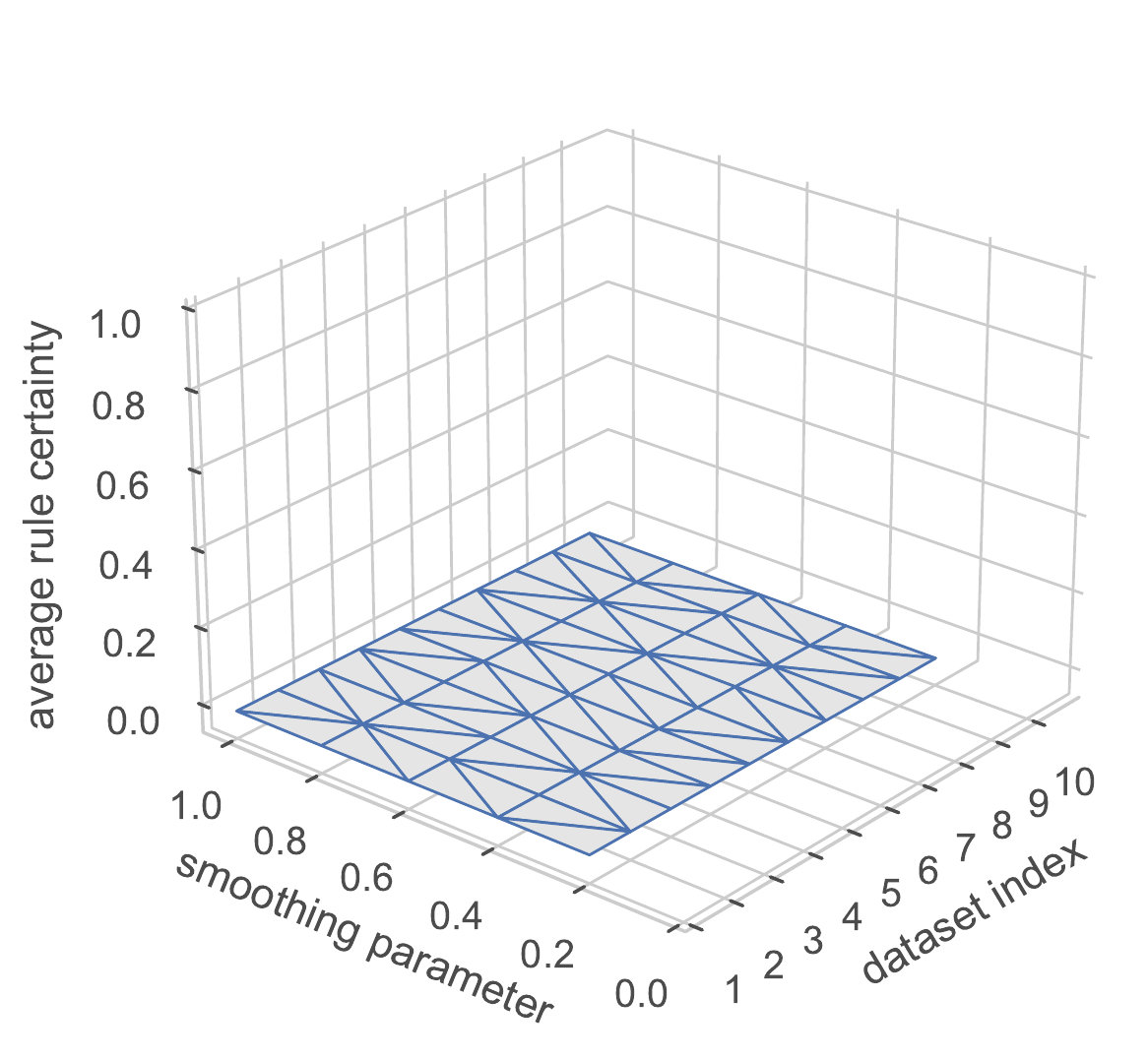}
	\caption{G\"odel and Equation \eqref{eq:distance-local1}}
	\end{subfigure}
	\begin{subfigure}{0.49\textwidth}
	\center
	\includegraphics[width=\textwidth]{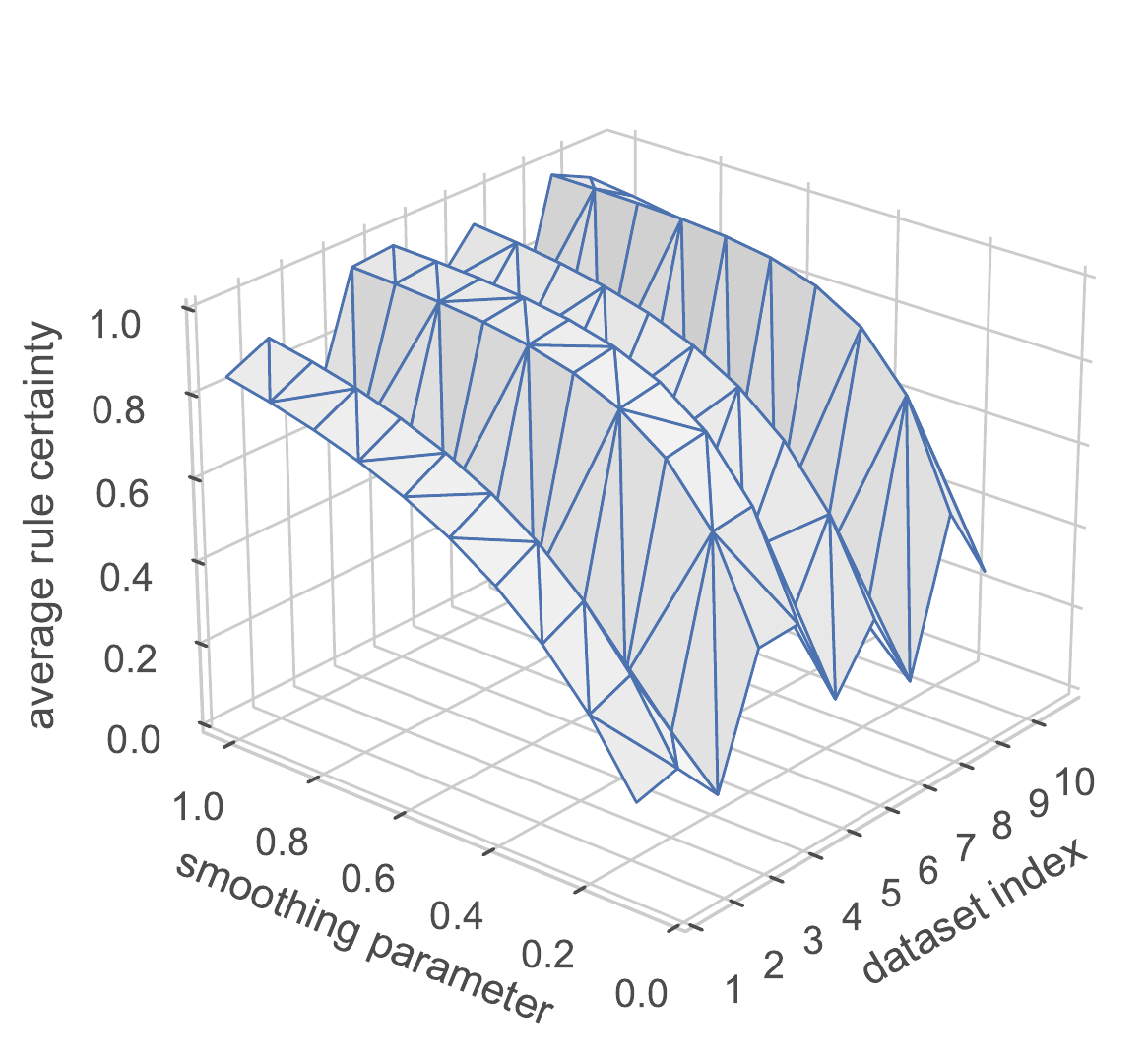}
	\caption{Łukasiewicz and Equation \eqref{eq:distance-local1}}
	\end{subfigure}
	\begin{subfigure}{0.49\textwidth}
	\center
	\includegraphics[width=\textwidth]{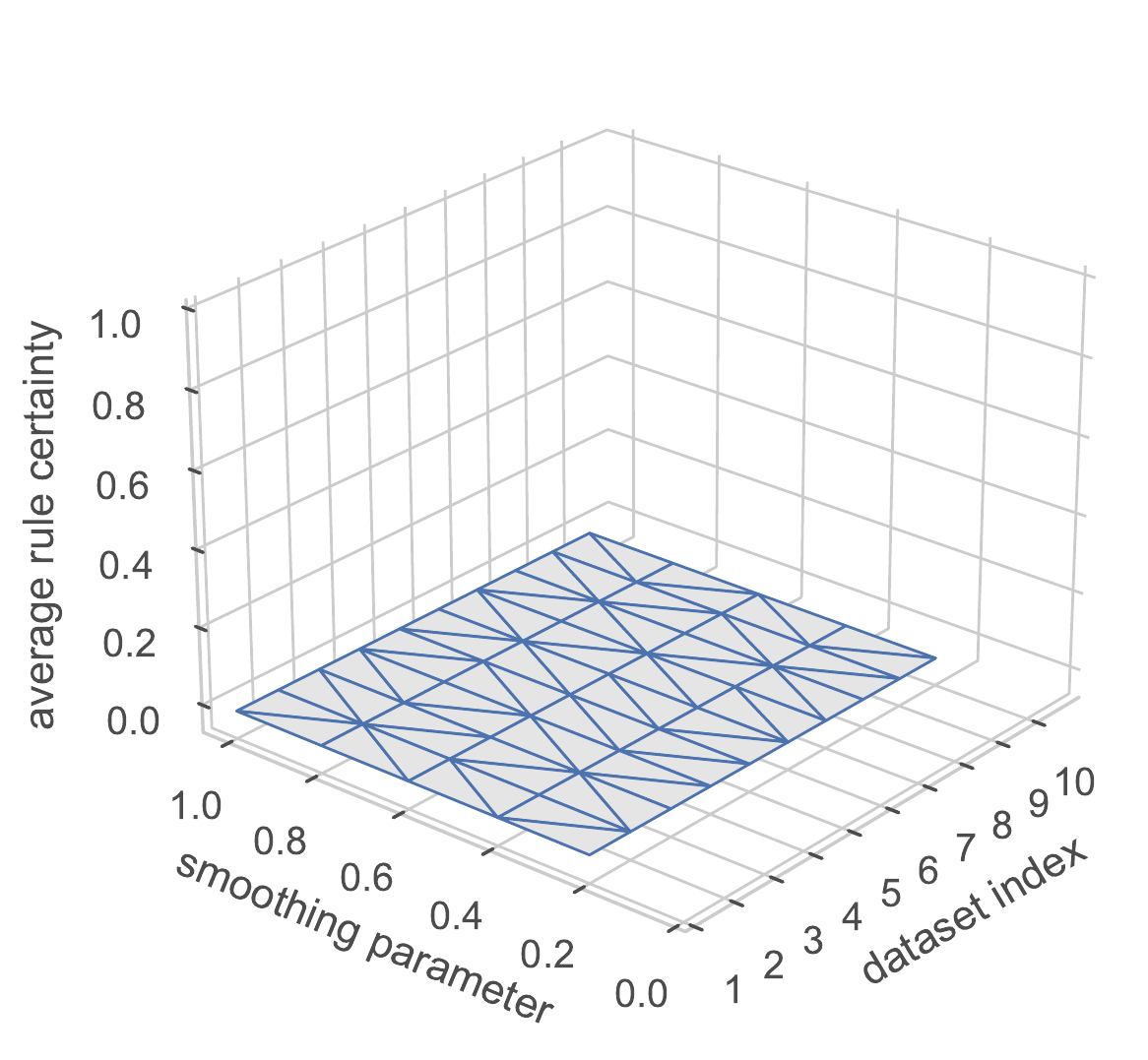}
	\caption{G\"odel and Equation \eqref{eq:distance-local2}}
	\end{subfigure}
	\begin{subfigure}{0.49\textwidth}
	\center
	\includegraphics[width=\textwidth]{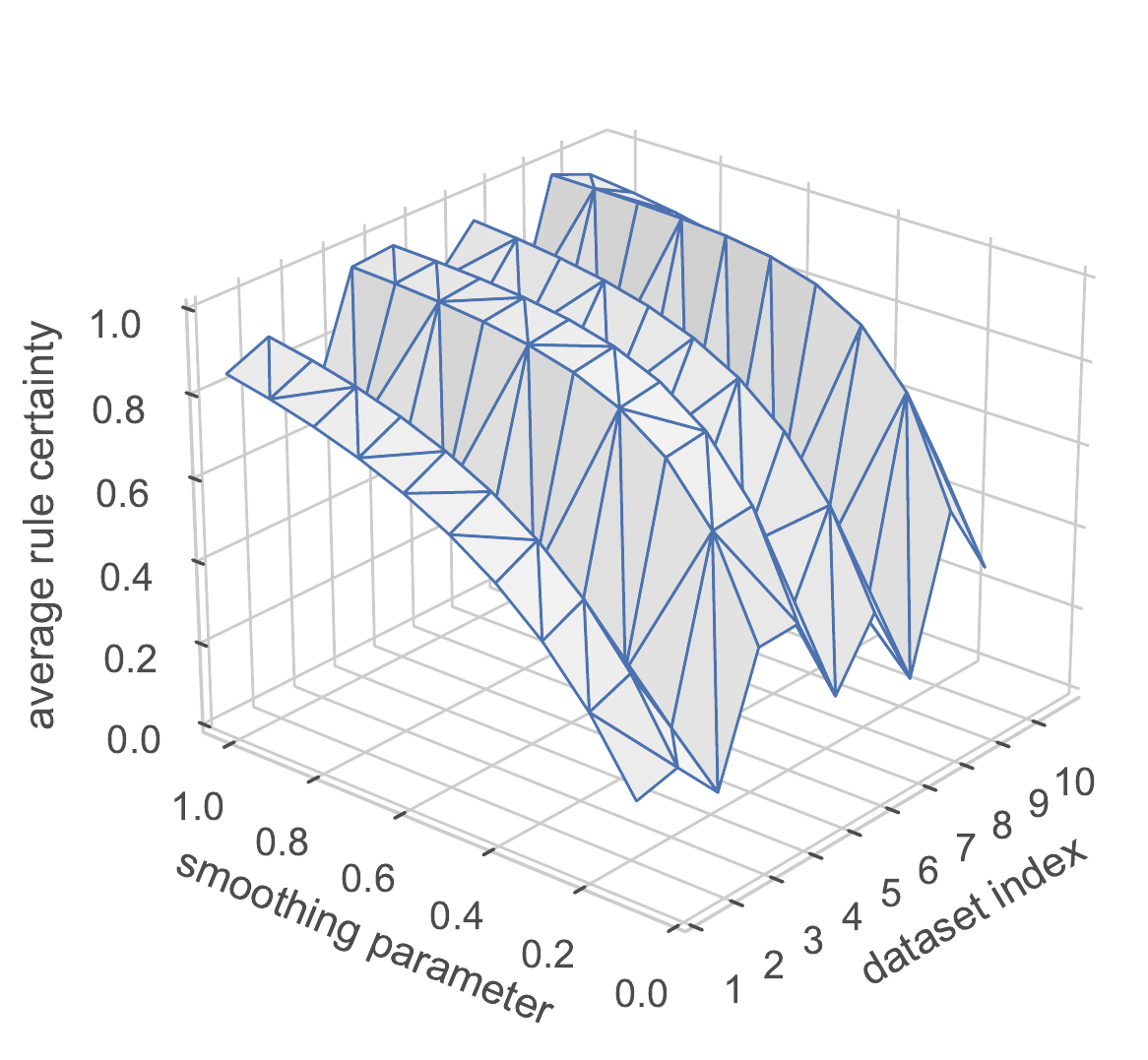}
	\caption{Łukasiewicz and Equation \eqref{eq:distance-local2}}
	\end{subfigure}	
	
	\captionsetup{justification=justified}
	\caption{Average rule confidence for each dataset when varying the smoothing parameter and the fuzzy implication function. G\"odel and Goguen report the same results, while the same behavior is observed for Łukasiewicz and Fodor. Hence, only G\"odel and Łukasiewicz are visualized. In these simulations, Random Forest is the black box.}
\label{fig:smoothing}
\end{figure*}

Figure \ref{fig:smoothing} shows the average rule confidence for each dataset when varying the smoothing parameter $\lambda>0$ in Equation \eqref{eq:fuzzy-relation} for different fuzzy implicators in Equation \eqref{eq:confidence}. Similar to the previous simulation results, Łukasiewicz and Fodor are the best-performing fuzzy operators. Moreover, it can be concluded that the larger smoothing parameter values lead to larger confidence values since the symbolic rules become easier to separate.

Figure \ref{fig:rule-confidence} shows the average antecedent confidence and the average rule confidence values for each dataset when changing the number of symbols. These figures also report the spread between the 10th and 90th percentiles. Similar to the previous simulations, the average antecedent confidence was calculated by averaging all confidence values attached to symbolic terms describing the problem features in the knowledge base. The average rule confidence was calculated by averaging the membership value of every rule in the knowledge base to the fuzzy-rough positive region corresponding to the decision class produced by the classifier. 

It can be seen that, for most datasets, after a certain amount of clusters, usually from 5 to 10, the confidence values have reached a high value and do not increase significantly. At this point, a higher granularity would not add value to the confidence values while harming interpretability. Therefore, such analysis can be used to derive a suitable trade-off between rule confidence and interpretability.

\begin{figure*}[!b]
\center
    \begin{subfigure}{0.49\textwidth}
	\center
    \includegraphics[width=0.95\textwidth]{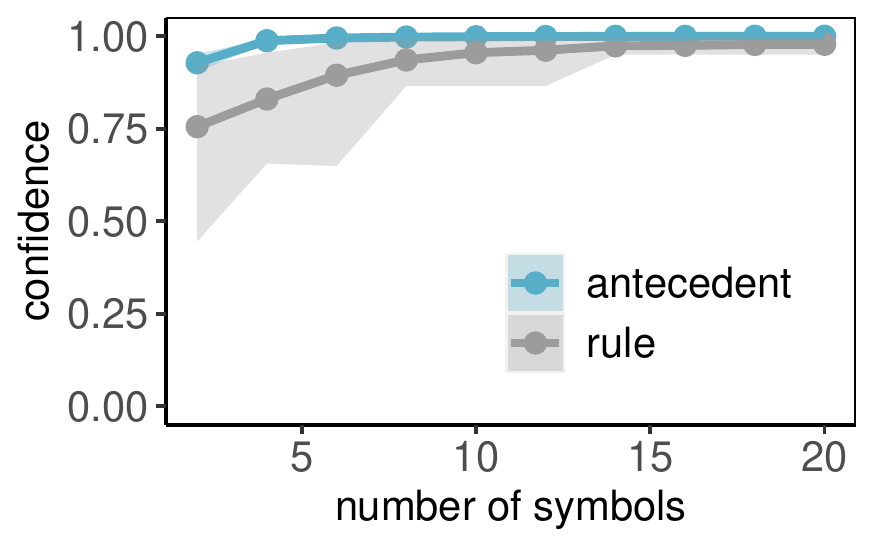}
	\caption{statlog}
	\end{subfigure}
	\begin{subfigure}{0.49\textwidth}
	\center
	\includegraphics[width=0.95\textwidth]{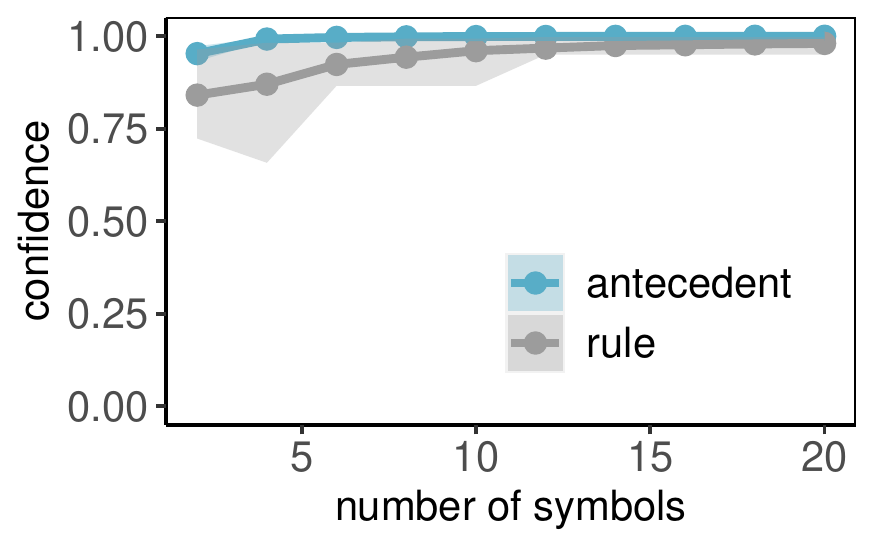}
	\caption{crx}
	\end{subfigure}
	\begin{subfigure}{0.49\textwidth}
	\center
	\includegraphics[width=0.95\textwidth]{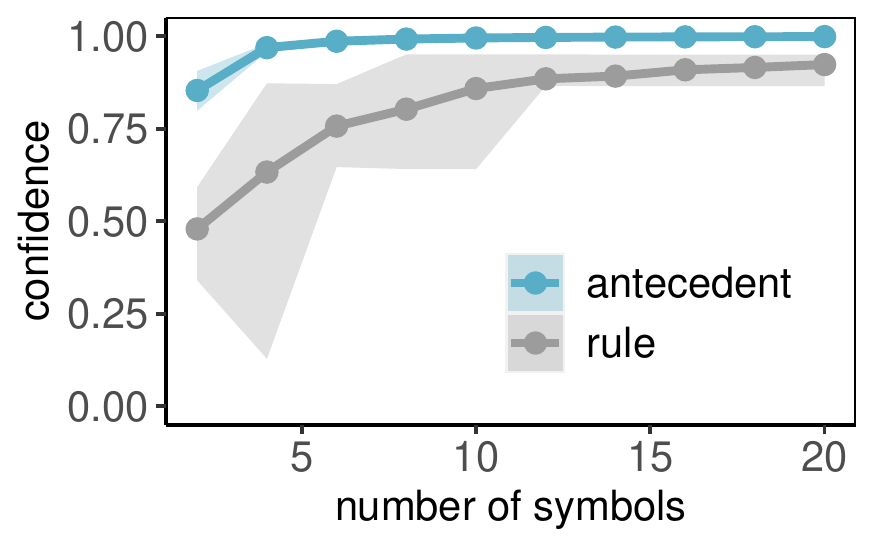}
	\caption{ecoli}
	\end{subfigure}
	\begin{subfigure}{0.49\textwidth}
	\center
	\includegraphics[width=0.95\textwidth]{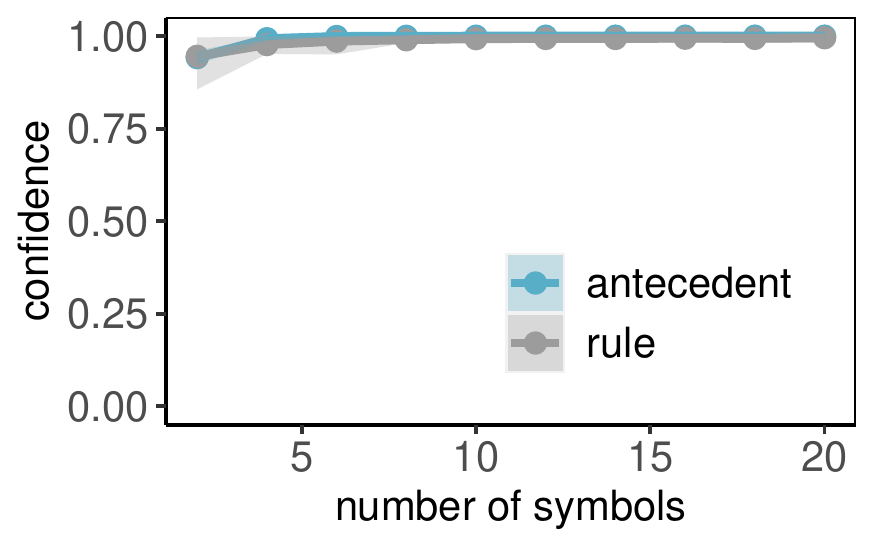}
	\caption{flags}
	\end{subfigure}
    \begin{subfigure}{0.49\textwidth}
    \center
    \includegraphics[width=0.95\textwidth]{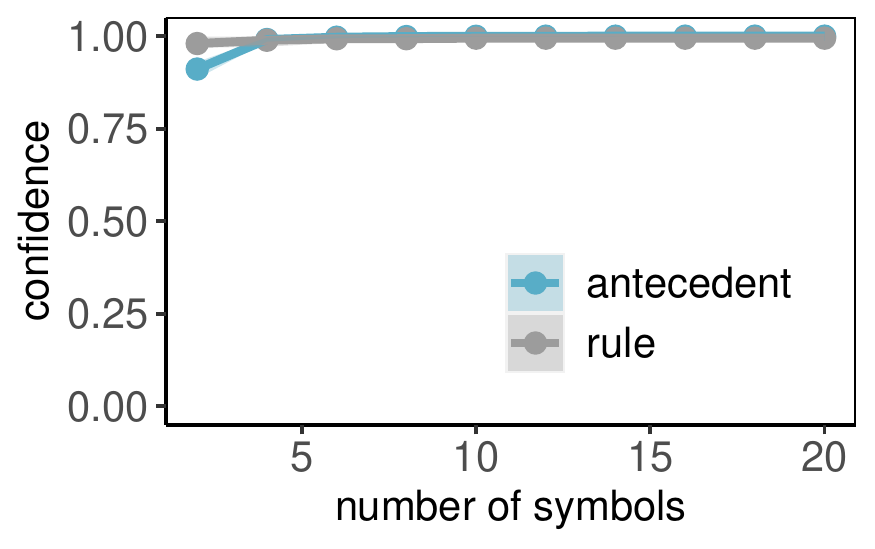}
    \caption{labor}
    \end{subfigure}
    \begin{subfigure}{0.49\textwidth}
    \center
    \includegraphics[width=0.95\textwidth]{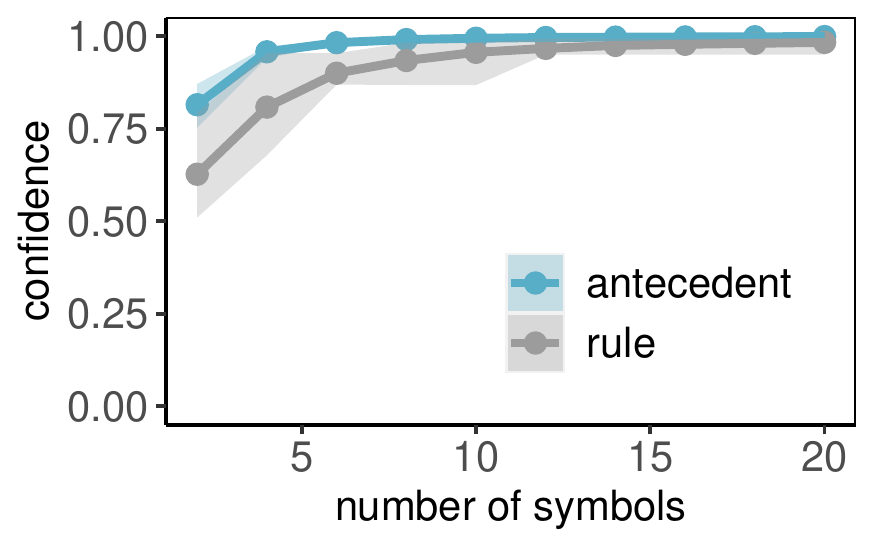}
    \caption{diabetes}
    \end{subfigure}
	\captionsetup{justification=justified}
	\caption{Average rule and antecedent confidence values for each dataset.}
\label{fig:rule-confidence}
\end{figure*}

\begin{figure*}[!ht]\ContinuedFloat
\center
    \begin{subfigure}{0.49\textwidth}
    \center
    \includegraphics[width=0.95\textwidth]{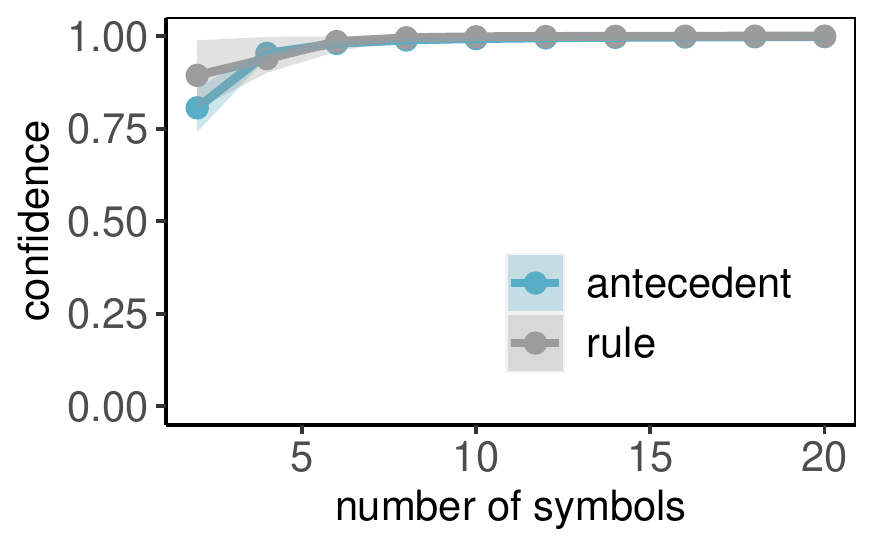}
    \caption{vehicle}
    \end{subfigure}
    \begin{subfigure}{0.49\textwidth}
    \center
    \includegraphics[width=0.95\textwidth]{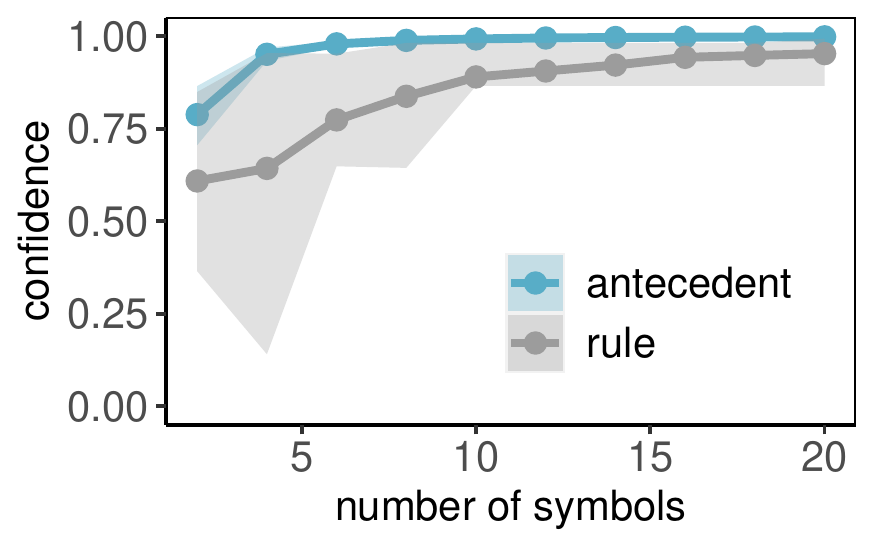}
    \caption{vertebral}
    \end{subfigure}
    \begin{subfigure}{0.49\textwidth}
    \center
    \includegraphics[width=0.95\textwidth]{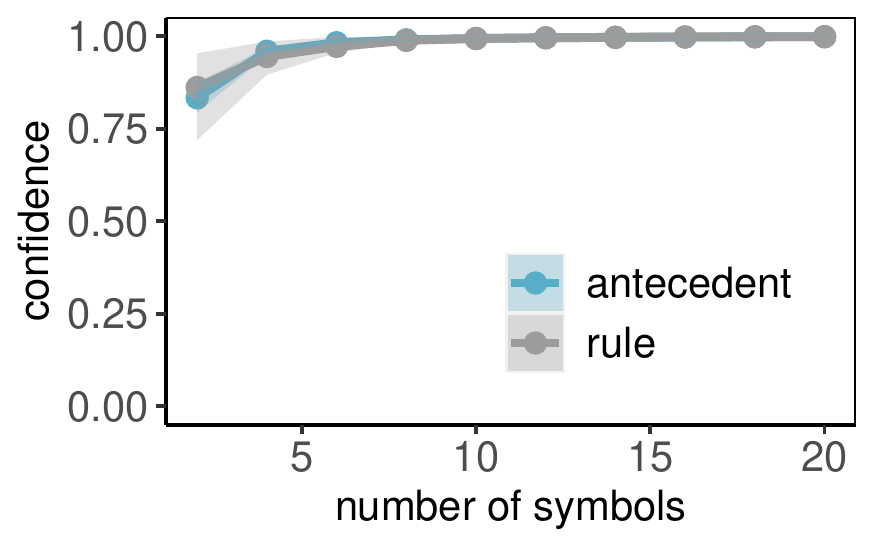}
    \caption{vowel}
    \end{subfigure}
    \begin{subfigure}{0.49\textwidth}
    \center
    \includegraphics[width=0.95\textwidth]{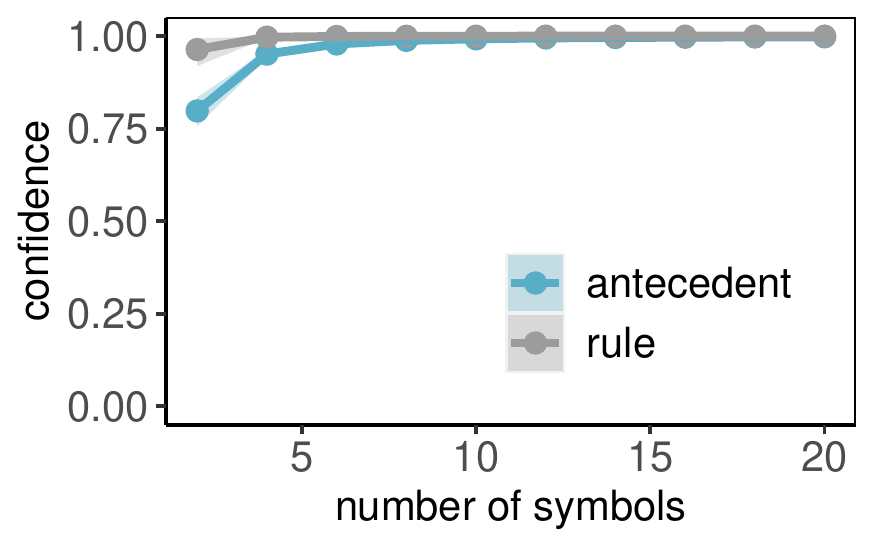}
    \caption{wine}
    \end{subfigure}
	\vspace{-10pt}
	\captionsetup{justification=justified}
	\caption{Average rule and antecedent confidence values for each dataset (cont.).}
\end{figure*}

We configure the parameters associated with the information granulation and uncertainty quantification of rules from the sensitivity analysis. Łukasiewicz and Fodor are the recommended fuzzy implication operators, while the proposed feature-wise distance function in Equation \eqref{eq:distance-local2} is advised when having fewer symbols. In contrast, the smoothing parameter in Equation \eqref{eq:fuzzy-relation} is set to one. We noticed that even larger values might lead to unrealistically high confidence values. This happens because the similarity classes generated by the fuzzy-rough set model under such strict conditions will be trivial (consisting of a single instance). As for the number of linguistic terms, five symbols lead to a reasonable trade-off between the rule certainty and interpretability. Of course, as the number of symbols approaches the infinite, the certainty of the symbolic rules will approach the certainty of the heterogeneous instances.

\subsection{Comparison with other approaches}
\label{sec:simulations:comparison}

To perform an analysis of similar approaches present in the literature, it makes the most sense to execute it on a qualitative level. Issues such as intuitiveness and the ability to provide domain-bound counterfactual explanations are not easily measurable quantitatively. Thus, we would like to draw attention to several features of black-box explanation models in the literature and compare them with the proposed approach. Table~\ref{tab:comparison_sota_theory} gives an overview of essential properties while simultaneously allowing us to compare our proposed method with state-of-the-art methods in terms of the absence/presence of a specific property.

\begin{table}[!ht]{
\caption{Overview of key properties of the proposed model and other state-of-the-art algorithms. The order of appearance of the state-of-the-art algorithms in the table is the same as in the discussion in the literature review section.}
\label{tab:comparison_sota_theory}
\footnotesize{
\begin{tabular}{|c|c|c|c|c|c|c|c|c|c|c|c|}
\hline
\textbf{property} &	\rotatebox{90}{\textbf{this paper}} &	\rotatebox{90}{\textbf{LIME}  \cite{Ribeiro2016}}	& \rotatebox{90}{\textbf{SHAP} \cite{lundberg2017unified}}	& \rotatebox{90}{\textbf{LRP} \cite{Bach2015}}	& \rotatebox{90}{\textbf{NAM} \cite{Agarwal2020}}	& \rotatebox{90}{\textbf{CHIRPS} \cite{Hatwell2020}}	& \rotatebox{90}{\textbf{LORE} \cite{Guidotti2018}}	& \rotatebox{90}{\textbf{MOC} \cite{Dandl2020}}	& \rotatebox{90}{\textbf{MAPLE} \cite{Plumb2018}}	& \rotatebox{90}{\textbf{DICE} \cite{Mothilal2020}}	& \rotatebox{90}{\textbf{FACE} \cite{Poyiadzi2020}}\\ \hline \hline
 fast	&\checkmark	&\checkmark	& 	&\checkmark	& 	&\checkmark&	\checkmark	& 	& 	& 	&\checkmark\\ \hline
deterministic	&\checkmark	& 	&\checkmark	&\checkmark	& 	& 	& 	& 	& 	& 	& \\ \hline
local	&\checkmark	&\checkmark	&\checkmark&	\checkmark	&\checkmark	&\checkmark	&\checkmark	&\checkmark	&\checkmark&	\checkmark	& \\ \hline
model-agnostic	&\checkmark	&\checkmark&	\checkmark	& 	& 	& 	&\checkmark	&\checkmark	& &\checkmark	&\checkmark\\ \hline
suitable for multi-class models&\checkmark&\checkmark	&\checkmark	&\checkmark	&\checkmark	&\checkmark	&	&\checkmark	&\checkmark	&\checkmark	&\checkmark \\\hline
user-parameter-free&&&&&&&\checkmark&&\checkmark&&\\\hline
post-hoc&	\checkmark	&\checkmark	&\checkmark&	\checkmark	& 	&\checkmark&	\checkmark	&\checkmark	&\checkmark	&\checkmark	&\checkmark\\ \hline
provides counterfactual 	&\multirow{2}{*}{\checkmark}	&\multirow{2}{*}{ }	&\multirow{2}{*}{ }	&\multirow{2}{*}{ }	&\multirow{2}{*}{ }	&\multirow{2}{*}{\checkmark}	&\multirow{2}{*}{\checkmark}	&\multirow{2}{*}{\checkmark}	&\multirow{2}{*}{\checkmark}	&\multirow{2}{*}{\checkmark}	&\multirow{2}{*}{\checkmark}\\ 
explanations &&&&&&&&&&&\\\hline
provides symbolic &	\multirow{2}{*}{\checkmark}	&\multirow{2}{*}{ }	&\multirow{2}{*}{ }	&\multirow{2}{*}{ }	&\multirow{2}{*}{ }	&\multirow{2}{*}{ }	&\multirow{2}{*}{\checkmark}	&\multirow{2}{*}{ }	&\multirow{2}{*}{ }	&\multirow{2}{*}{ }	&\multirow{2}{*}{ }\\ 
explanations &&&&&&&&&&&\\\hline
explanations from the domain&\checkmark & & &\checkmark &&\checkmark&&&\checkmark&\checkmark&\\\hline
certainty quantification&\checkmark&&&&\checkmark&\checkmark&&&\checkmark&&\\\hline\hline
\end{tabular}
}}
\end{table}

The first two properties (fast and deterministic) refer mainly to the manner of method execution. A typical reason why a method was assumed to be slow is when it works in a way that it generates a vast number of candidate explanations/candidate solutions and then performs some computationally expensive optimization procedure that aims at slimming the initial model composition. This optimization procedure is often heuristic, directly implying the lack of determinism. Being at the same time fast and deterministic is a rare combination of properties. We want to emphasize that the proposed approach satisfies both of these conditions.

The following two properties, local and model-agnostic, relate to the applicability of an explanation algorithm. Local models aim to describe feature importance for a particular prediction. All but one of the selected approaches operate in a local mode. We have focused on this group as the proposed method falls into it. Notably, the proposed method uses all domain instances in the knowledge base when it computes a local explanation. Being model-agnostic means that a given method does not need a specific internal representation of the black box it explains. An often-met restriction is that some explanation algorithms only work for tree-based or neural models. For instance, NAM \cite{Agarwal2020} works with a neural network, while CHIRPS \cite{Hatwell2020} is dedicated to the Random Forest model. The proposed approach is model-agnostic, which broadens its applicability.

Being user-parameter-free, suitable for multi-class classifiers, and post-hoc relates to how an explaining module was designed to operate. Not all methods are suitable for multi-class classifiers, for example, LORE \cite{Guidotti2018}, which otherwise shares many other features with the proposed method (being user-parameter-free means that the entire procedure does not involve any user-set parameters). Most existing methods, including ours, need some parametrization. Post-hoc relates to the moment at which the explanation model is launched. With one exception, all mentioned methods, ours including, are executed after their black box was trained.

We have mentioned several features related to explanation generation; the first is whether or not a model provides counterfactual explanations. There is a slim group of methods that do so. An even narrower group of methods generates symbolic explanations, enabling more user-friendly communication of the mined knowledge. The proposed explanation module satisfies these conditions, which we believe are critical advantages in contrast to what is already available in the literature. Furthermore, we would like to mention that it's a relatively rare property that an algorithm provides counterfactual explanations for particular cases from the domain. Instead, some methods use synthetically generated instances that may or may not occur in the domain. Lastly, we should mention that some methods, including ours, provide additional tools for certainty quantification for generated explanations, making it more convenient for the end-user.

\subsection{Detailed case study}
\label{sec:simulations:results}

Next, we will illustrate the different questions supported by the proposed conversational agent using the \textit{diabetes} dataset. The structure of these questions and their categories are introduced in Section \ref{sec:queries}.

First, the user should ask the chatbot to load the desired dataset, for example, by typing ``Load the diabetes dataset''. The recognized intent is ``load\_data''. Rasa extracts the dataset to be loaded as the entity ``dataset''. Once the dataset is loaded, the user can request further information, such as the features describing the problem and the number of instances. Figure \ref{fig:data_stats} shows an example of this step where ``data\_stats'' is the intent recognized by the conversational agent.

\begin{figure}[!ht]
    \centering
    \begin{rightbubbles}
        Tell me more about the data.
     \end{rightbubbles}
     
    \begin{leftbubbles}
        The dataset contains the following variables: Plas, Pres, Skin, Insu, Mass, Pedi, and Age. The dataset has 768 instances in total.
    \end{leftbubbles}
    \caption{Basic statistics after having loaded a dataset.}
    \label{fig:data_stats}
\end{figure}

\begin{figure}[!ht]
    \centering
    \begin{rightbubbles}
        How is BMI distributed?
     \end{rightbubbles}
     
    \begin{leftbubbles}
        \includegraphics[width=0.5\columnwidth]{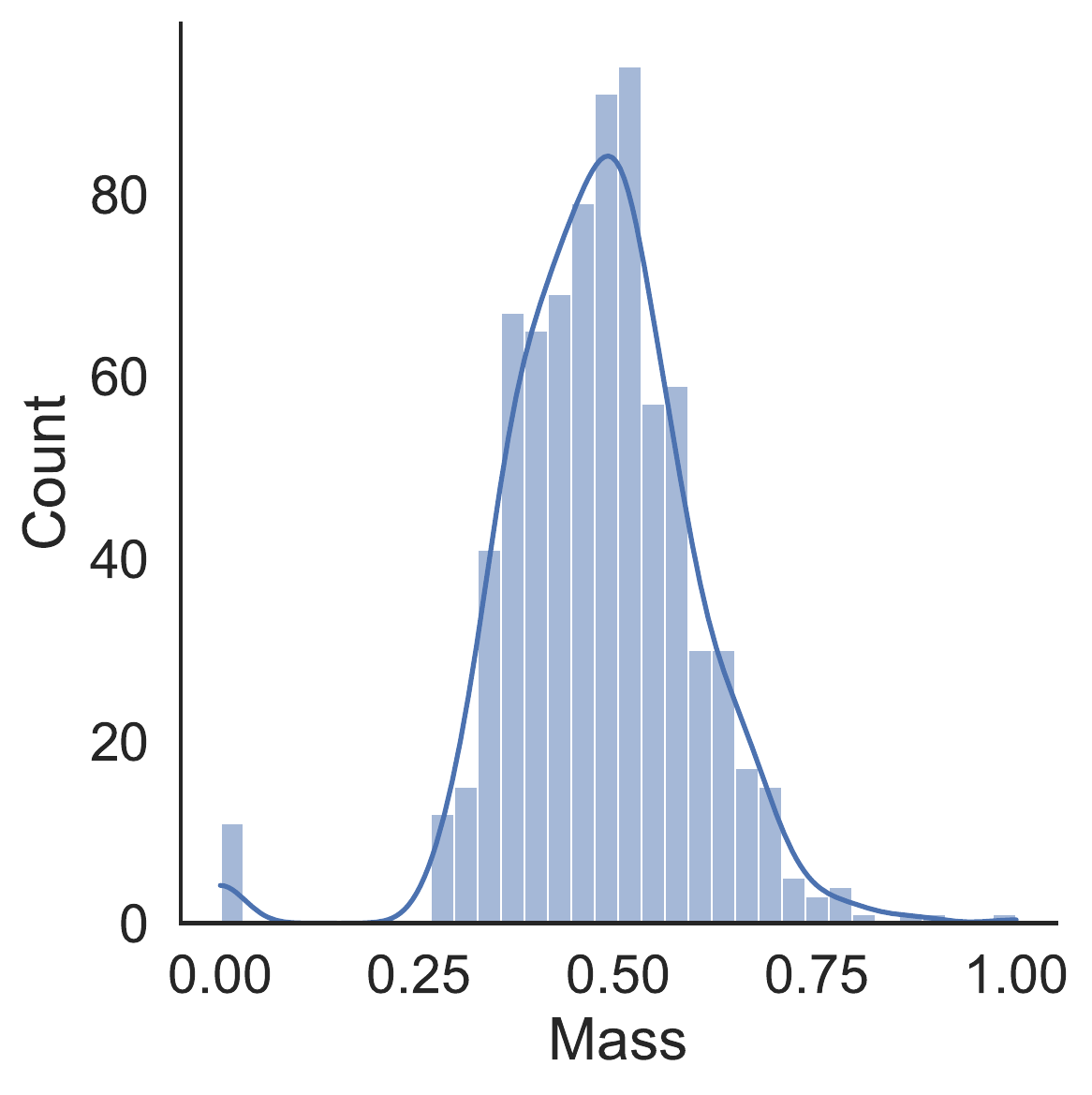}
        
        There you go!
    \end{leftbubbles}
    \caption{An example of a histogram generated by the chatbot.}
    \label{fig:plot_histogram}
\end{figure}

\begin{figure}[!ht]
    \centering
    \begin{rightbubbles}
        How are Age and Pres correlated?
     \end{rightbubbles}
     
    \begin{leftbubbles}
        \includegraphics[width=0.5\columnwidth]{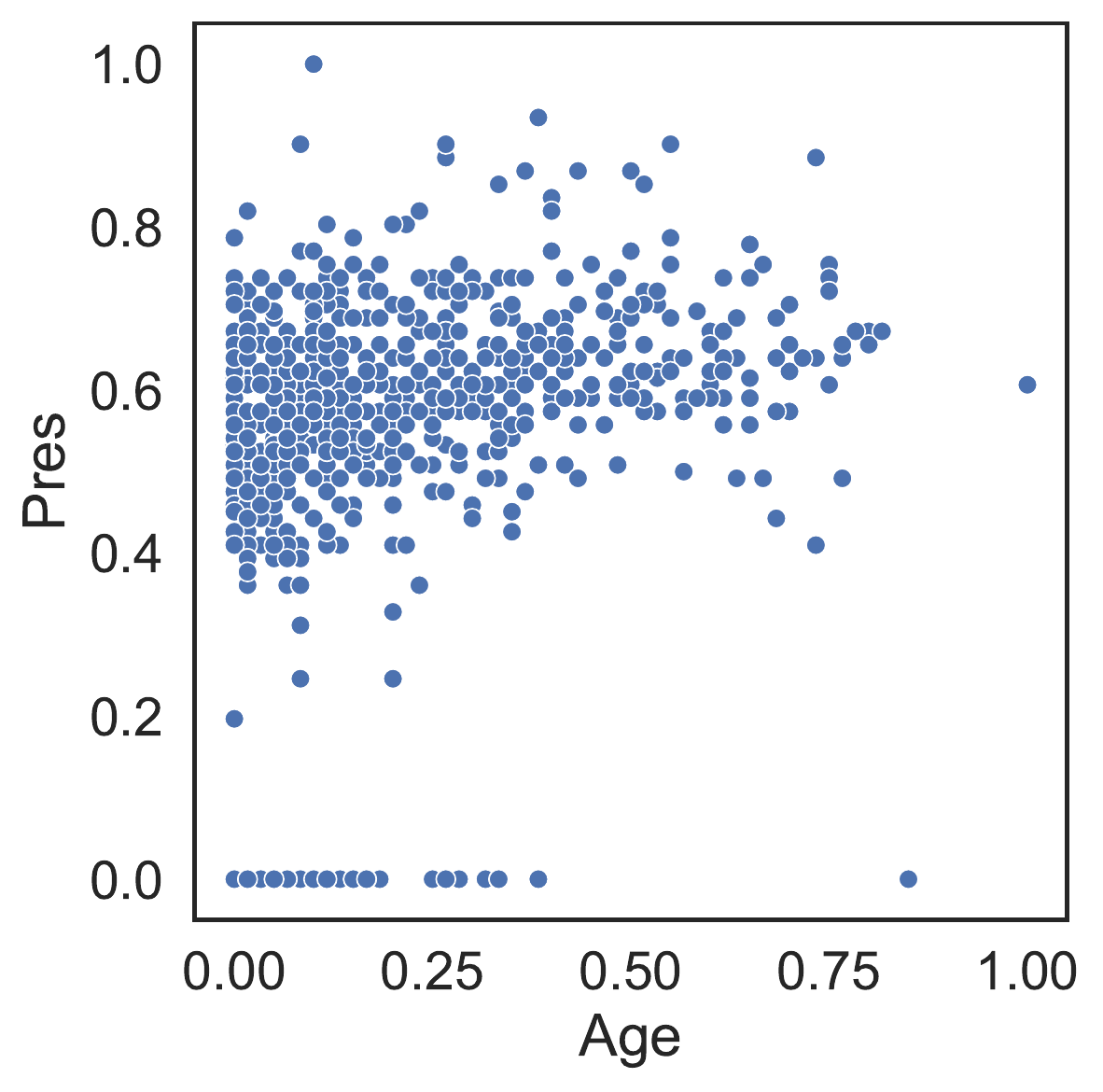} 
        
        The correlation value between the submitted variables is 0.24. The associated p-value is 1.75e-11. The correlation does not seem to be very strong.
    \end{leftbubbles}
    \caption{An example of a pairwise correlation analysis performed by the chatbot.}
    \label{fig:plot_correlation}
\end{figure}

Furthermore, the user can ask questions about problem features, e.g., how a continuous variable is distributed or how two continuous variables are related (using a scatter plot). In the former, the recognized intent is ``plot\_histogram'', and the extracted entity is ``variable''. In the latter, the intent is ``plot\_correlation'', and Rasa extracts twice the entity ``variable''. Examples can be seen in Figure \ref{fig:plot_histogram} and Figure \ref{fig:plot_correlation}.

\begin{figure}[!ht]
    \centering
    \begin{rightbubbles}
        Please show me the correlation matrix.
     \end{rightbubbles}
     
    \begin{leftbubbles}
        \includegraphics[width=0.7\columnwidth]{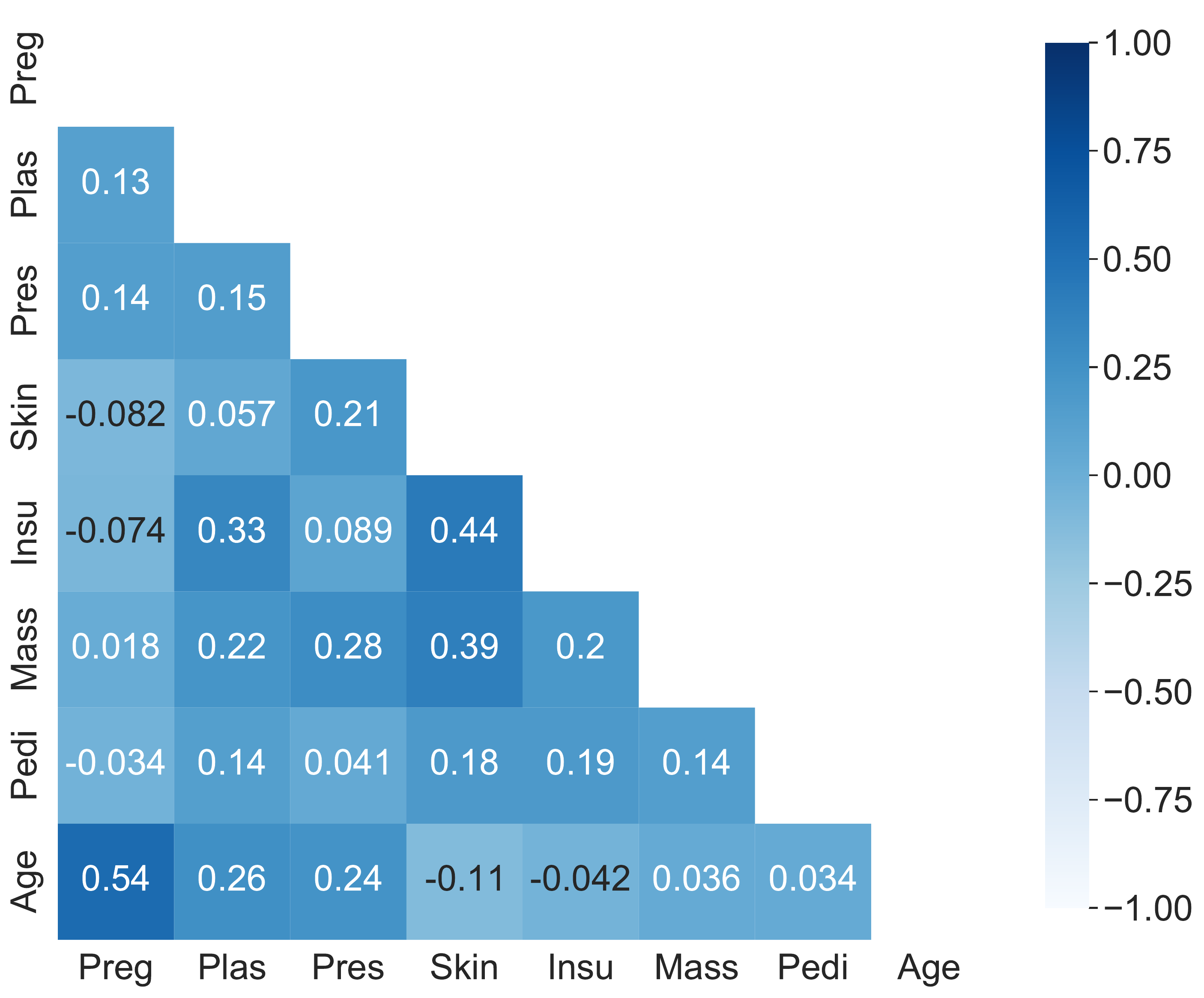} 
    
        This is what the correlation matrix looks like.
    \end{leftbubbles}
    \caption{An example of a correlation matrix generated by the chatbot.}
    \label{fig:correlation_matrix}
\end{figure}

A correlation matrix may also be generated, shown in Figure \ref{fig:correlation_matrix}, provided that the dataset contains numeric features. This is performed when the intent ``correlation\_matrix'' is recognized.

\begin{figure}[!ht]
    \centering
    \begin{rightbubbles}
        Train the model on this data.
     \end{rightbubbles}
     
    \begin{leftbubbles}
        \includegraphics[width=0.50\columnwidth]{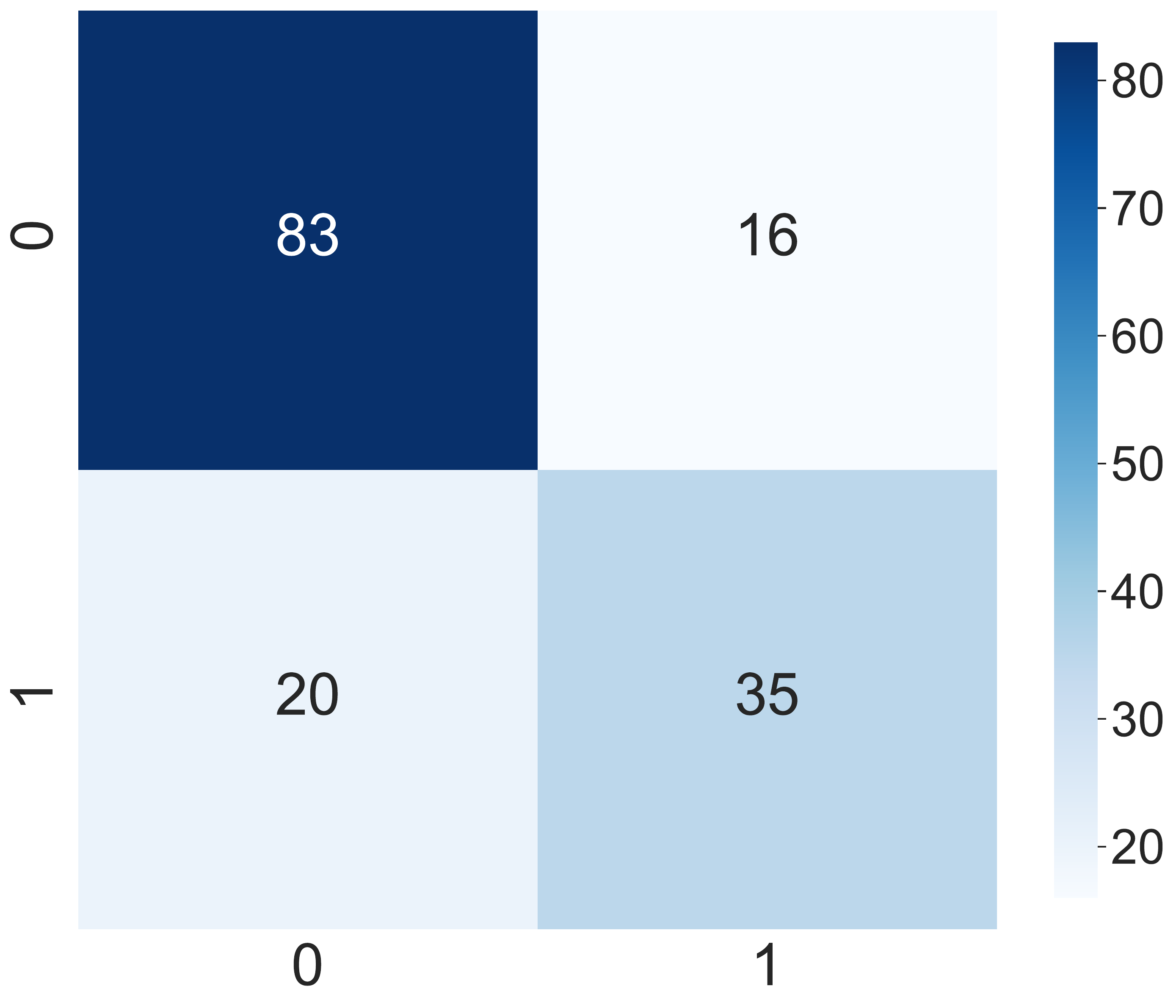}
        
        I successfully trained the Random Forest classifier on the dataset! The accuracy on the test set is 0.766. I included the confusion matrix for the predictions on the test set. If you want higher accuracy, feel free to train the model again with other parameters for the number of estimators and/or maximum depth. If you are satisfied with the achieved accuracy, we can continue!
    \end{leftbubbles}
    \caption{Training of a Random Forest classifier by the chatbot.}
    \label{fig:train_model}
\end{figure}

Random Forest classifier can be trained on the dataset, as the intent ``train\_model'' is recognized. The number of estimators and the maximum depth for the Random Forest can be specified. The chatbot also returns the training accuracy (see Figure \ref{fig:train_model}). It is worth mentioning that including other classifiers is possible and would require little effort.

The user can request more information about the data split used during the training process, as illustrated in Figure \ref{fig:train_test_samples}. The default configuration is 80\% for training and 20\% for testing purposes. Upon recognizing the intent ``train\_test\_samples'', the chatbot returns the train-test split configuration as a percentage and number of instances.

\begin{figure}[!ht]
    \centering
    \begin{rightbubbles}
        How did you split the data?
     \end{rightbubbles}
     
    \begin{leftbubbles}
        I used 80\% of the data (614 instances) for training the classification model. For testing, I used the remaining 20\% of the data (154 instances).
    \end{leftbubbles}
    \caption{An example of requesting the data split in the chatbot.}
    \label{fig:train_test_samples}
\end{figure}

\begin{figure*}[!htbp]
\center

    \begin{subfigure}{0.49\textwidth}
	\center
	\includegraphics[width=\textwidth]{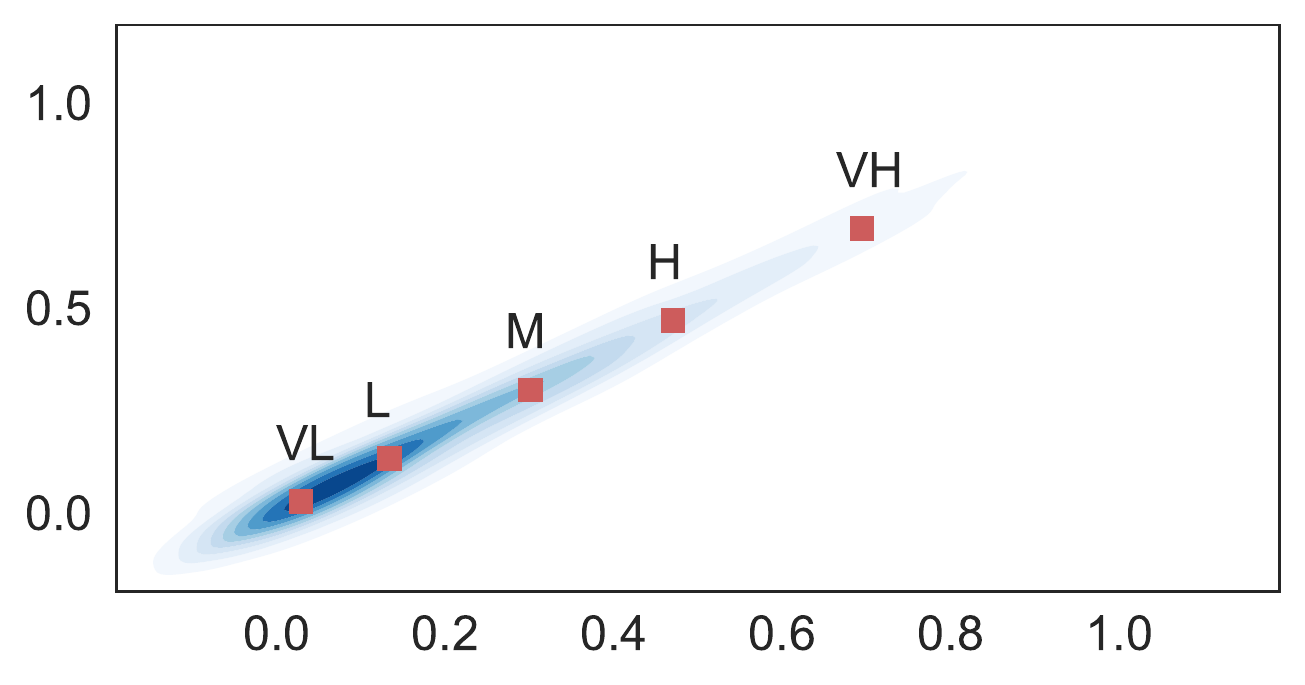}
	\caption{Age}
	\end{subfigure}
	\begin{subfigure}{0.49\textwidth}
	\center
	\includegraphics[width=\textwidth]{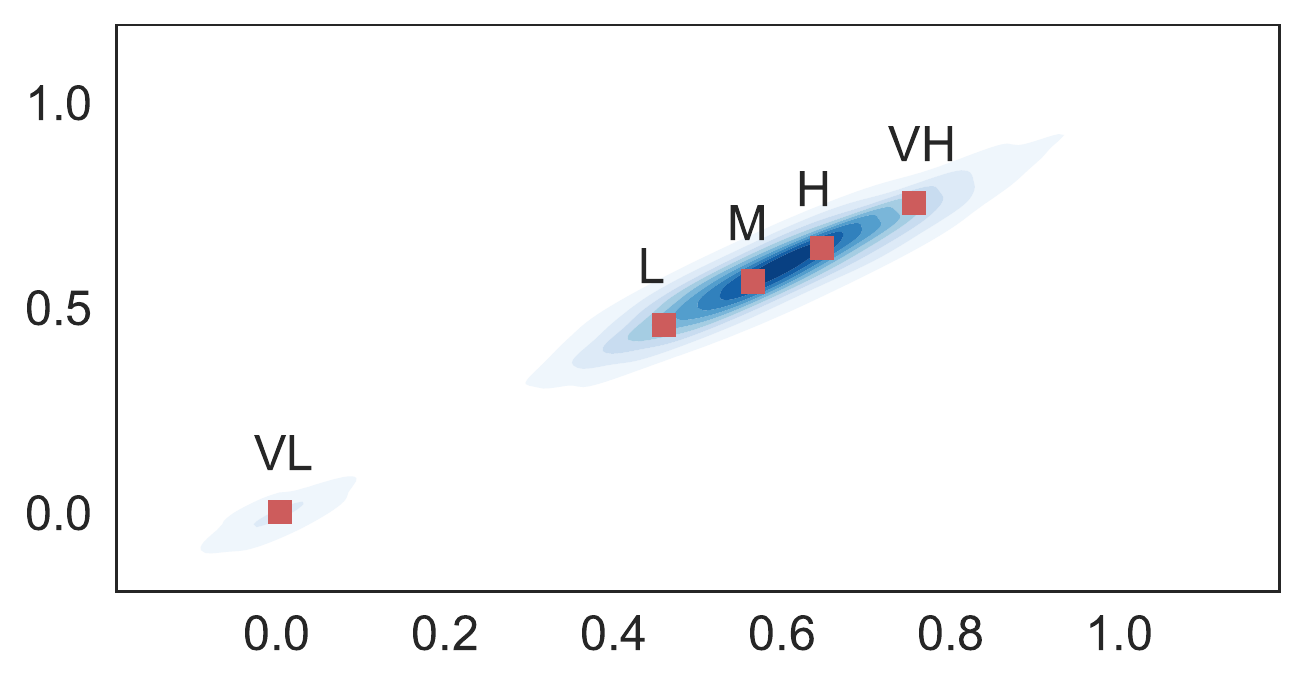}
	\caption{BloodPressure}
	\end{subfigure}
	
	\begin{subfigure}{0.49\textwidth}
	\center
	\includegraphics[width=\textwidth]{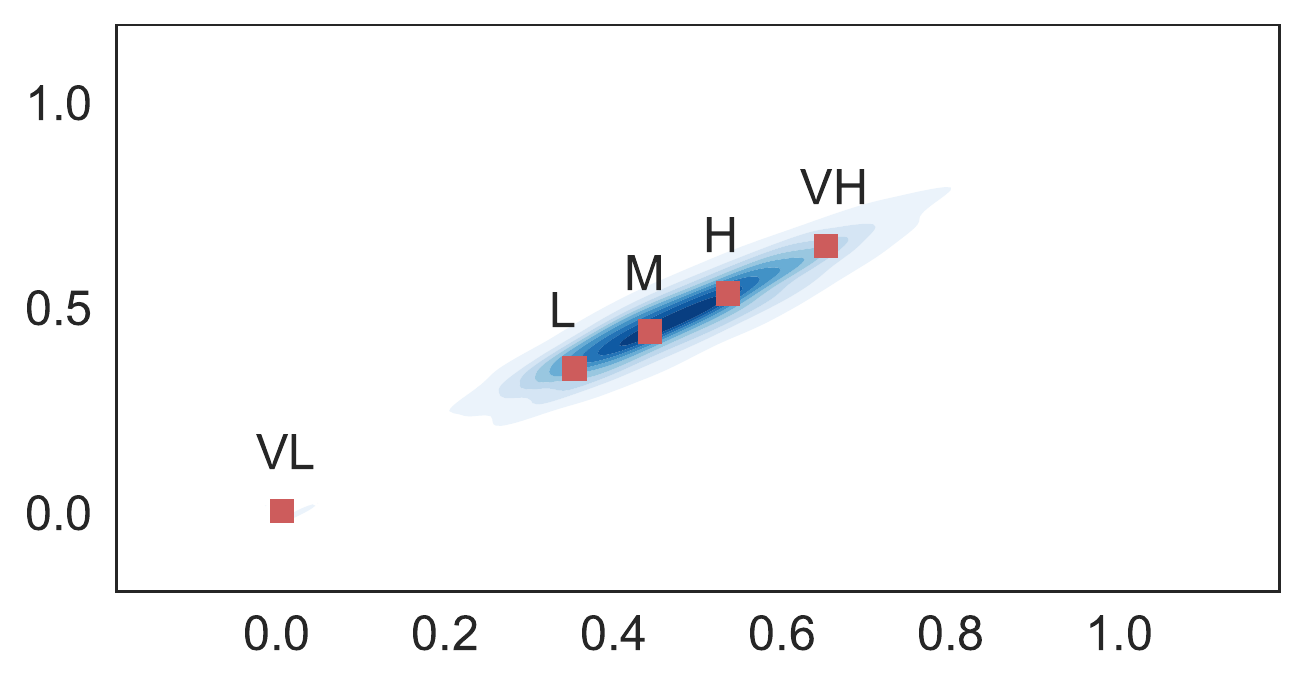}
	\caption{BMI}
	\end{subfigure}
	\begin{subfigure}{0.49\textwidth}
	\center
	\includegraphics[width=\textwidth]{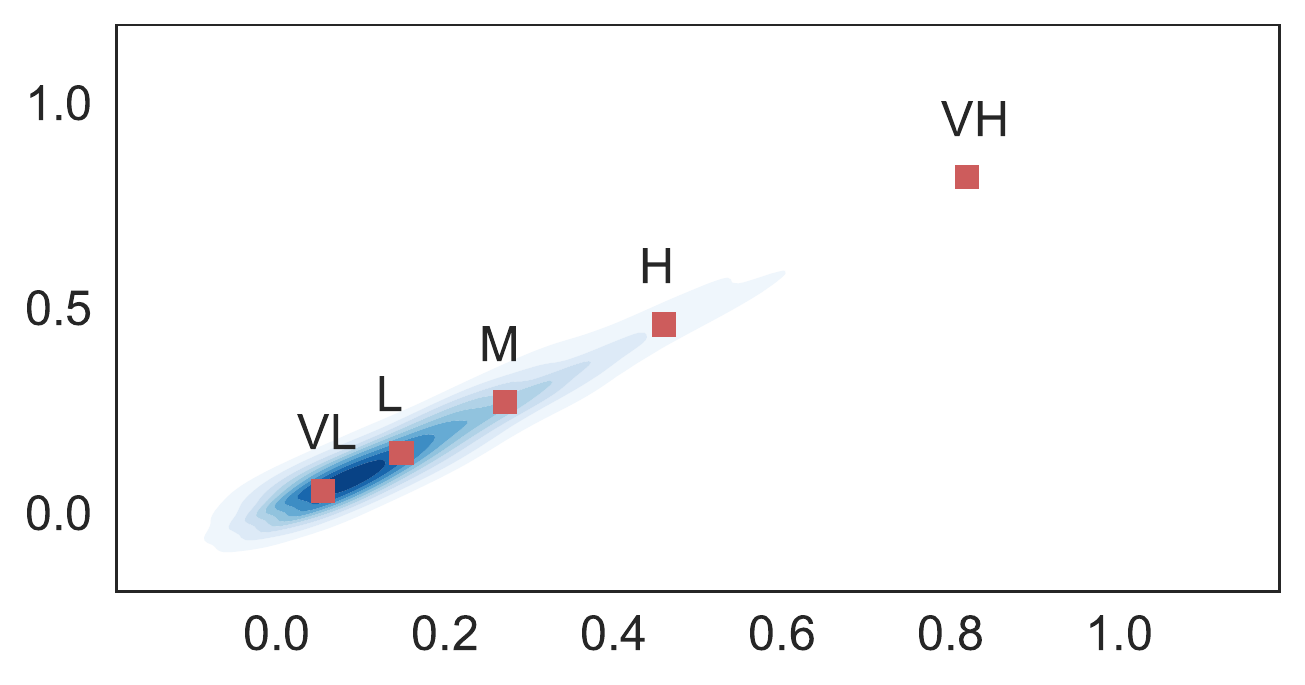}
	\caption{DiabetesPedigreeFunction}
	\end{subfigure}
	
    \begin{subfigure}{0.49\textwidth}
	\center
	\includegraphics[width=\textwidth]{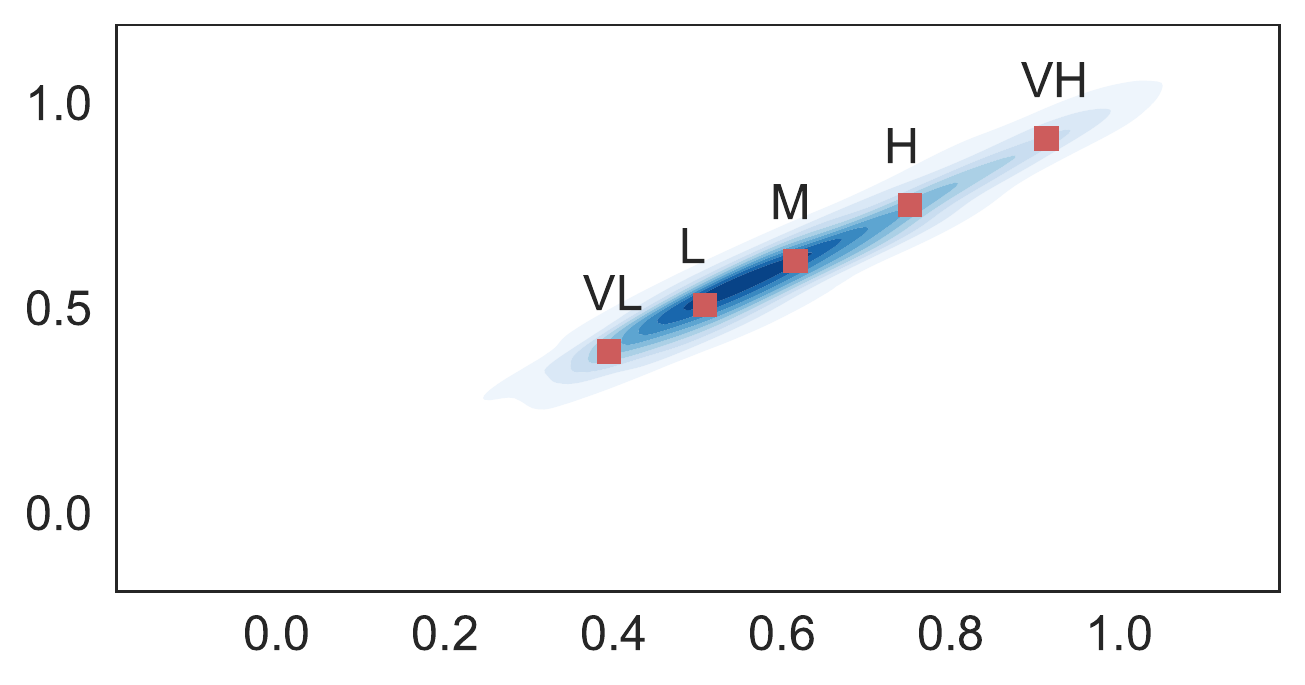}
	\caption{Glucose}
	\end{subfigure}
	\begin{subfigure}{0.49\textwidth}
	\center
	\includegraphics[width=\textwidth]{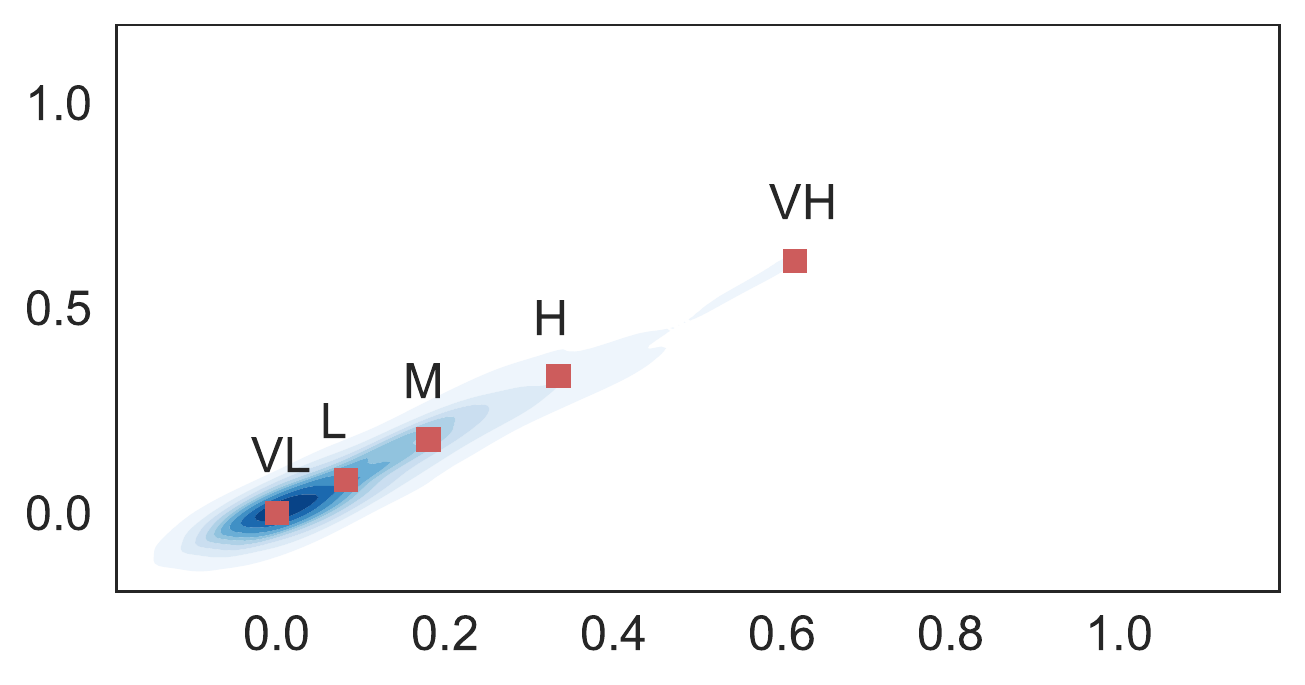}
	\caption{Insulin}
	\end{subfigure}
	
	\begin{subfigure}{0.49\textwidth}
	\center
	\includegraphics[width=\textwidth]{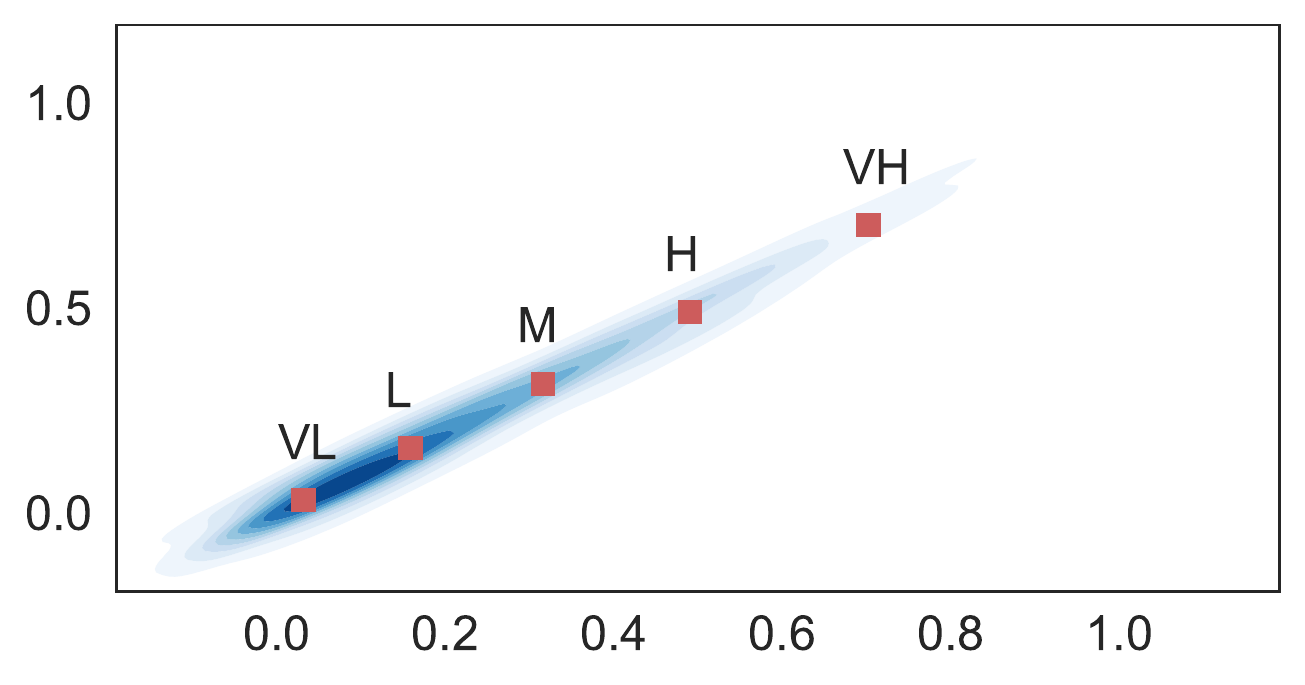}
	\caption{Pregnancies}
	\end{subfigure}
	\begin{subfigure}{0.49\textwidth}
	\center
	\includegraphics[width=\textwidth]{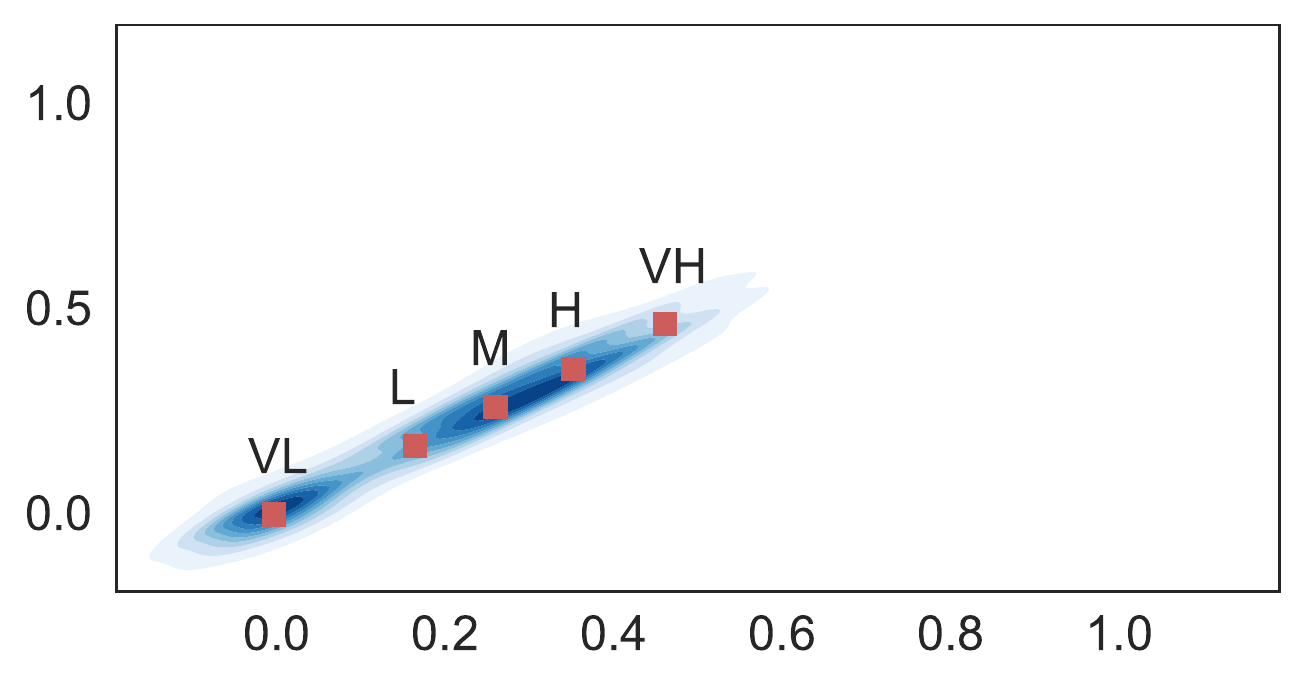}
	\caption{SkinThickness}
	\end{subfigure}	
	
	\captionsetup{justification=justified}
	\caption{Fuzzy prototypes extracted from numerical features describing the diabetes dataset, labeled with predefined linguistic terms.}
\label{fig:terms-pima}
\end{figure*}

\begin{figure}[!ht]
    \centering
    \includegraphics[width=\textwidth]{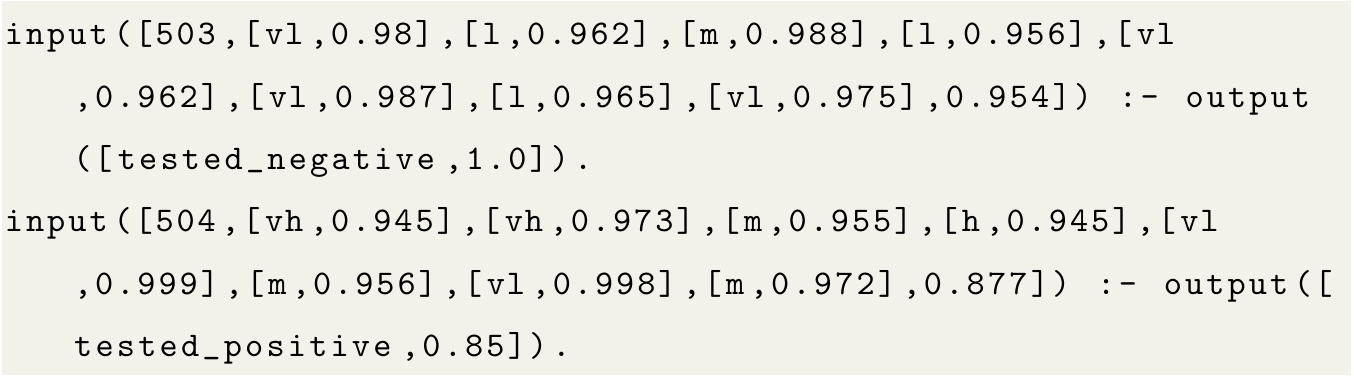}
    \caption{An excerpt from the Prolog knowledge base.}
    \label{fig:kbexcerpt}
\end{figure}

The user can ask the chatbot to construct the symbolic explanation module. As this request is recognized by the intent ``train\_explanation\_module'', the chatbot performs several steps that range from building the symbolic terms to quantifying the uncertainty of symbolic rules. Figure \ref{fig:terms-pima} shows the fuzzy prototypes extracted from numerical features describing the diabetes dataset. The axes denote the normalized values of these features. Thus, prototypes will be located along the identity line. We extract $c=5$ fuzzy prototypes for each numerical feature in this example using the fuzzy $c$-means algorithm. Next, these prototypes are ordered and labeled with a predefined set of symbolic terms (\textit{very low} - VL, \textit{low} - L, \textit{medium} - M, \textit{high} - H, and \textit{very high} - VH) that will be used to create the symbolic fuzzy rules. Figure \ref{fig:symbolic-terms} displays the message shown by the conversational agent. Finally, the fuzzy-rough regions are created, and the Prolog knowledge base is built. Figure \ref{fig:kbexcerpt} shows an excerpt from this knowledge base.

\begin{figure}[!ht]
    \centering
    \begin{rightbubbles}
        Can you construct the symbolic explanation module?
     \end{rightbubbles}
     
    \begin{leftbubbles}
        Done! I calculated the symbolic terms for the values of the input variables. I created fuzzy-rough regions and built a knowledge base in Prolog to run queries. I would appreciate it if you used the correct feature and class names. As a reminder, the feature names are Las, Pres, Skin, Insu, Mass, Pedi, Age, and Class. The decision classes are tested\_negative and tested\_positive.
    \end{leftbubbles}
    \caption{Construction of the symbolic explanation module in the chatbot.}
    \label{fig:symbolic-terms}
\end{figure}

The chatbot can also quantify the problem complexity (see Figure \ref{fig:complexity}). The chatbot computes this measure to complement the average confidence of the symbolic rules in the knowledge base. The intuition is that more difficult problems should involve rules with smaller confidence values.

\begin{figure}[!ht]
    \centering
    \begin{rightbubbles}
        What is the complexity of the loaded problem?
     \end{rightbubbles}
     
    \begin{leftbubbles}
        The complexity of this problem is 0.138. This complexity measure provides evidence of the extent to which the rules in the knowledge base conflict with each other.
    \end{leftbubbles}
    \caption{Complexity of the loaded classification problem.}
    \label{fig:complexity}
\end{figure}

Once the symbolic explanation module is built, the user can pose what-if or counterfactual questions (the latter being discussed later). What-if questions are recognized by the intent ``run\_full\_query''. Rasa extracts the entities ``known\_concept'' for variables specified as known, ``value'' for their respective values, ``outcome'' for the desired outcome, and ``unknown\_concept'' for the variable to be resolved by the chatbot. Figure \ref{fig:run_full_query} shows an example of the answer provided by the conversational agent for a what-if question.

\begin{figure}[!ht]
    \centering
    \begin{rightbubbles}
        If Preg is very low, Gluc is low, and the outcome is test\_negative, what is Age?
     \end{rightbubbles}
     
    \begin{leftbubbles}
        I have run the query for you. These are the results: Age is very low, with a certainty of 0.991. The entire rule has a certainty of 0.954.
    \end{leftbubbles}
    \caption{Example of a what-if question resolved by the chatbot.}
    \label{fig:run_full_query}
\end{figure}

Furthermore, the user can request the closest rule to the one used to resolve the last query. This is recognized by the intent ``closest\_instance''; an example can be seen in Figure \ref{fig:closest_instance}.

\begin{figure}[!ht]
    \centering
    \begin{rightbubbles}
        What rule is closest to this one?
     \end{rightbubbles}
     
    \begin{leftbubbles}
        I found that the following rule is the closest: If Preg is very low, Gluc is low, Pres is medium, Skin is low, Insu is very low, Mass is low, Pedi is very low, Age is very low, then Class is tested\_negative.
    \end{leftbubbles}
    \caption{Closest rule to the last one used to resolve a query.}
    \label{fig:closest_instance}
\end{figure}

As mentioned, the user can also ask the chatbot counterfactual questions, an example of which can be seen in Figure \ref{fig:run_cf_query}. Rasa recognizes such a question by the intent ``run\_cf\_query''. It extracts the variables that should be regarded as unknown as the entity ``unknown\_concept'', the variables that are known as the entity ``known\_concept'', their values as the entity ``value'', and finally, the outcome as the entity ``outcome''.

\begin{figure}[!ht]
    \centering
    \begin{rightbubbles}
        What values should Preg and Gluc take while Age is medium for the outcome to be tested\_negative instead of tested\_positive?
     \end{rightbubbles}
     
    \begin{leftbubbles}
        I have run the query for you. These are the results: Preg is very high, with a certainty of 0.929. Gluc is high with a certainty of 0.924. The rule has a certainty of 0.883.
    \end{leftbubbles}
    \caption{Example of a counterfactual question resolved by the chatbot.}
    \label{fig:run_cf_query}
\end{figure}

The user can also request the fuzzy-rough uncertainty \cite{koumeri2021bias} associated with a protected feature. This measure quantifies the fuzzy-rough regions' change when the protected feature is removed, which can be seen as a proxy for measuring explicit bias. An example can be seen in Figure \ref{fig:bias}. Rasa recognizes this as the intent ``bias'' and extracts the entity ``variable'' which indicates which feature is to be suppressed.

\begin{figure}[!ht]
    \centering
    \begin{rightbubbles}
        What is the explicit bias associated with Age?
     \end{rightbubbles}
     
    \begin{leftbubbles}
        The fuzzy-rough uncertainty (explicit bias) against this feature is 0.662. This measure quantifies the changes in the decision boundaries after removing a protected feature as a proxy for bias quantification.
    \end{leftbubbles}
    \caption{Explicit bias attached to a protected feature.}
    \label{fig:bias}
\end{figure}

Additionally, the user can request the top rules in the knowledge base, as seen in Figure \ref{fig:top_rules_kb}. The chatbot recognizes this task through the intent ``top\_rules\_kb''. These rules provide valuable knowledge about the problem that domain experts can further inspect. 

\begin{figure}[!ht]
    \centering
    \begin{rightbubbles}
        What are the top rules in the knowledge base?
     \end{rightbubbles}
    
    \begin{leftbubbles}
        These are the top 3 rules in the knowledge base: \newline
        Rule \#1: If Preg is medium, Plas is low, Pres is low, Skin is high, Insu is medium, Mass is low, Pedi is high, Age is low, then Class is tested\_negative. \newline
        Rule \#2: If Preg is high, Plas is very high, Pres is very high, Skin is medium, Insu is high, Mass is very high, Pedi is medium, Age is high, then Class is tested\_positive. \newline
        Rule \#3: If Preg is high, Plas is very high, Pres is low, Skin is high, Insu is high, Mass is medium, Pedi is medium, Age is low, then Class is tested\_positive.
    \end{leftbubbles}
    \caption{Top rules in the knowledge base.}
    \label{fig:top_rules_kb}
\end{figure}

Before concluding our paper, we should mention that the proposed conversational agent can be extended. For example, we can integrate popular fairness measures to quantify and mitigate bias, include pre-processing pipelines, or incorporate feature attribution metrics to provide interpretability.

\section{Concluding remarks}
\label{sec:remarks}

In this paper, we presented a reasoning module powered by Prolog to generate symbolic explanations from the predictions made by a black-box classifier. The main procedure of this module concerns the induction of fuzzy rules used to derive the explanations. The main advantages of our proposal are summarized as follows. Firstly, the rules are automatically mined using a data-driven approach that relies on granular computing. Secondly, the explanation module is agnostic and can be coupled with any machine learning classifier. Thirdly, the module computes the confidence of rules as a tool to assess the correctness of explanations. Lastly, users can pose questions using natural language queries and receive understandable answers without involving complex theoretical concepts.

The numerical simulations using several pattern classification problems allowed us to draw interesting conclusions about the parameters involved in fuzzy rule construction. Concerning the number of symbols, we noticed that having more granularity does not always lead to higher confidence values but harms interpretability. It was also seen that the proposed feature-wise distance function yields better results than the baseline when having less granularity, while the smoothing parameter should be as close to one as possible. Concerning the fuzzy operators, Łukasiewicz and Fodor were the best-performing operators in our simulations.

Within the limitations of our proposal, we can mention the trade-off between precision and significance. On the one hand, more precision will harm interpretability since we need more symbols to describe numerical variables. Less precision will improve interpretability but at the expense of harming the confidence of rules. Another limitation is that experts must define the symbols they want to produce semantically coherent explanations. For example, it would be grammatically odd to describe the variable age in terms of ``low age" and ``high age" instead of using the linguistic terms ``young" and ``old," respectively. Future research will be devoted to inferring such linguistic terms from the feature names.

\bibliographystyle{elsarticle-harv}
\bibliography{references}

\appendix
\section{Extended sensitivity analysis}
\label{sec:appendix}

To investigate the differences in the explanation module quality that may arise when using different classifiers as black boxes to be explained, we conduct an extended sensitivity analysis in this appendix. The indicator of the model quality in the case of the proposed approach is the rule confidence, while the parameters to be changed are the fuzzy implication function, the distance function, and the smoothing parameter. The classifiers include Logistic Regression, Na\"ive Bayes, Support Vector Machines, Light Gradient Boosting Machine, and $k$-Nearest Neighbors.

\paragraph{Support Vector Machines}

\begin{figure*}[!htbp]
\center
    \begin{subfigure}{0.49\textwidth}
	\center
	\includegraphics[width=\textwidth]{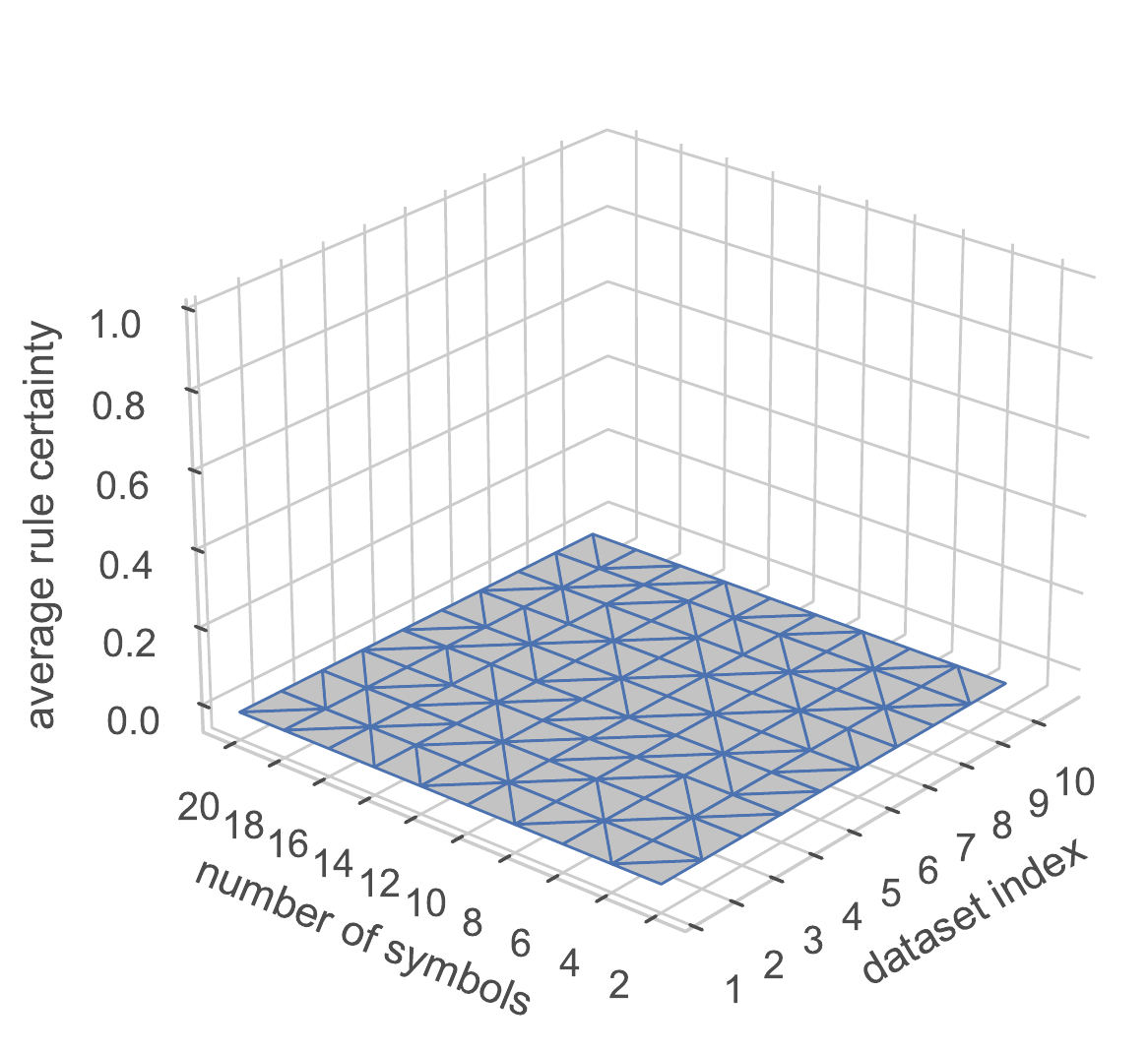}
	\caption{G\"odel and Equation \eqref{eq:distance-local1}}
	\end{subfigure}
	\begin{subfigure}{0.49\textwidth}
	\center
	\includegraphics[width=\textwidth]{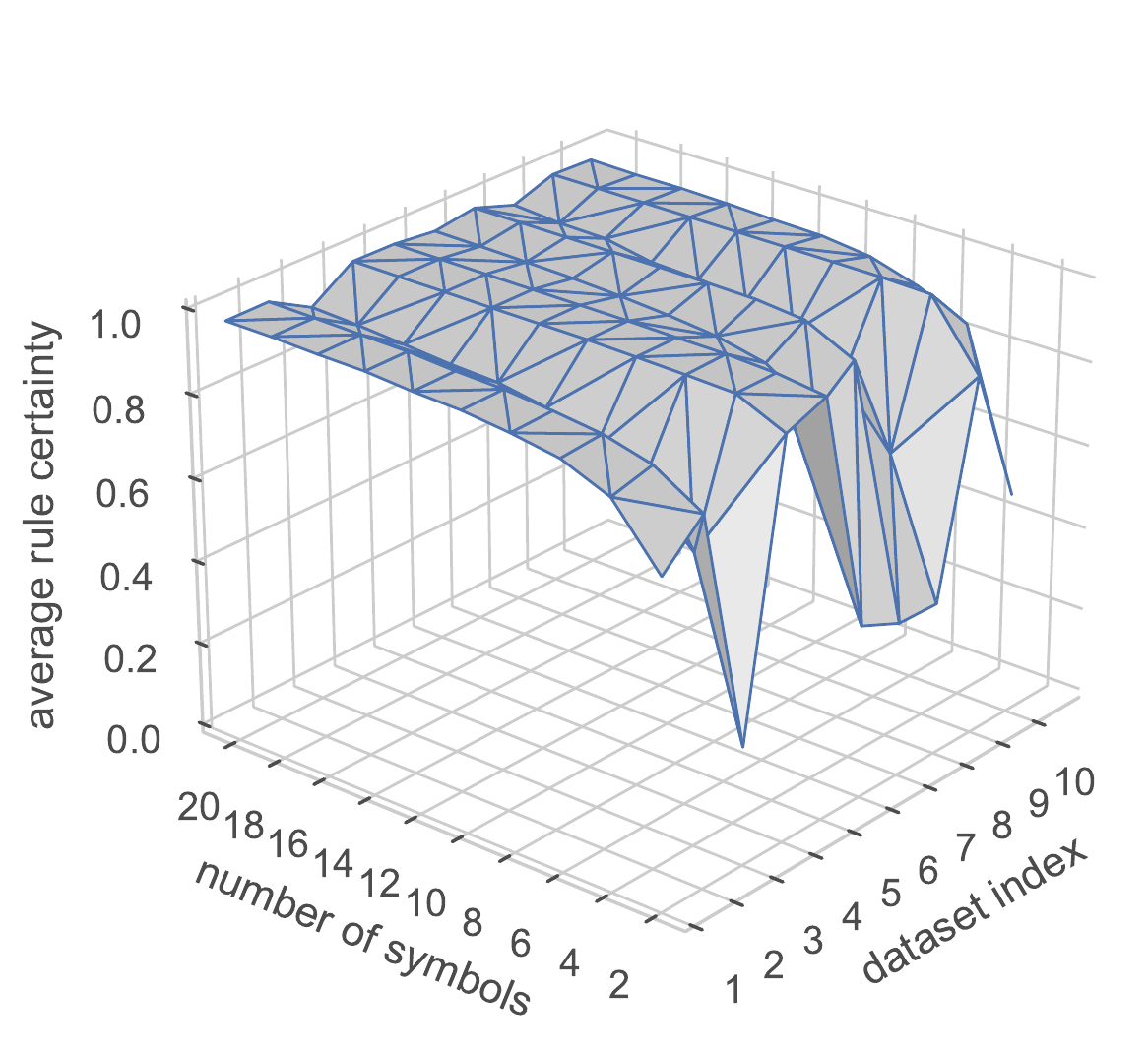}
	\caption{Łukasiewicz and Equation \eqref{eq:distance-local1}}
	\end{subfigure}
	\begin{subfigure}{0.49\textwidth}
	\center
	\includegraphics[width=\textwidth]{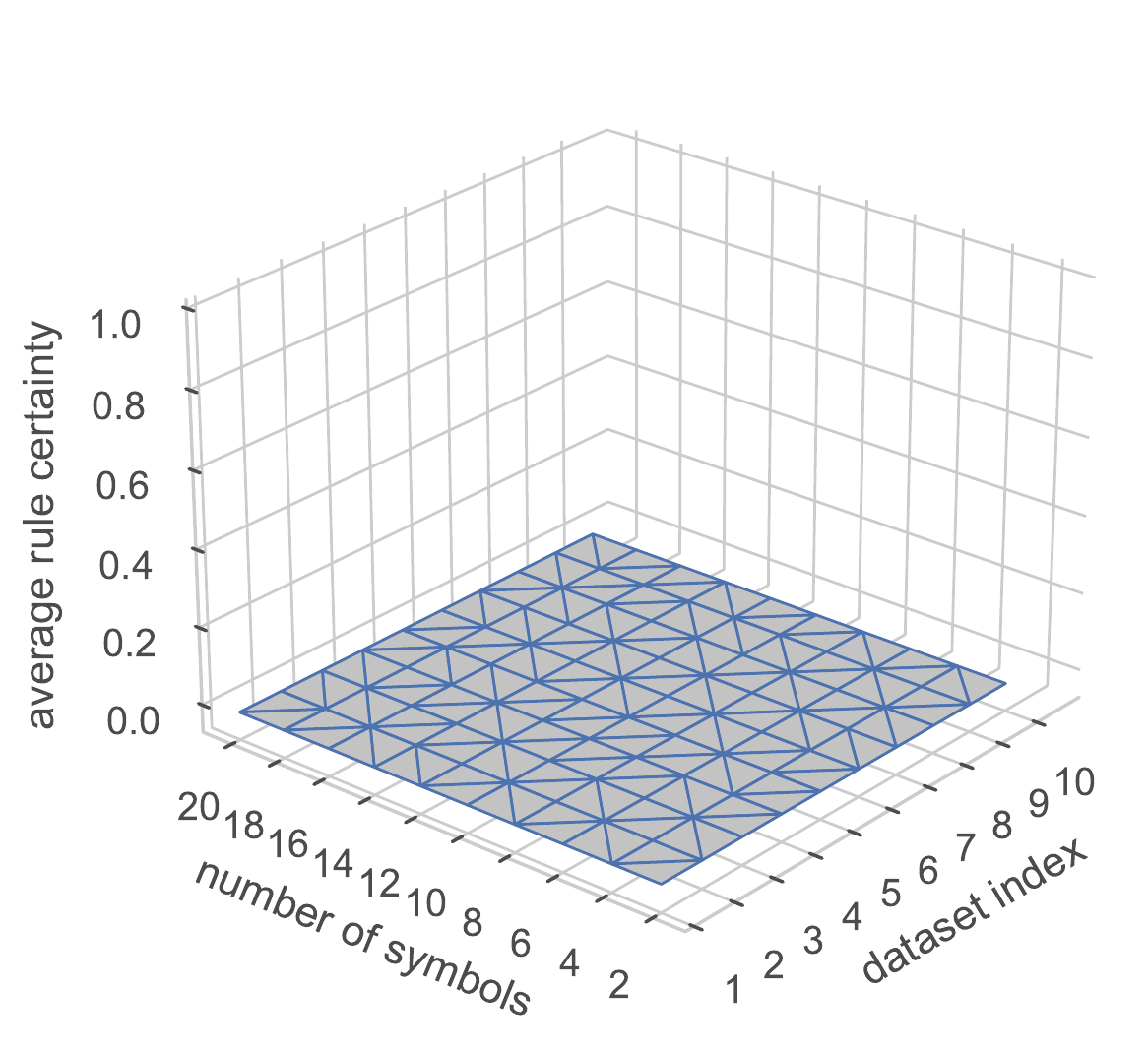}
	\caption{G\"odel and Equation \eqref{eq:distance-local2}}
	\end{subfigure}
	\begin{subfigure}{0.49\textwidth}
	\center
	\includegraphics[width=\textwidth]{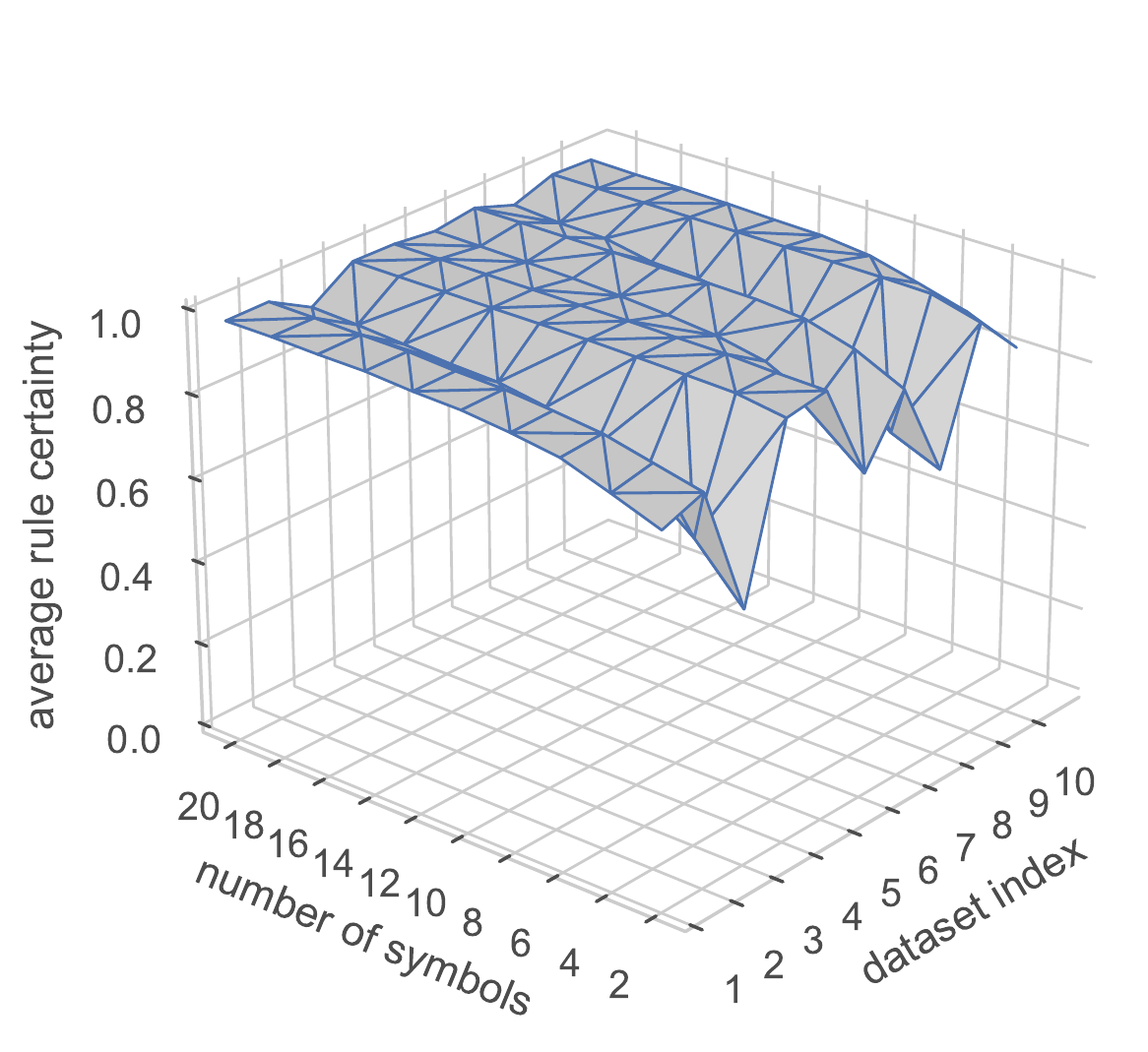}
	\caption{Łukasiewicz and Equation \eqref{eq:distance-local2}}
	\end{subfigure}	
	
	\captionsetup{justification=justified}
	\caption{Average rule confidence for each dataset when varying the number of symbols and the fuzzy implication function. Only G\"odel and Łukasiewicz are visualized. In these simulations, a Support Vector Machine is a black box.}
\label{fig:svm-terms}
\end{figure*}

\begin{figure*}[!htbp]
\center
    \begin{subfigure}{0.49\textwidth}
	\center
	\includegraphics[width=\textwidth]{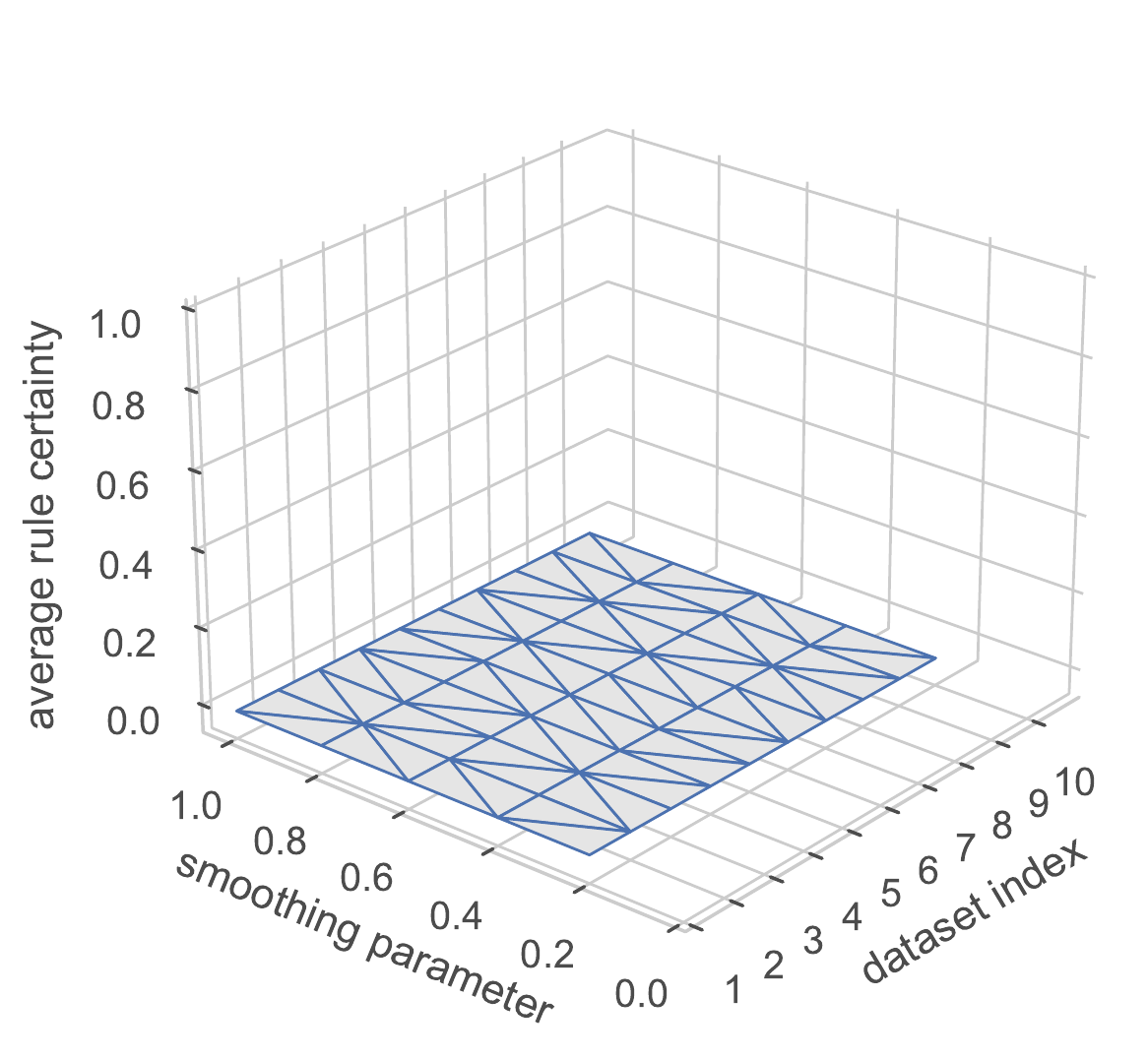}
	\caption{G\"odel and Equation \eqref{eq:distance-local1}}
	\end{subfigure}
	\begin{subfigure}{0.49\textwidth}
	\center
	\includegraphics[width=\textwidth]{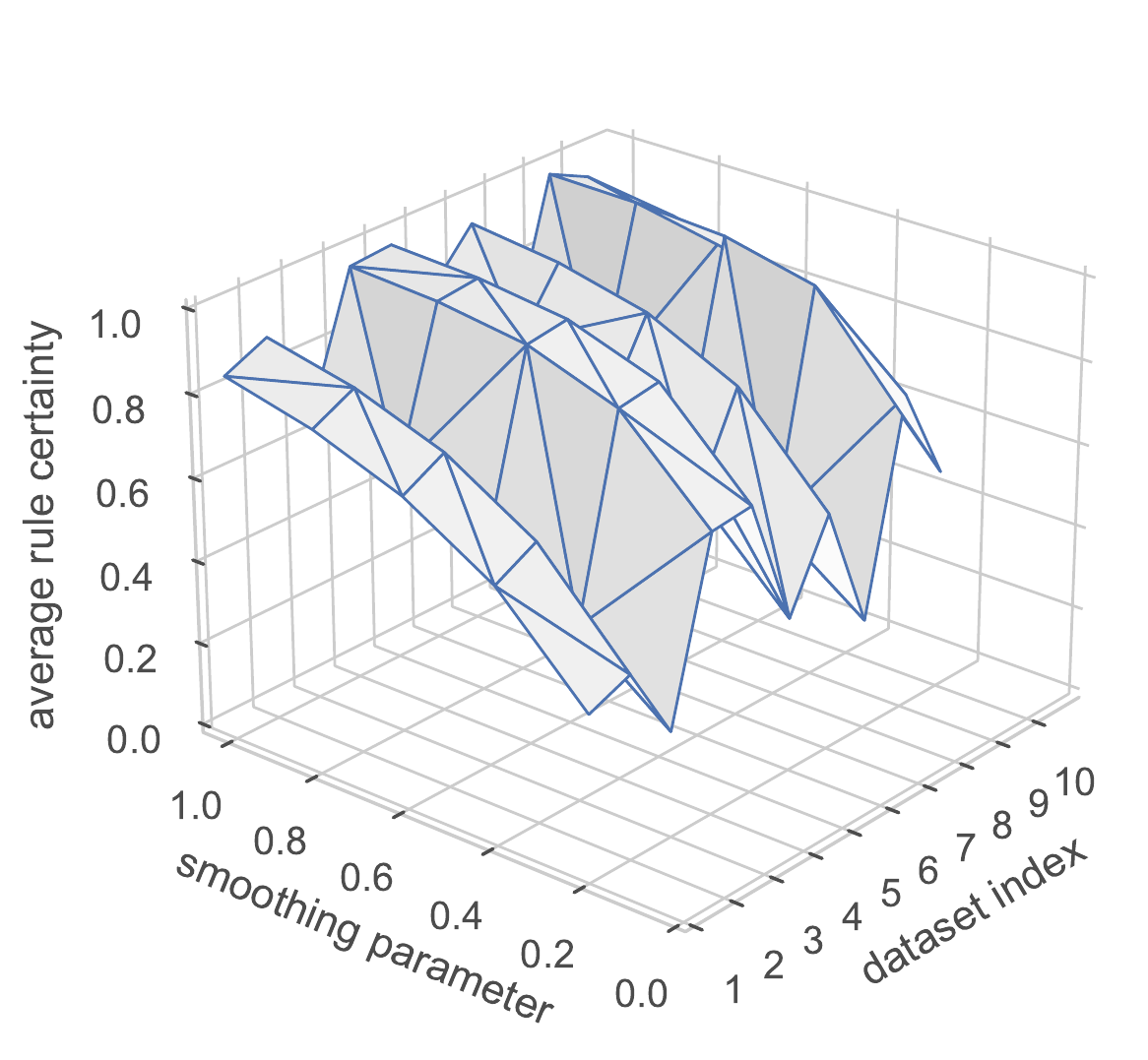}
	\caption{Łukasiewicz and Equation \eqref{eq:distance-local1}}
	\end{subfigure}
	\begin{subfigure}{0.49\textwidth}
	\center
	\includegraphics[width=\textwidth]{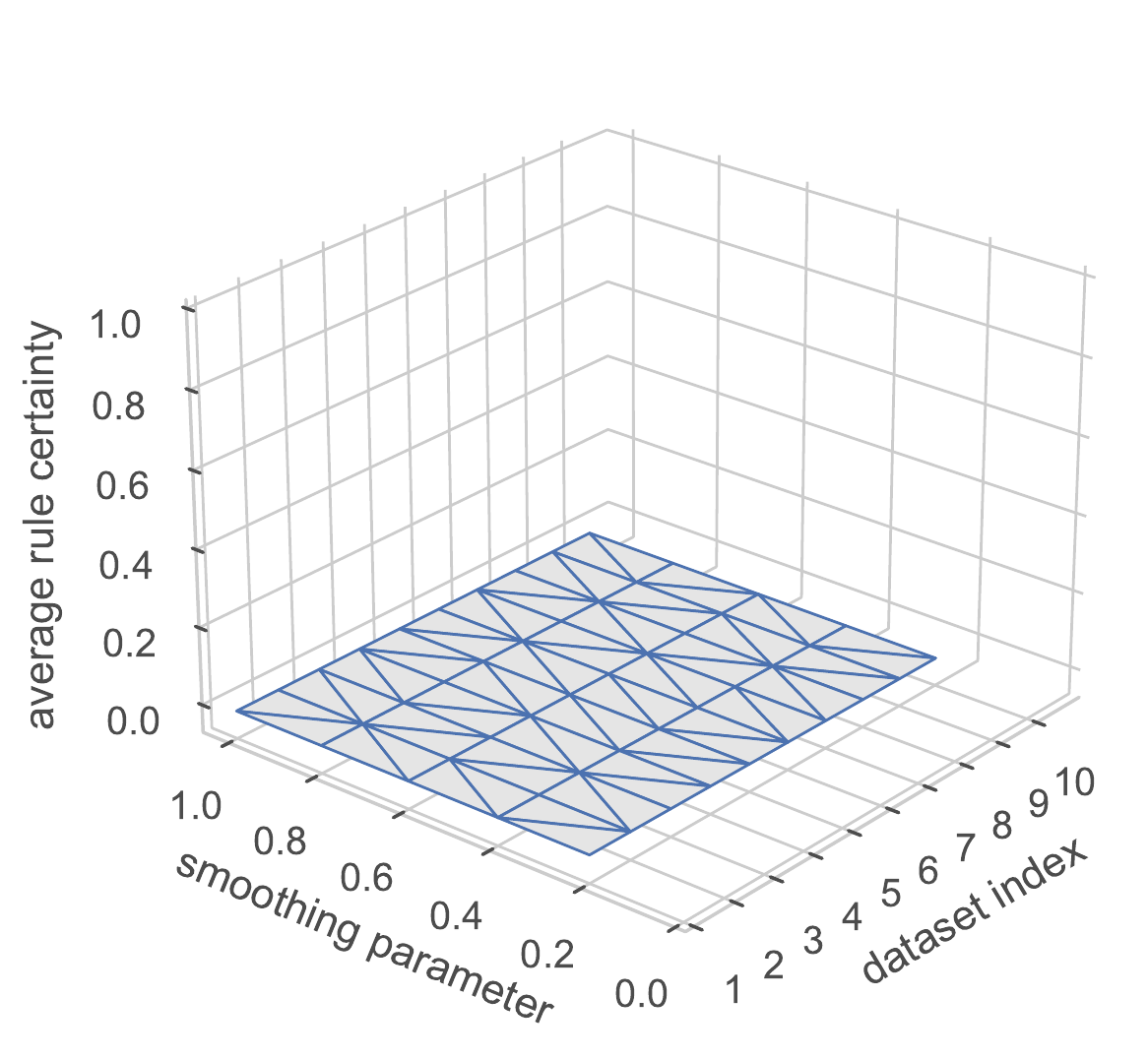}
	\caption{G\"odel and Equation \eqref{eq:distance-local2}}
	\end{subfigure}
	\begin{subfigure}{0.49\textwidth}
	\center
	\includegraphics[width=\textwidth]{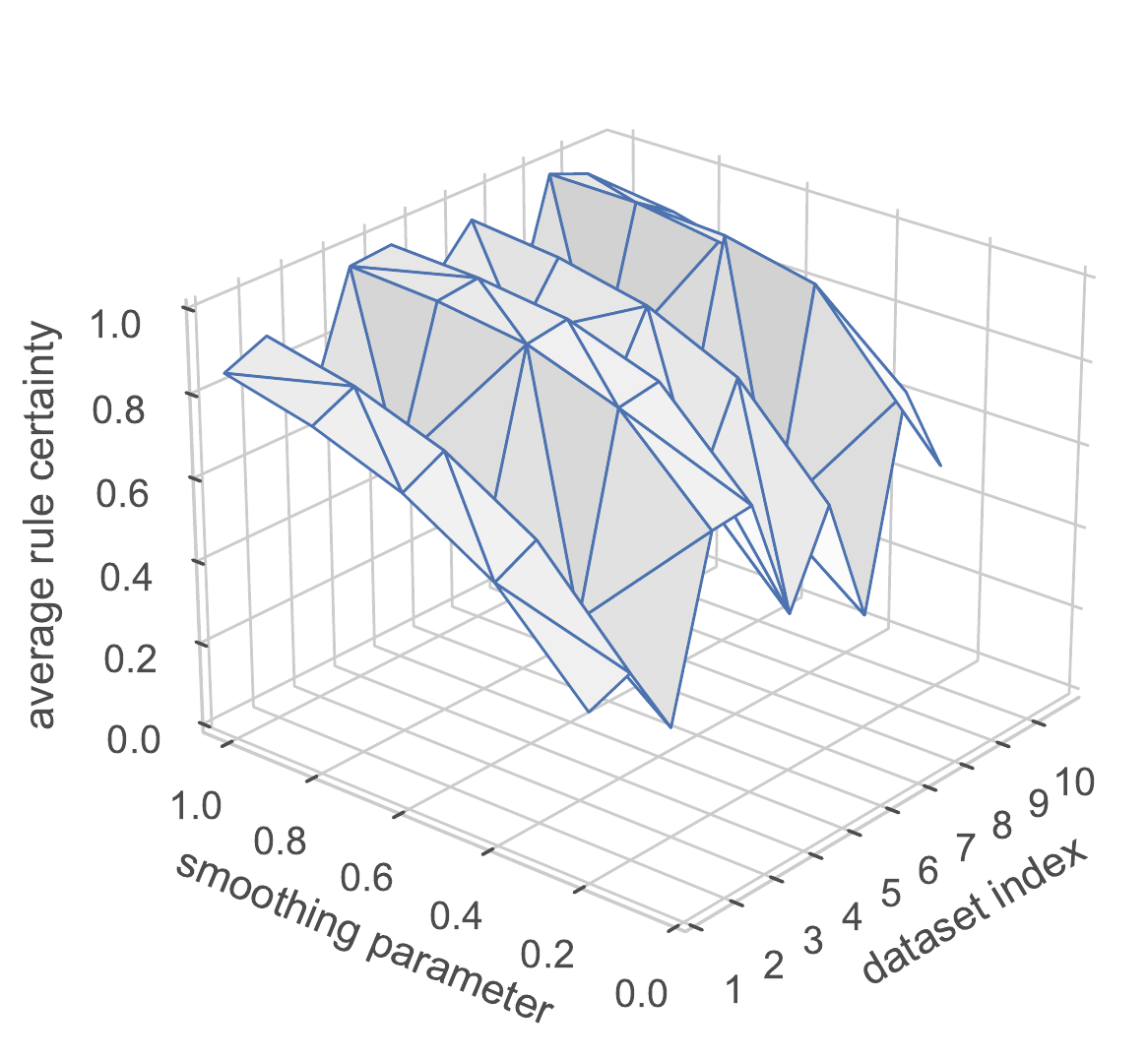}
	\caption{Łukasiewicz and Equation \eqref{eq:distance-local2}}
	\end{subfigure}	
	
	\captionsetup{justification=justified}
	\caption{Average rule confidence for each dataset when varying the smoothing parameter and the fuzzy implication function. Only G\"odel and Łukasiewicz are visualized. In these simulations, a Support Vector Machine is a black box.}
\label{fig:svm-smoothing}
\end{figure*}

Figures \ref{fig:svm-terms} and \ref{fig:svm-smoothing} present the average rule confidence obtained when changing the number of symbols and the smoothing parameter, respectively. Here, we use a Support Vector Machine using the default parameters in the \texttt{scikit-learn} library.

Figures \ref{fig:svm-terms} and \ref{fig:svm-smoothing} allow inspecting the differences between the implication functions and the distance functions. As expected, the average rule confidence is first and foremost dependent on the ``essence'' of the model, that is, the chosen implication function. Secondly, the behavior concerning the value of the smoothing parameter and the number of symbols is very consistent. We shall observe that increasing either makes the average confidence saturate. After reaching a certain threshold, it shall stabilize. These outcomes are analogous to the results presented in the core part of the manuscript concerning the random forest classifier.

\paragraph{Logistic Regression}

\begin{figure*}[!htbp]
\center
    \begin{subfigure}{0.49\textwidth}
	\center
	\includegraphics[width=\textwidth]{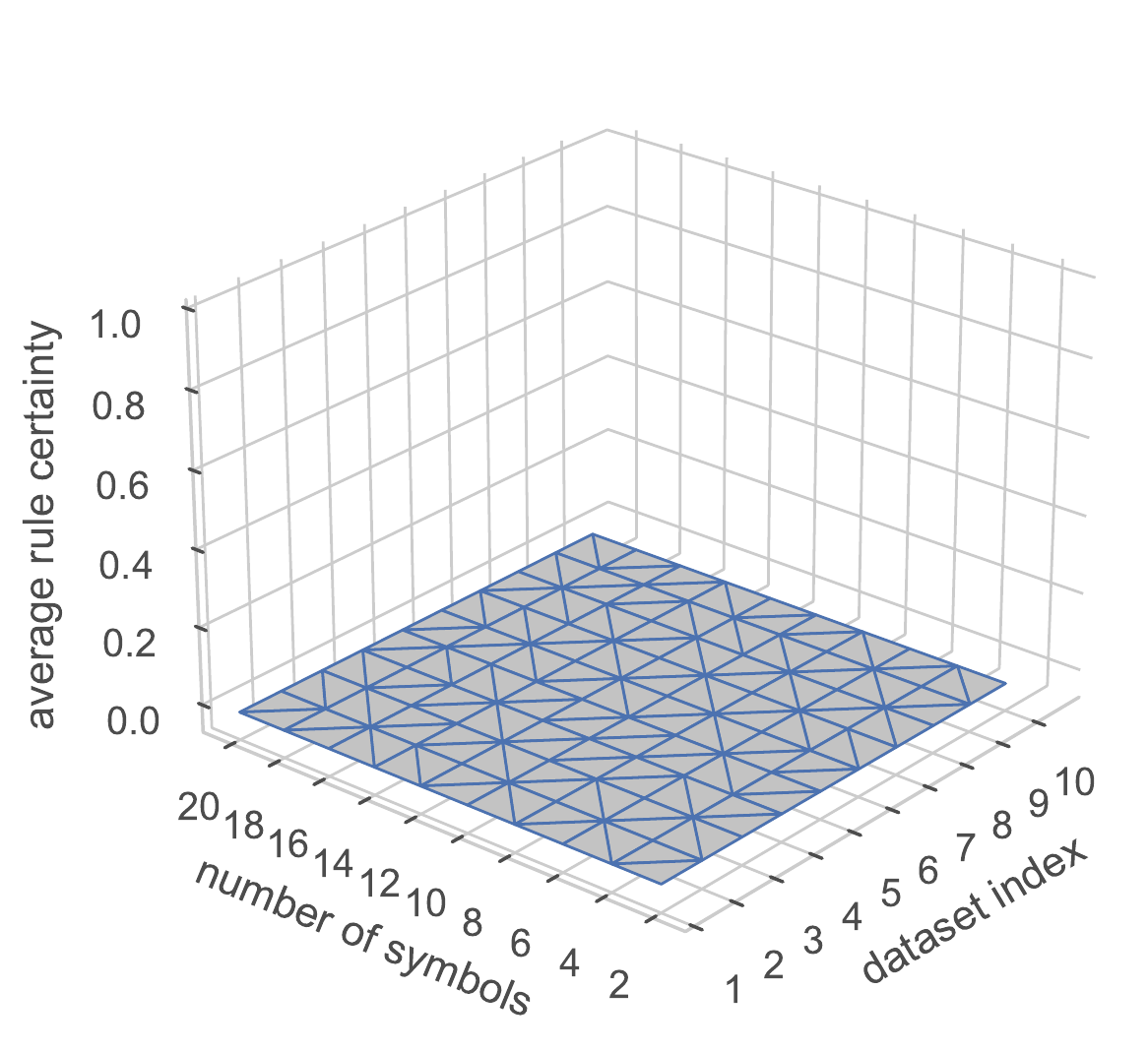}
	\caption{G\"odel and Equation \eqref{eq:distance-local1}}
	\end{subfigure}
	\begin{subfigure}{0.49\textwidth}
	\center
	\includegraphics[width=\textwidth]{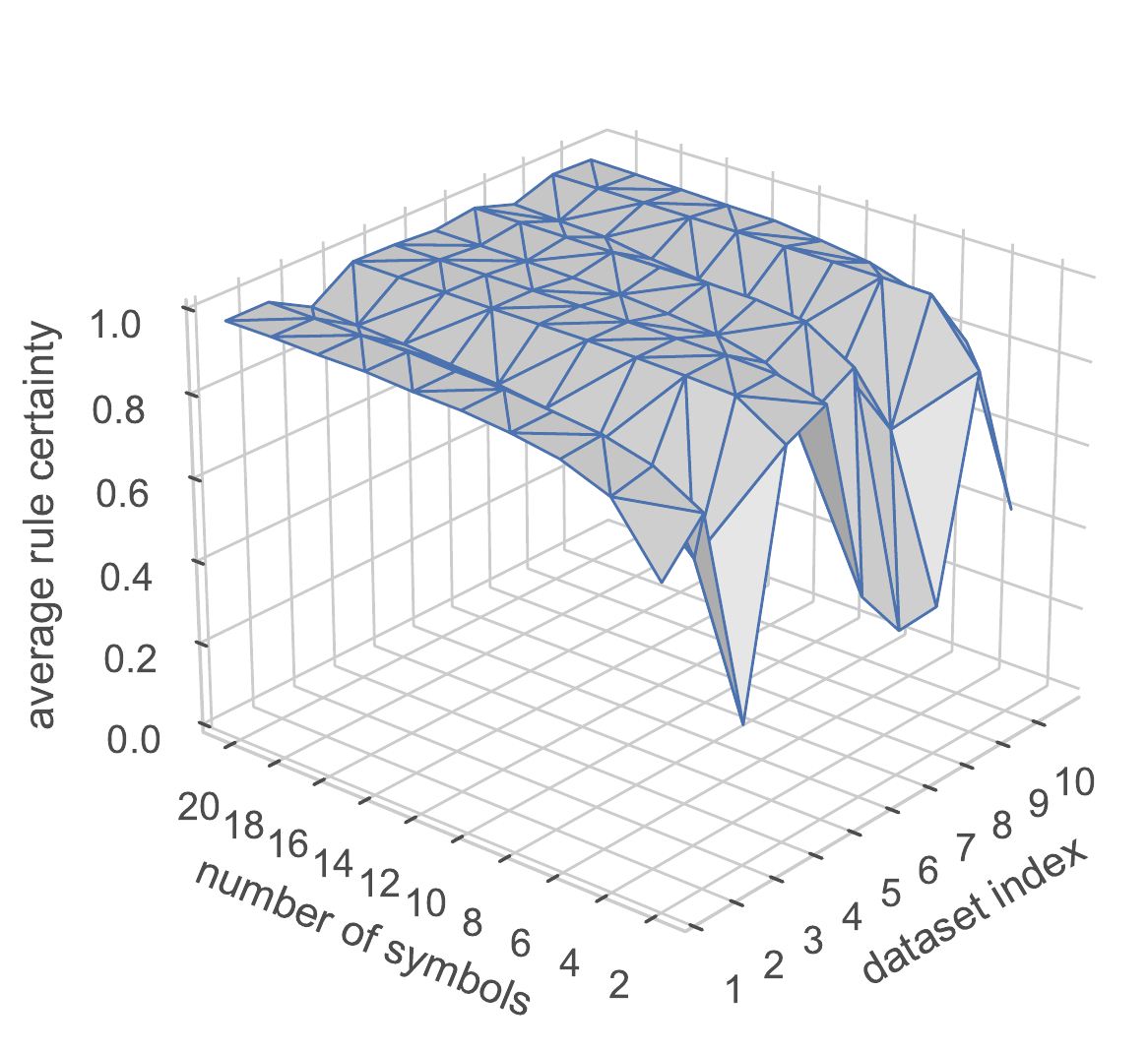}
	\caption{Łukasiewicz and Equation \eqref{eq:distance-local1}}
	\end{subfigure}
	\begin{subfigure}{0.49\textwidth}
	\center
	\includegraphics[width=\textwidth]{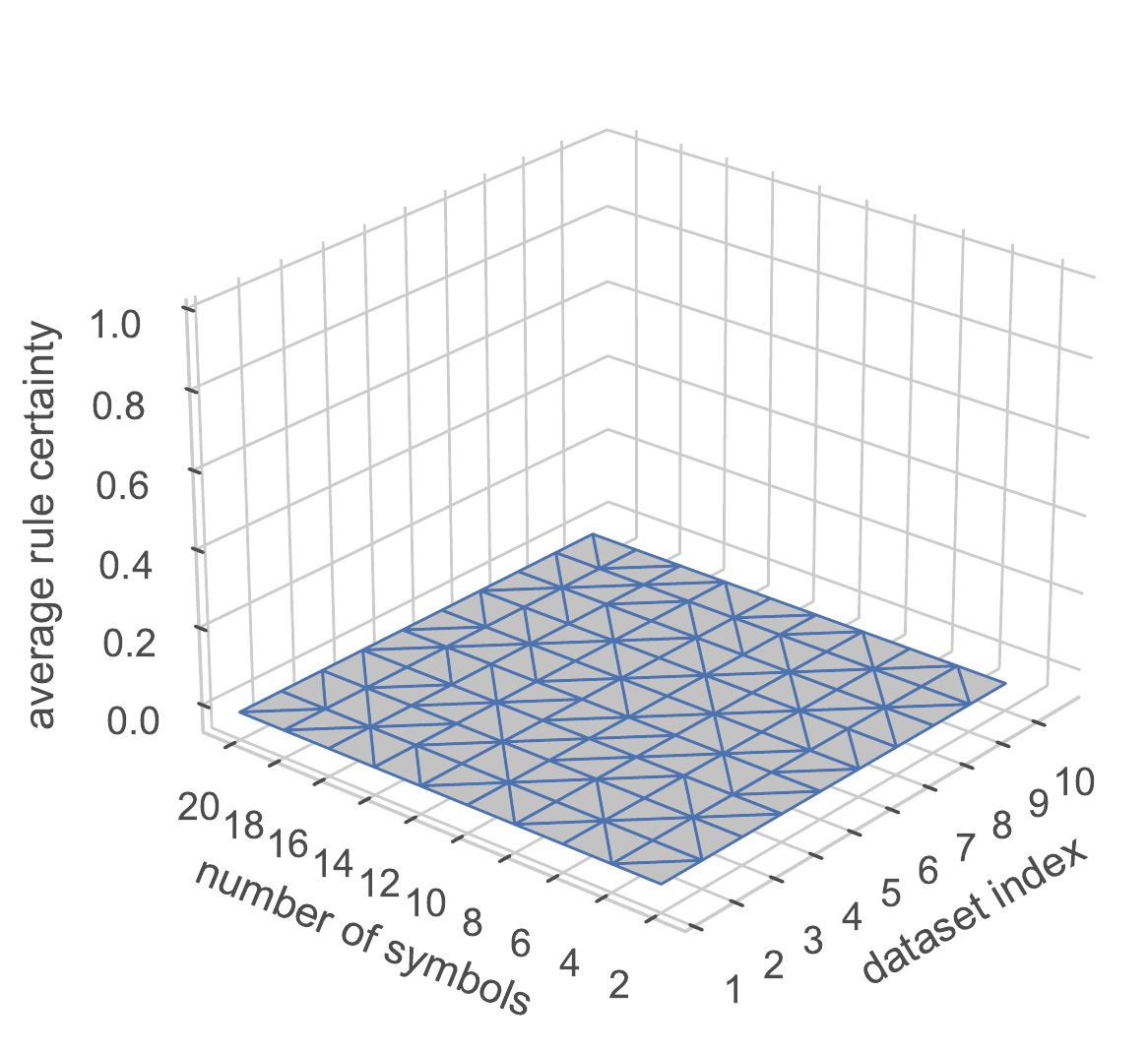}
	\caption{G\"odel and Equation \eqref{eq:distance-local2}}
	\end{subfigure}
	\begin{subfigure}{0.49\textwidth}
	\center
	\includegraphics[width=\textwidth]{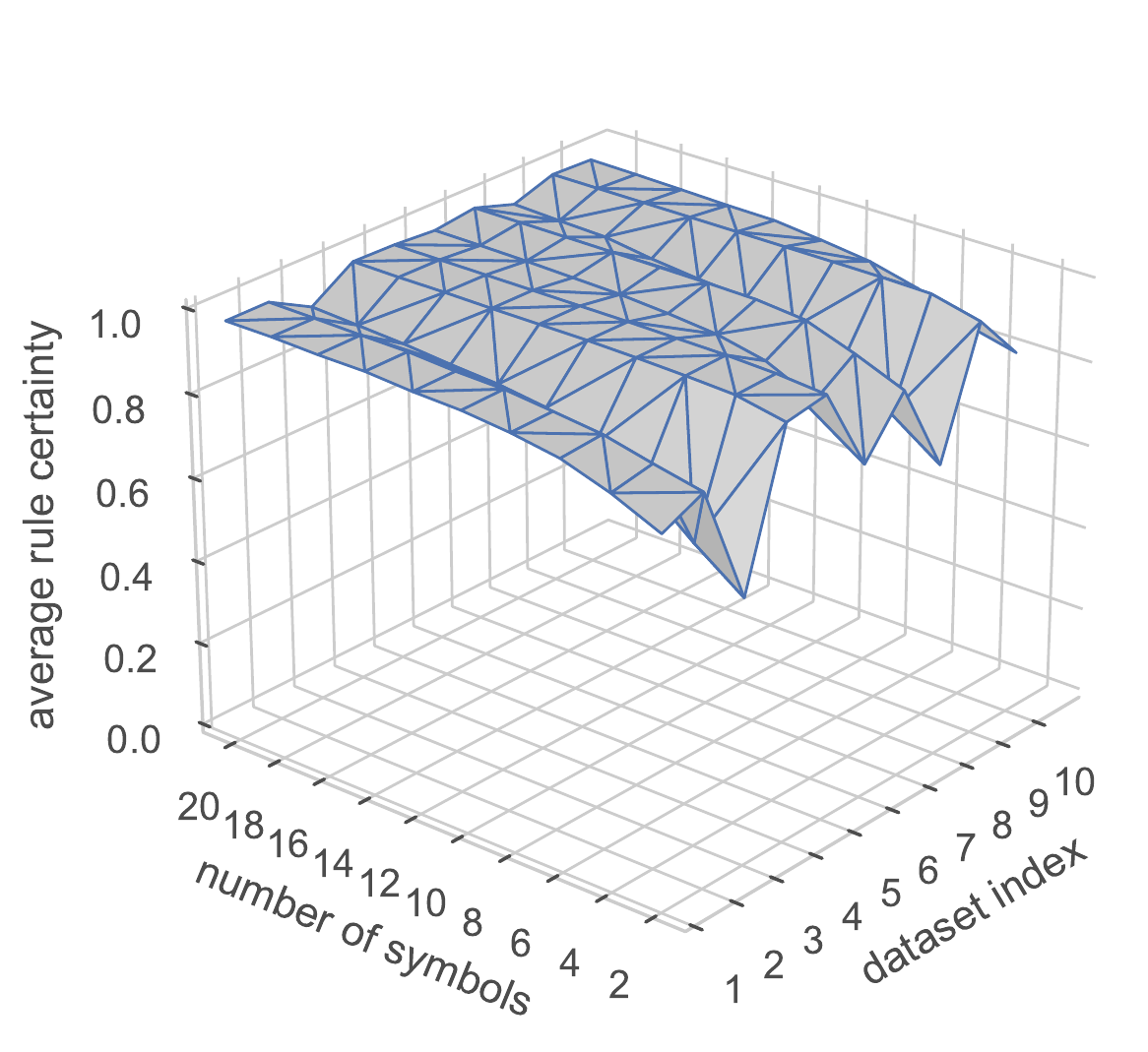}
	\caption{Łukasiewicz and Equation \eqref{eq:distance-local2}}
	\end{subfigure}

    \captionsetup{justification=justified}
	\caption{Average rule confidence for each dataset when varying the number of symbols and the fuzzy implication function. Only G\"odel and Łukasiewicz are visualized. In these simulations, Logistic Regression is the black box.}
\label{fig:lr-terms}
\end{figure*}

\begin{figure*}[!htbp]
\center

    \begin{subfigure}{0.49\textwidth}
	\center
	\includegraphics[width=\textwidth]{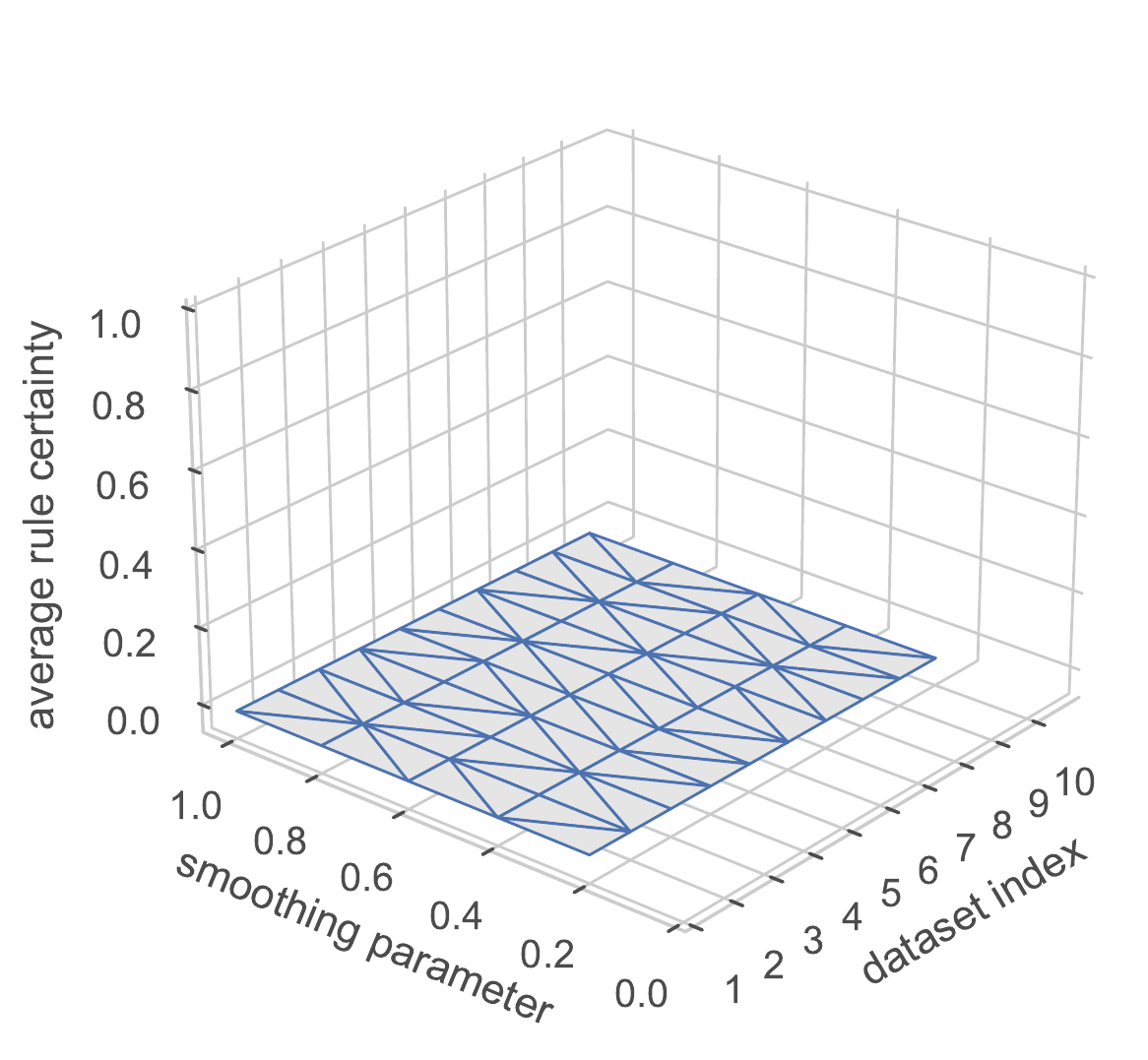}
	\caption{G\"odel and Equation \eqref{eq:distance-local1}}
	\end{subfigure}
	\begin{subfigure}{0.49\textwidth}
	\center
	\includegraphics[width=\textwidth]{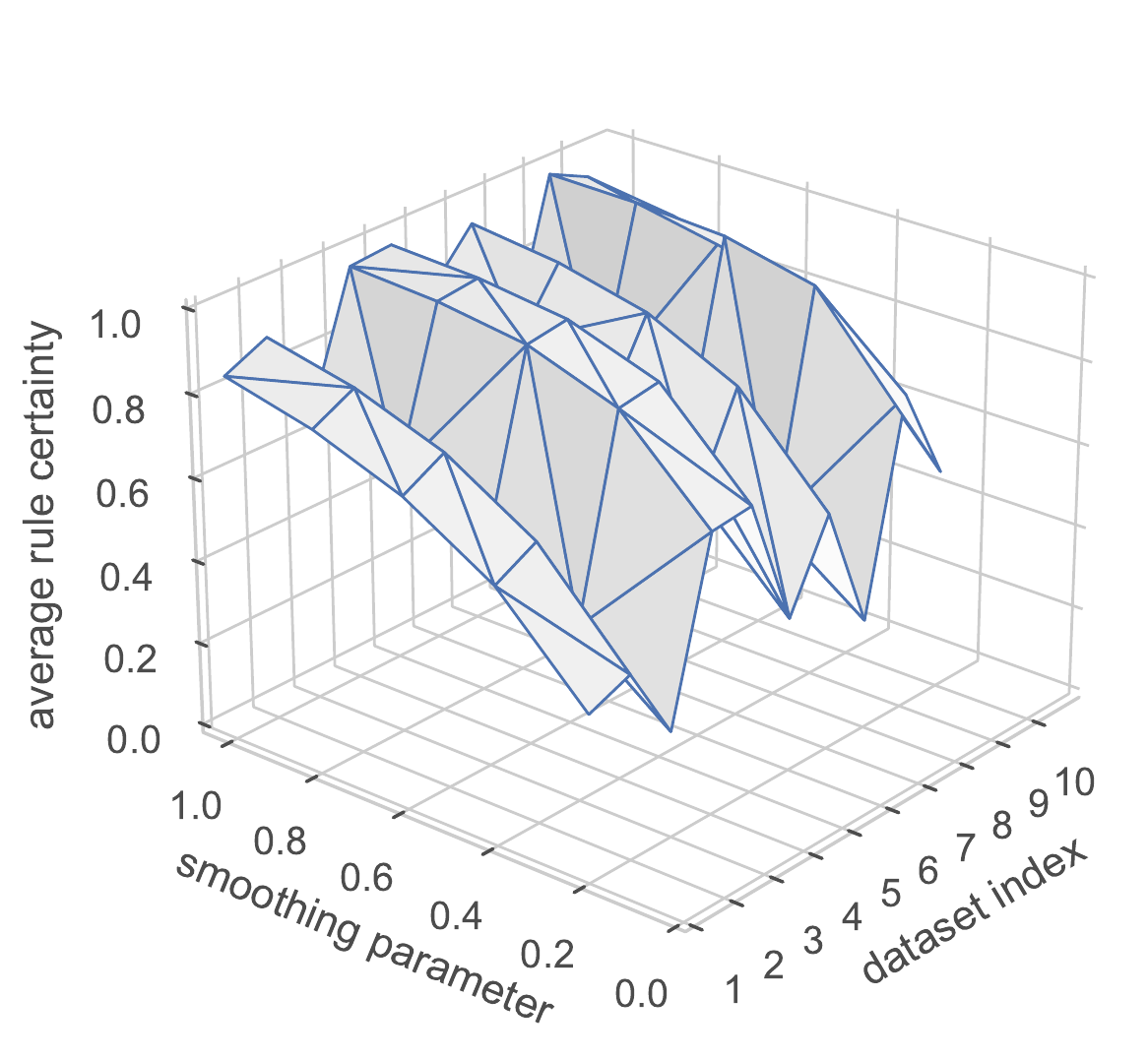}
	\caption{Łukasiewicz and Equation \eqref{eq:distance-local1}}
	\end{subfigure}
	\begin{subfigure}{0.49\textwidth}
	\center
	\includegraphics[width=\textwidth]{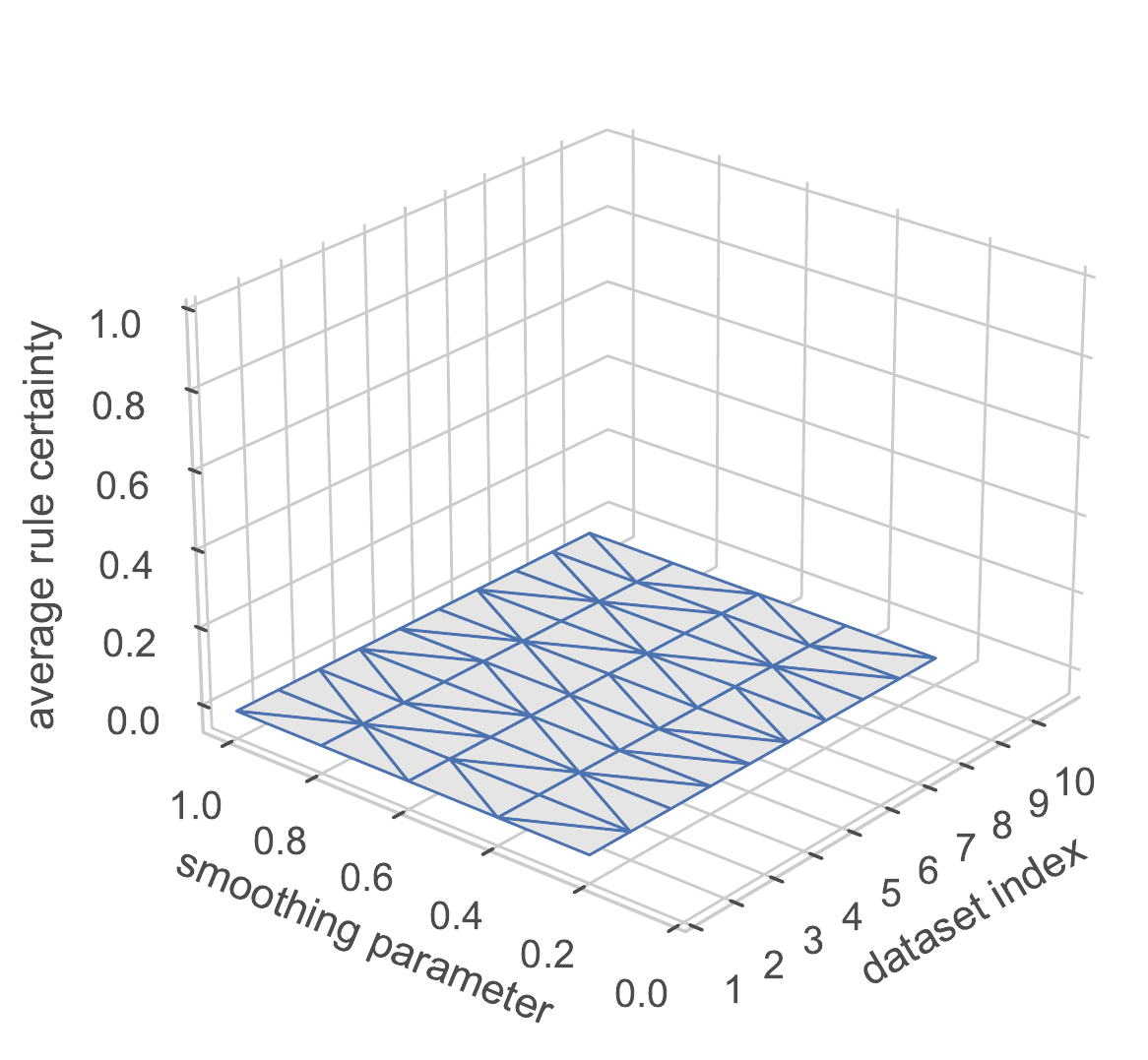}
	\caption{G\"odel and Equation \eqref{eq:distance-local2}}
	\end{subfigure}
	\begin{subfigure}{0.49\textwidth}
	\center
	\includegraphics[width=\textwidth]{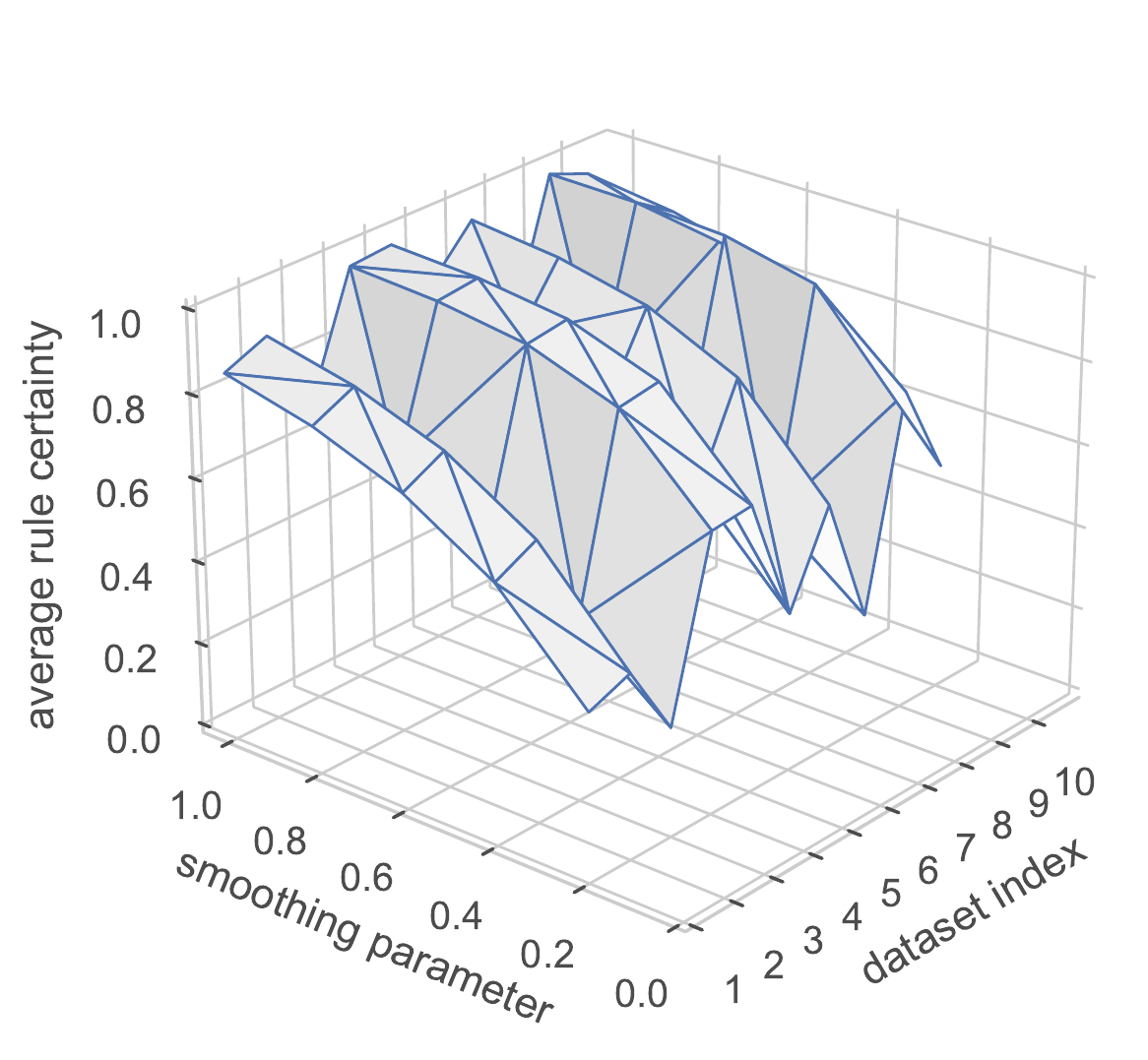}
	\caption{Łukasiewicz and Equation \eqref{eq:distance-local2}}
	\end{subfigure}	
    
    \captionsetup{justification=justified}
	\caption{Average rule confidence for each dataset when varying the smoothing parameter and the fuzzy implication function. Only G\"odel and Łukasiewicz are visualized. In these simulations, Logistic Regression is the black box.}
\label{fig:lr-smoothing}
\end{figure*}

Subsequently, we built a Logistic Regression model using the default parameters in the \texttt{scikit-learn} library. Figures \ref{fig:lr-terms} and \ref{fig:lr-smoothing} show the influence of the number of symbols and the smoothing parameter on the average rule confidence. The plots exhibit similar patterns to the ones observed for Random Forests and Support Vector Machines. If we want to draw more specific conclusions, we should mention that the sensitivity analysis revealed that the smoothing parameter has a greater influence on the average rule confidence than the number of symbols. This was also observed in the cases addressed before.

\paragraph{k-Nearest Neighbors}

\begin{figure*}[!htbp]
\center
    \begin{subfigure}{0.49\textwidth}
	\center
	\includegraphics[width=\textwidth]{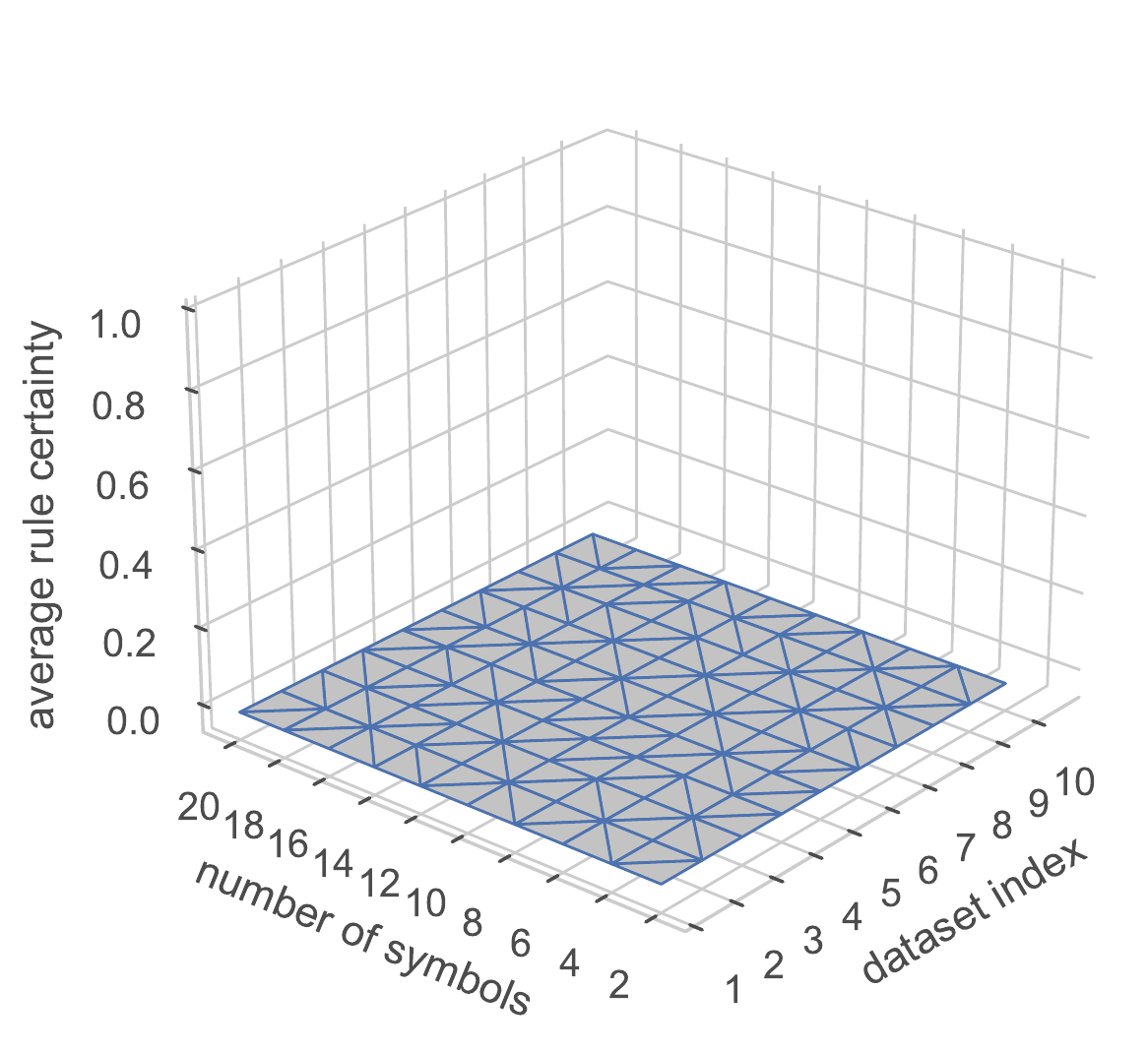}
	\caption{G\"odel and Equation \eqref{eq:distance-local1}}
	\end{subfigure}
	\begin{subfigure}{0.49\textwidth}
	\center
	\includegraphics[width=\textwidth]{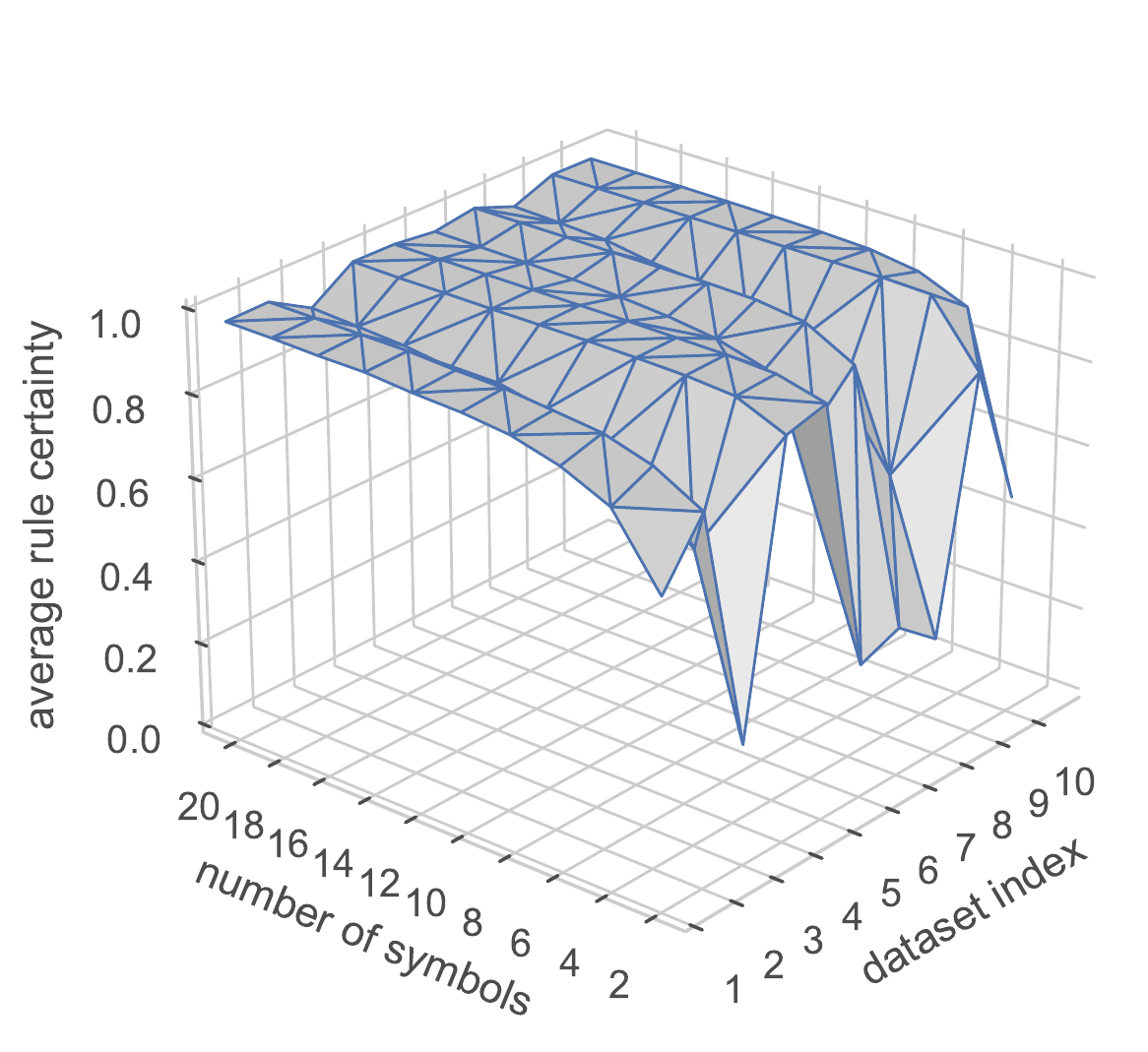}
	\caption{Łukasiewicz and Equation \eqref{eq:distance-local1}}
	\end{subfigure}
	\begin{subfigure}{0.49\textwidth}
	\center
	\includegraphics[width=\textwidth]{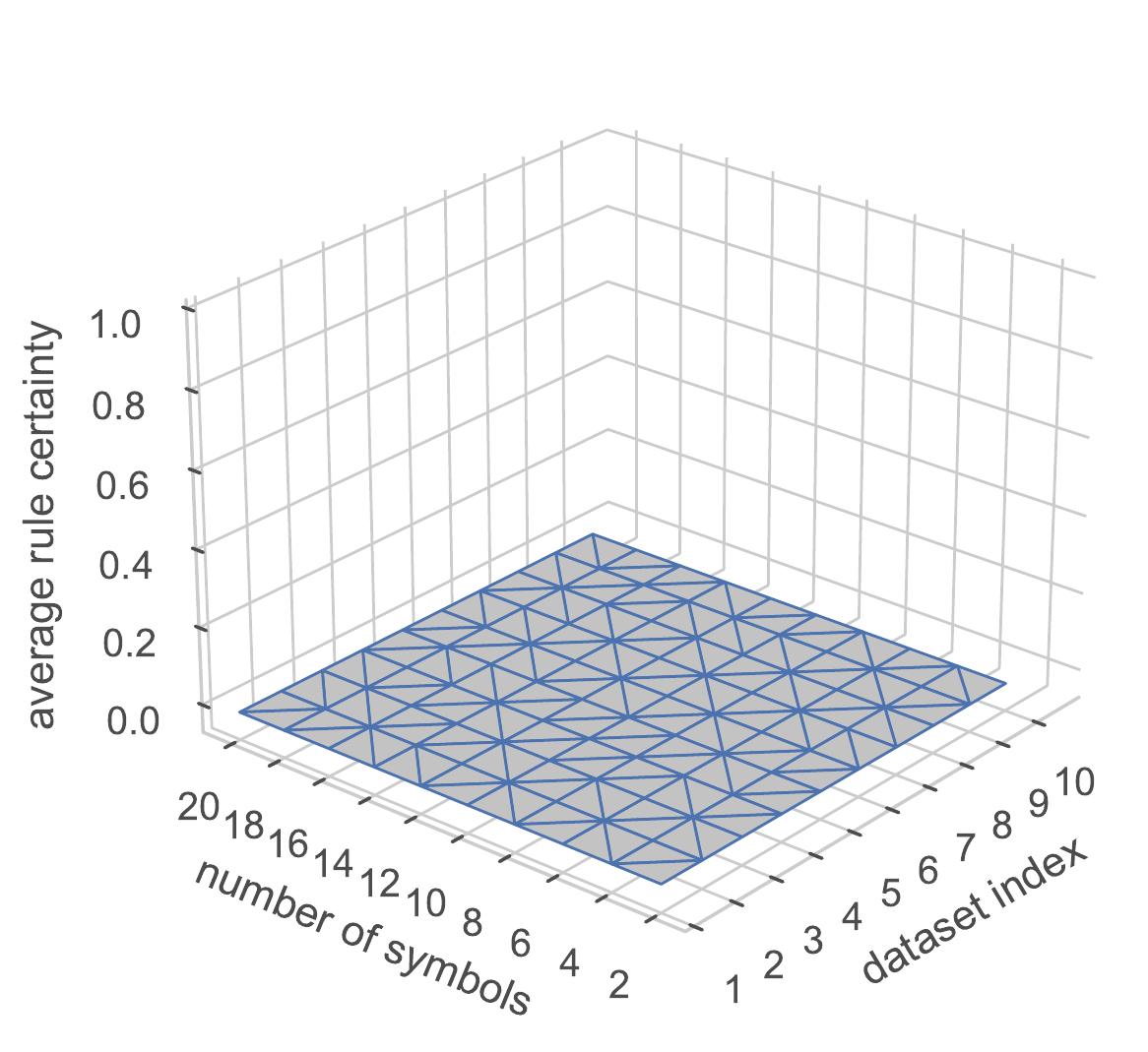}
	\caption{G\"odel and Equation \eqref{eq:distance-local2}}
	\end{subfigure}
	\begin{subfigure}{0.49\textwidth}
	\center
	\includegraphics[width=\textwidth]{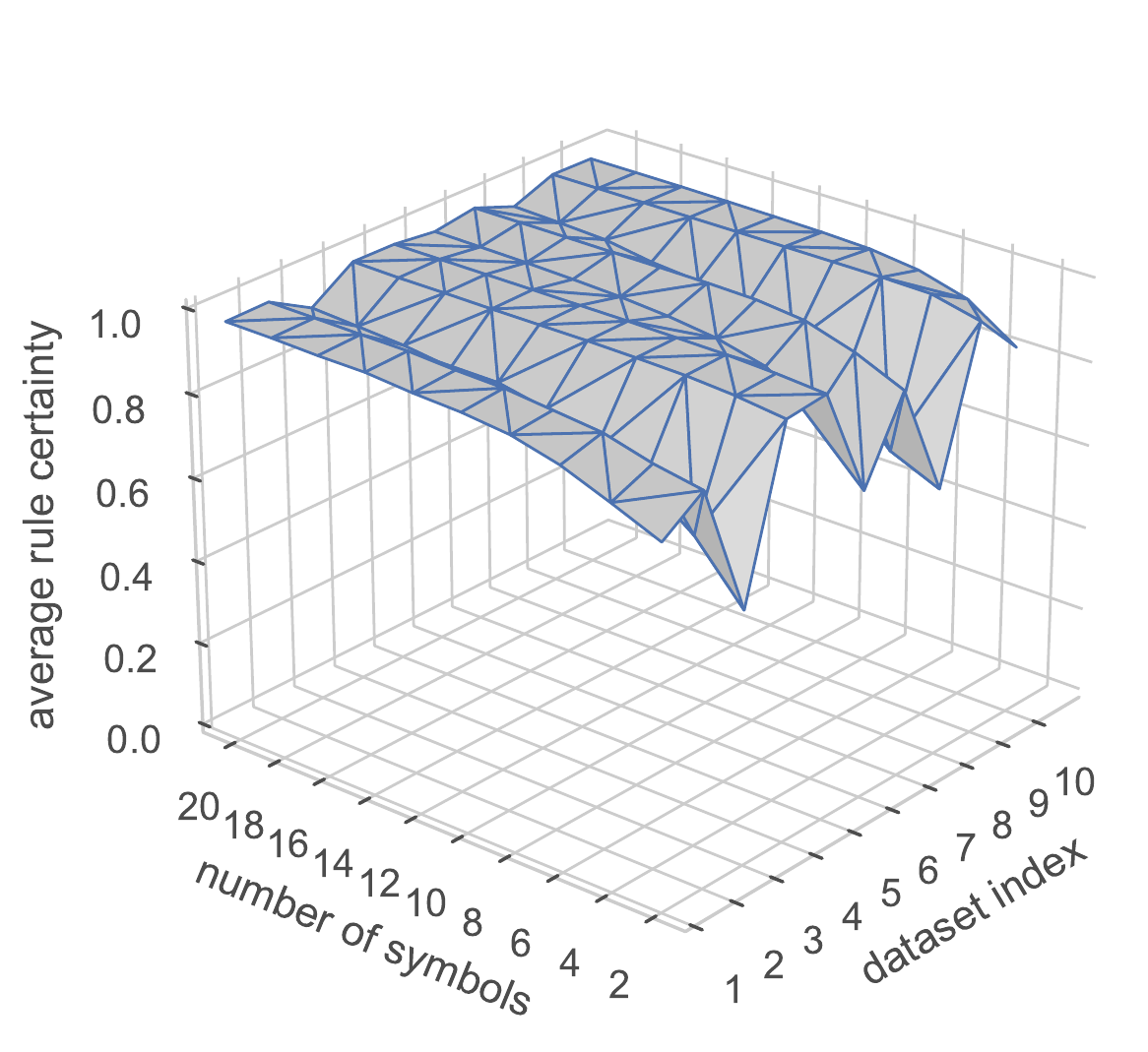}
	\caption{Łukasiewicz and Equation \eqref{eq:distance-local2}}
	\end{subfigure}	
	
	\captionsetup{justification=justified}
	\caption{Average rule confidence for each dataset when varying the number of symbols and the fuzzy implication function. Only G\"odel and Łukasiewicz are visualized. In these simulations, $k$-Nearest Neighbors is the black box.}
\label{fig:knn-terms}
\end{figure*}

\begin{figure*}[!htbp]
\center

    \begin{subfigure}{0.49\textwidth}
	\center
	\includegraphics[width=\textwidth]{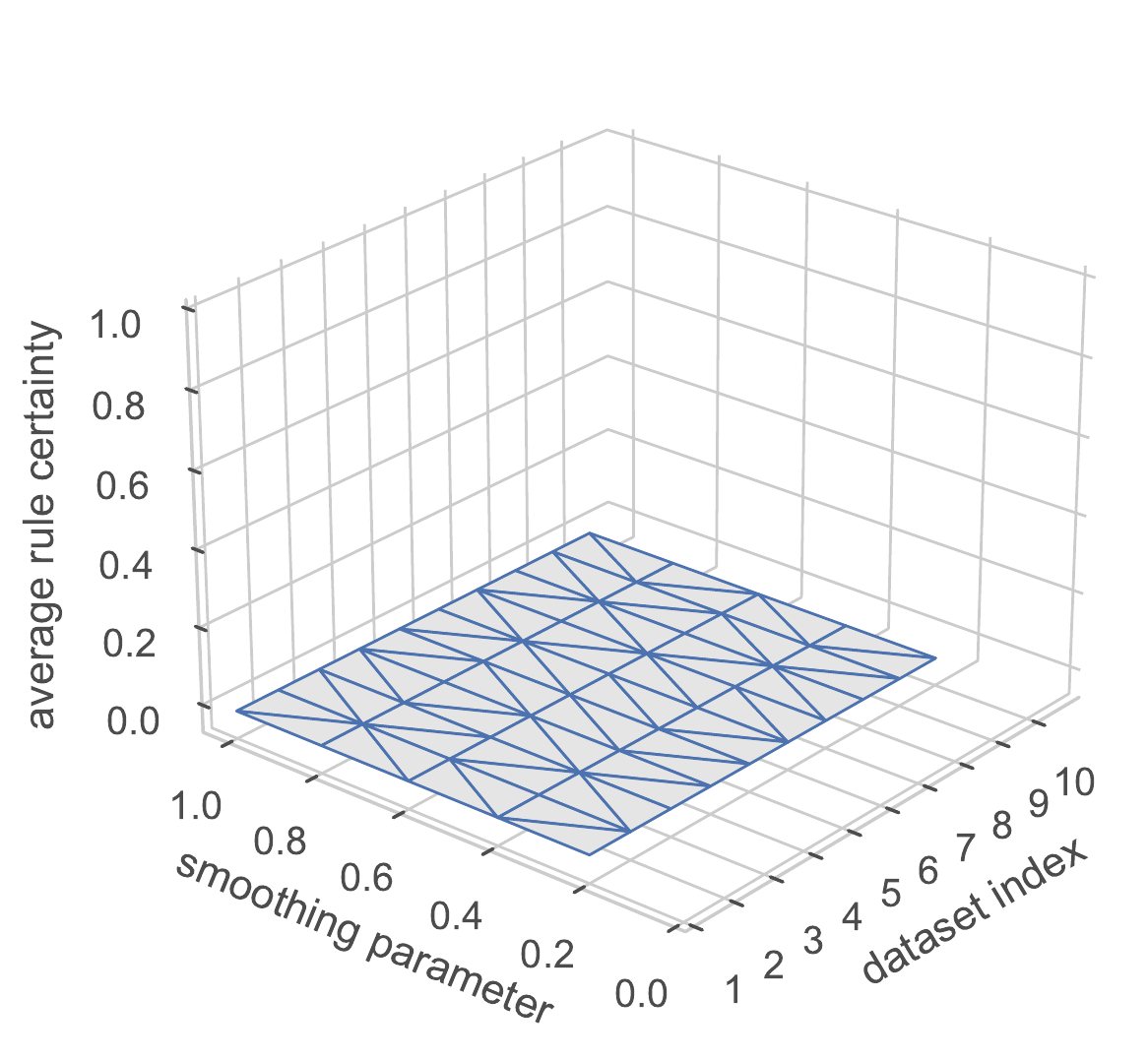}
	\caption{G\"odel and Equation \eqref{eq:distance-local1}}
	\end{subfigure}
	\begin{subfigure}{0.49\textwidth}
	\center
	\includegraphics[width=\textwidth]{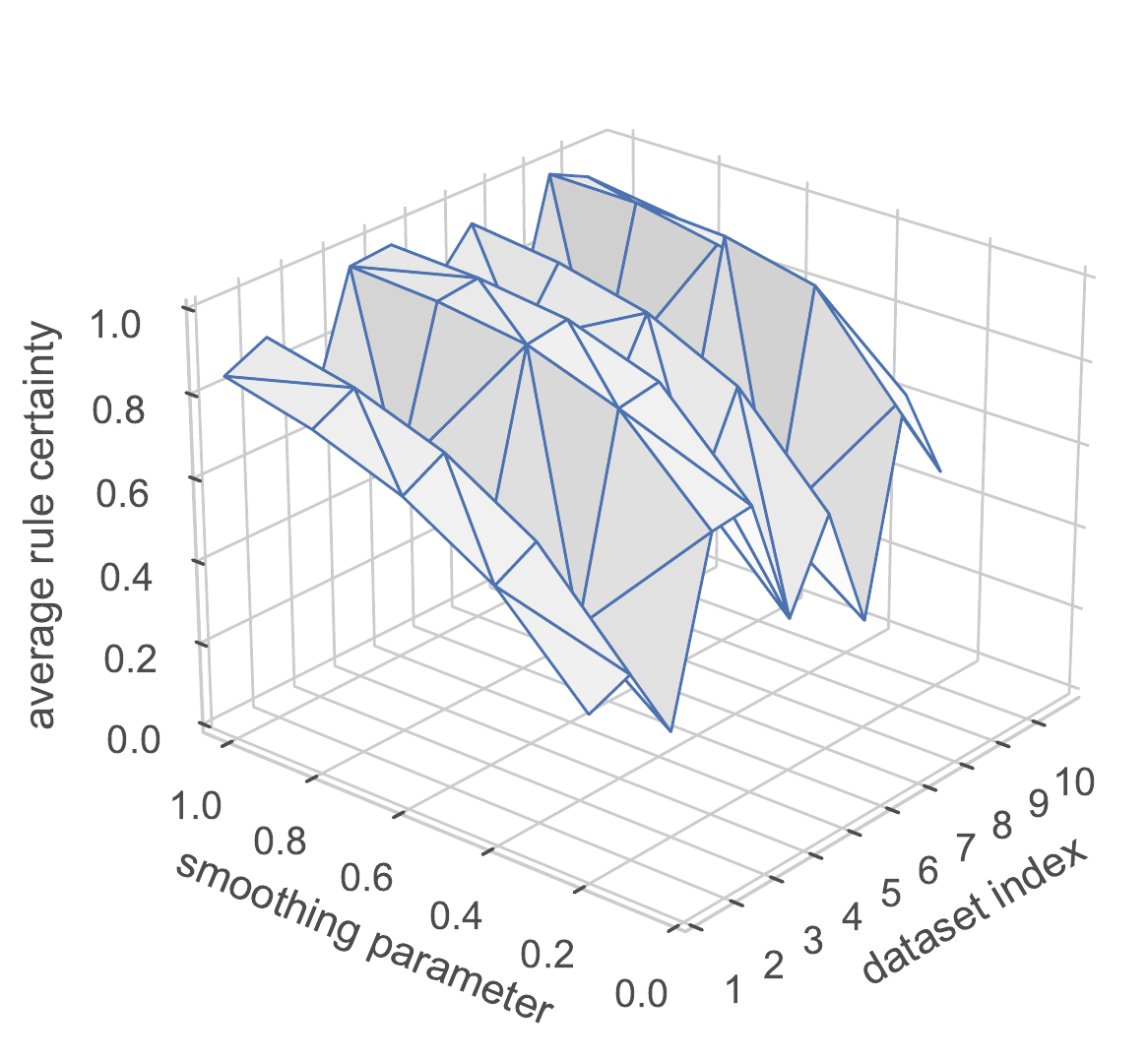}
	\caption{Łukasiewicz and Equation \eqref{eq:distance-local1}}
	\end{subfigure}
	\begin{subfigure}{0.49\textwidth}
	\center
	\includegraphics[width=\textwidth]{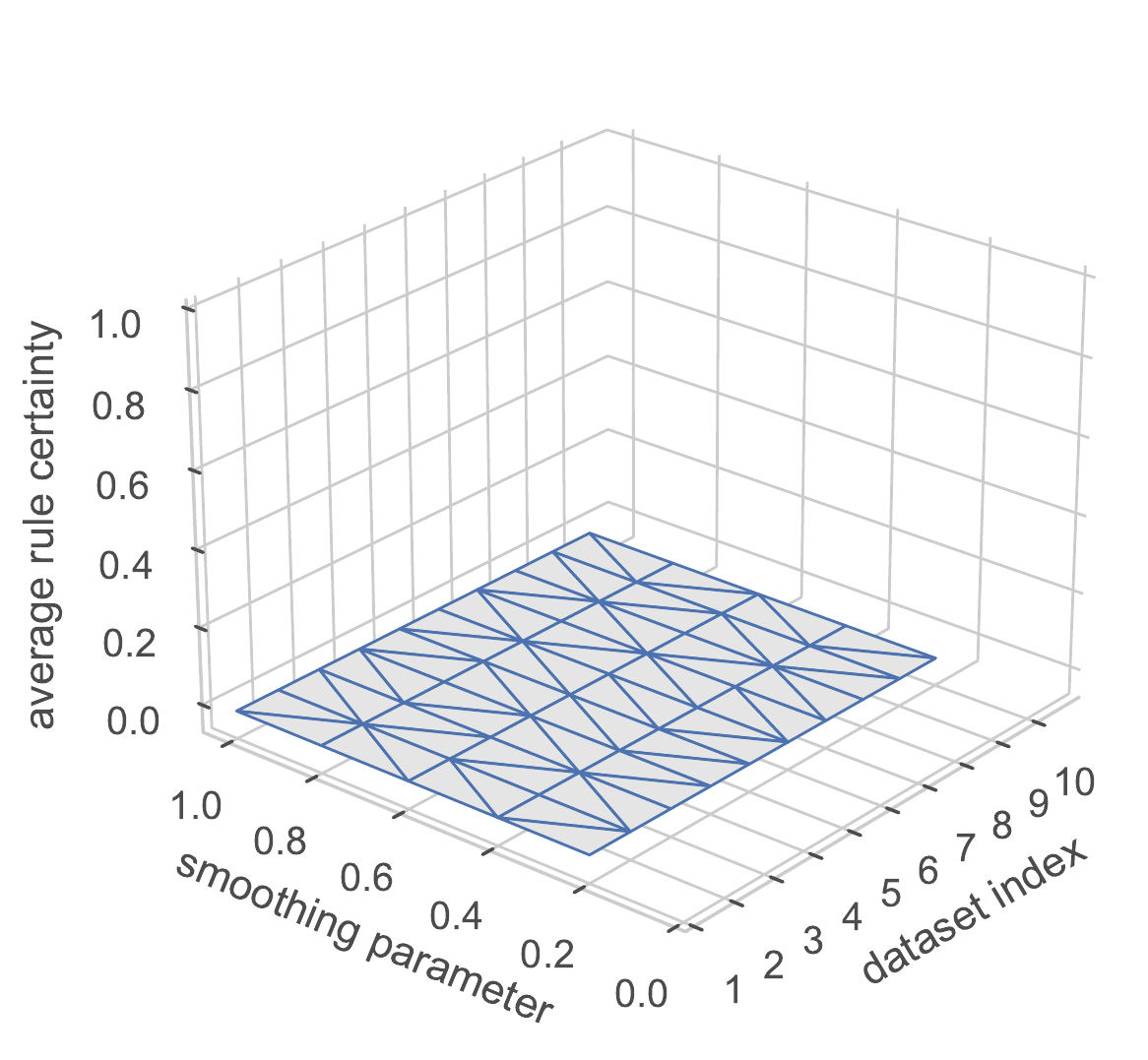}
	\caption{G\"odel and Equation \eqref{eq:distance-local2}}
	\end{subfigure}
	\begin{subfigure}{0.49\textwidth}
	\center
	\includegraphics[width=\textwidth]{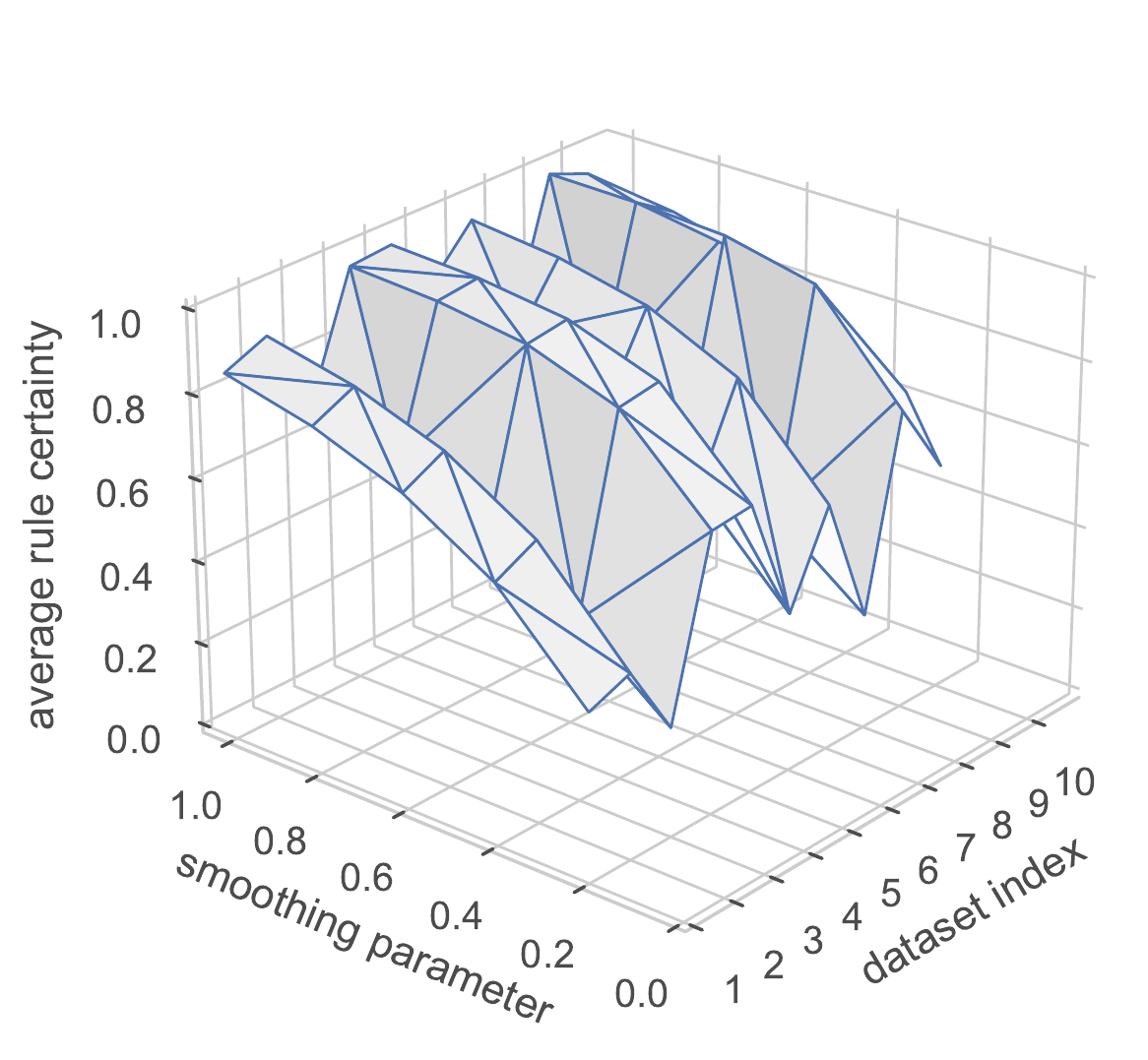}
	\caption{Łukasiewicz and Equation \eqref{eq:distance-local2}}
	\end{subfigure}	
	
	\captionsetup{justification=justified}
	\caption{Average rule confidence for each dataset when varying the smoothing parameter and the fuzzy implication function. Only G\"odel and Łukasiewicz are visualized. In these simulations, $k$-Nearest Neighbors is the black box.}
\label{fig:knn-smoothing}
\end{figure*}

The subsequent classifier that was tested is the k-Nearest Neighbor model. Figures \ref{fig:knn-terms} and \ref{fig:knn-smoothing} allow analyzing the sensitivity of the proposed rule-based approach to the changing values of the number of symbols and the smoothness parameter. The outcomes of this experiment, when framed in the form of generalized conclusions, are the same as the conclusions drawn from the previous classifiers. 

\paragraph{Na\"ive Bayes}

\begin{figure*}[!htbp]
\center
    \begin{subfigure}{0.49\textwidth}
	\center
	\includegraphics[width=\textwidth]{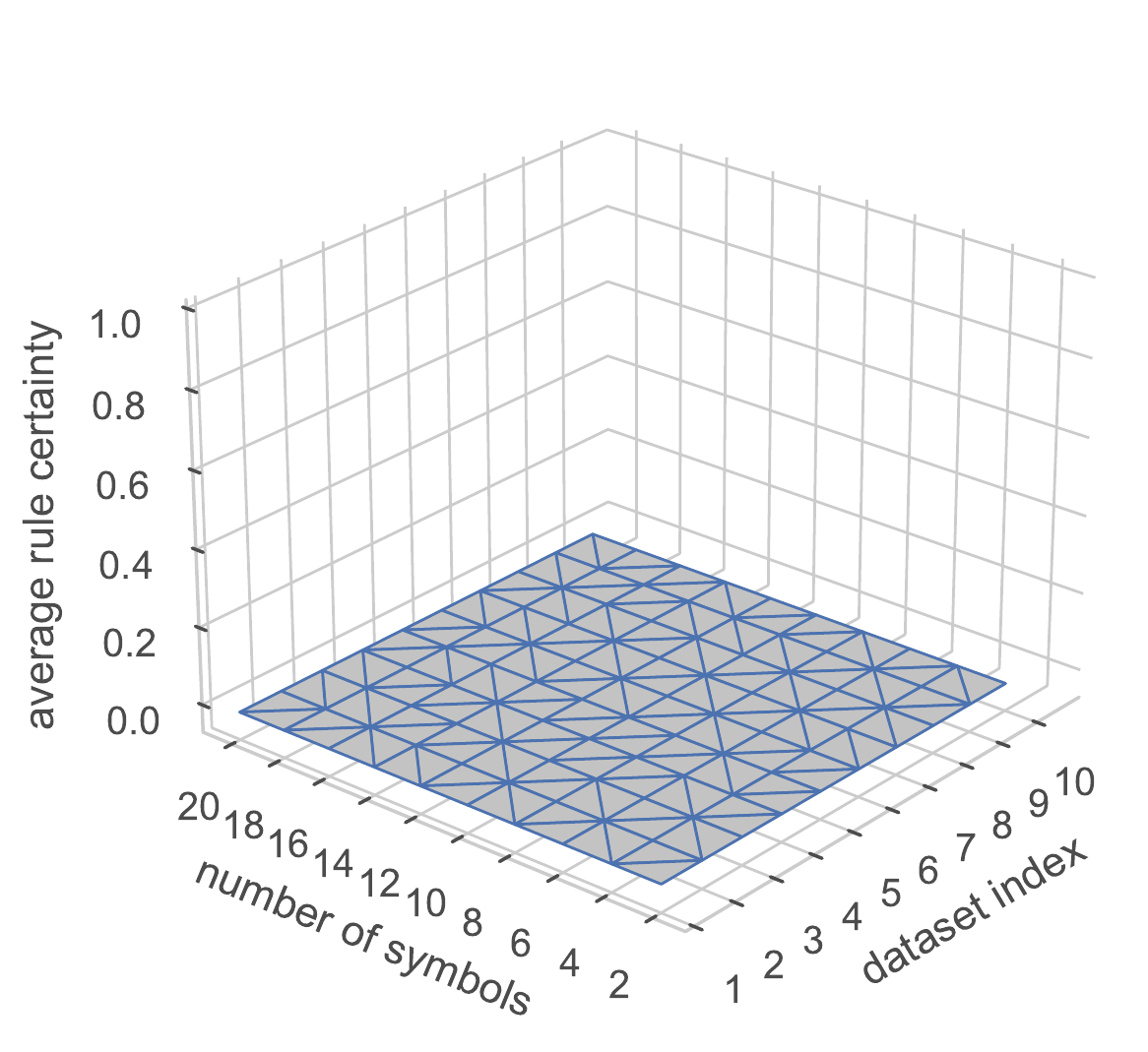}
	\caption{G\"odel and Equation \eqref{eq:distance-local1}}
	\end{subfigure}
	\begin{subfigure}{0.49\textwidth}
	\center
	\includegraphics[width=\textwidth]{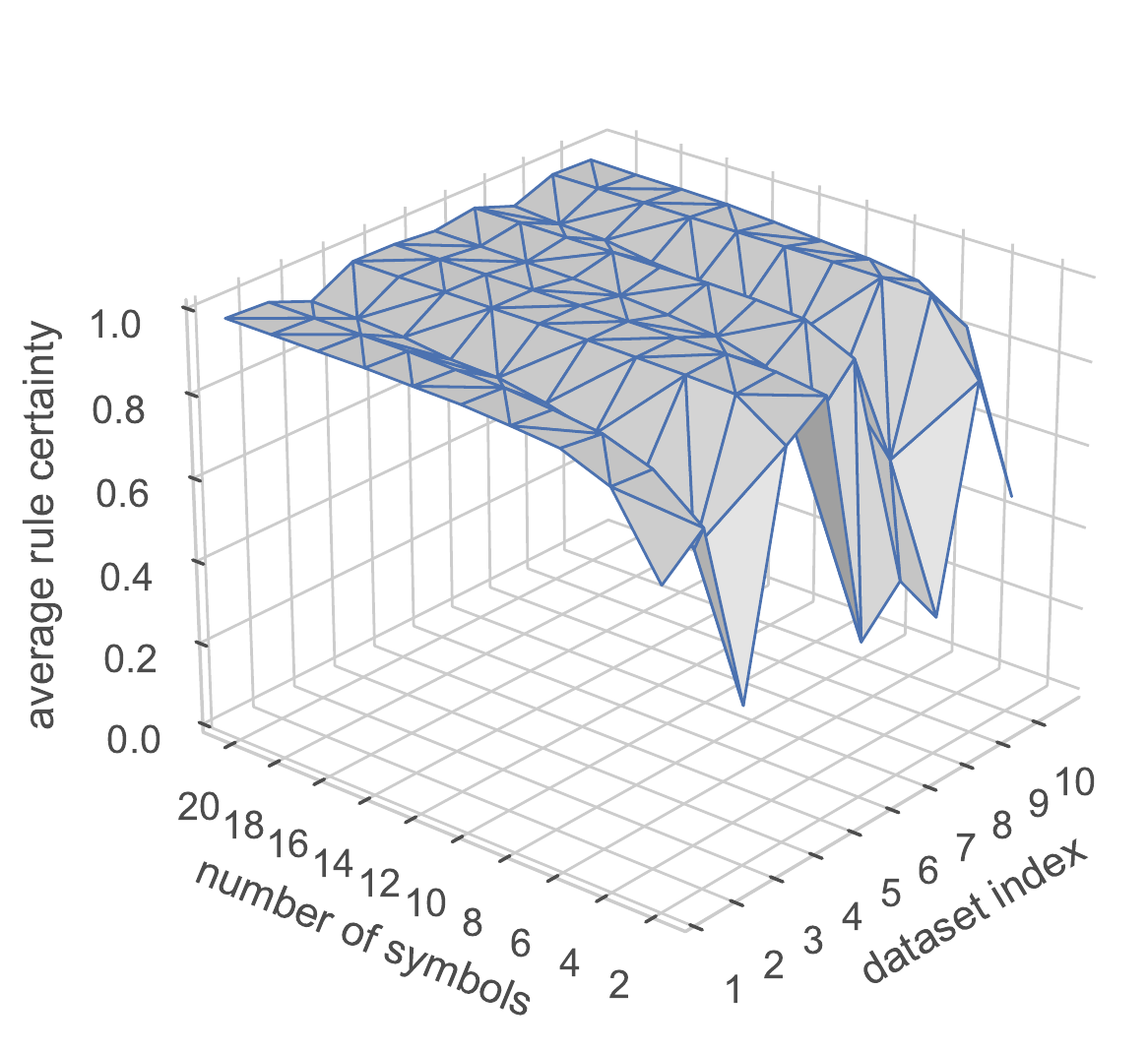}
	\caption{Łukasiewicz and Equation \eqref{eq:distance-local1}}
	\end{subfigure}
	\begin{subfigure}{0.49\textwidth}
	\center
	\includegraphics[width=\textwidth]{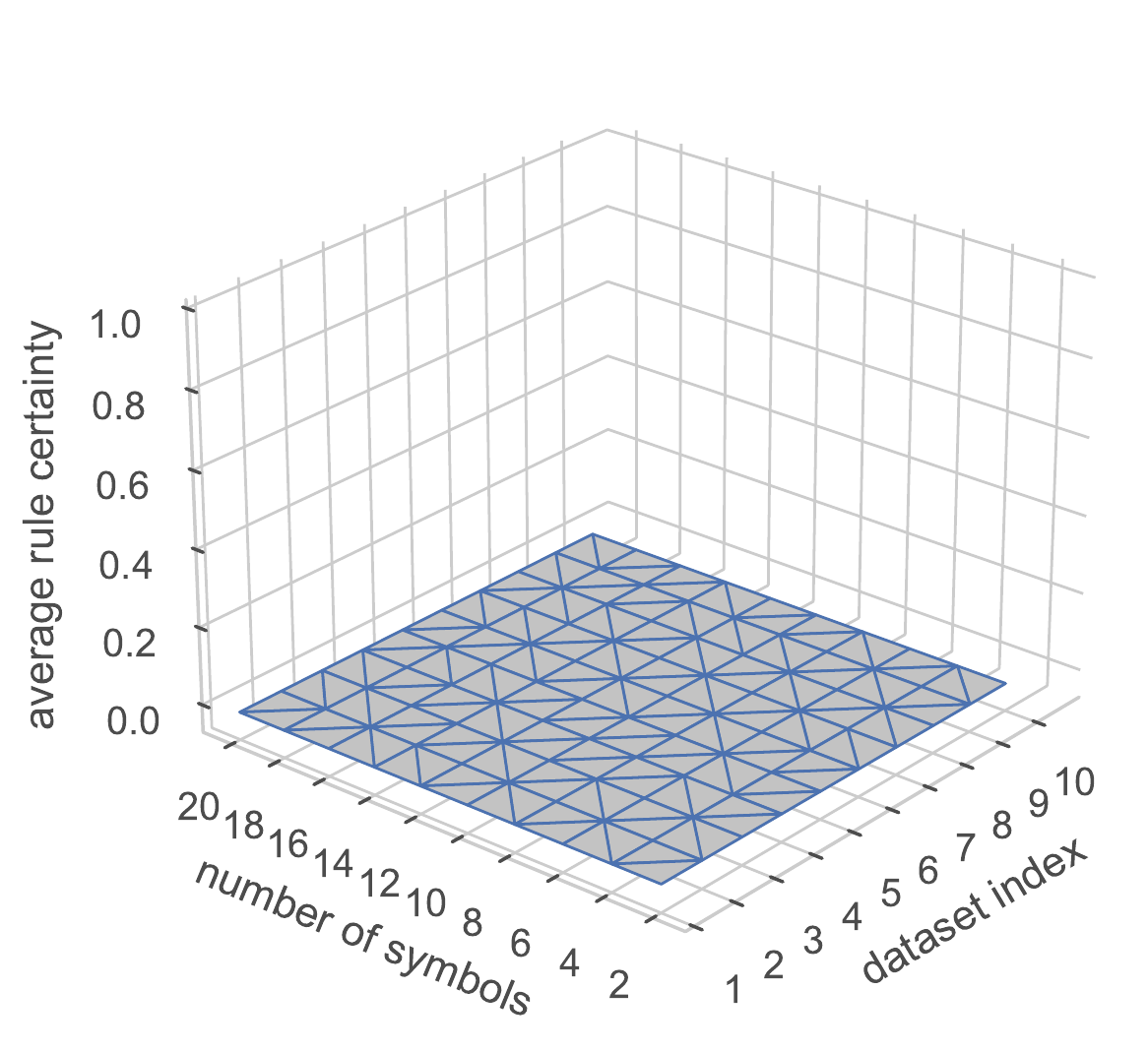}
	\caption{G\"odel and Equation \eqref{eq:distance-local2}}
	\end{subfigure}
	\begin{subfigure}{0.49\textwidth}
	\center
	\includegraphics[width=\textwidth]{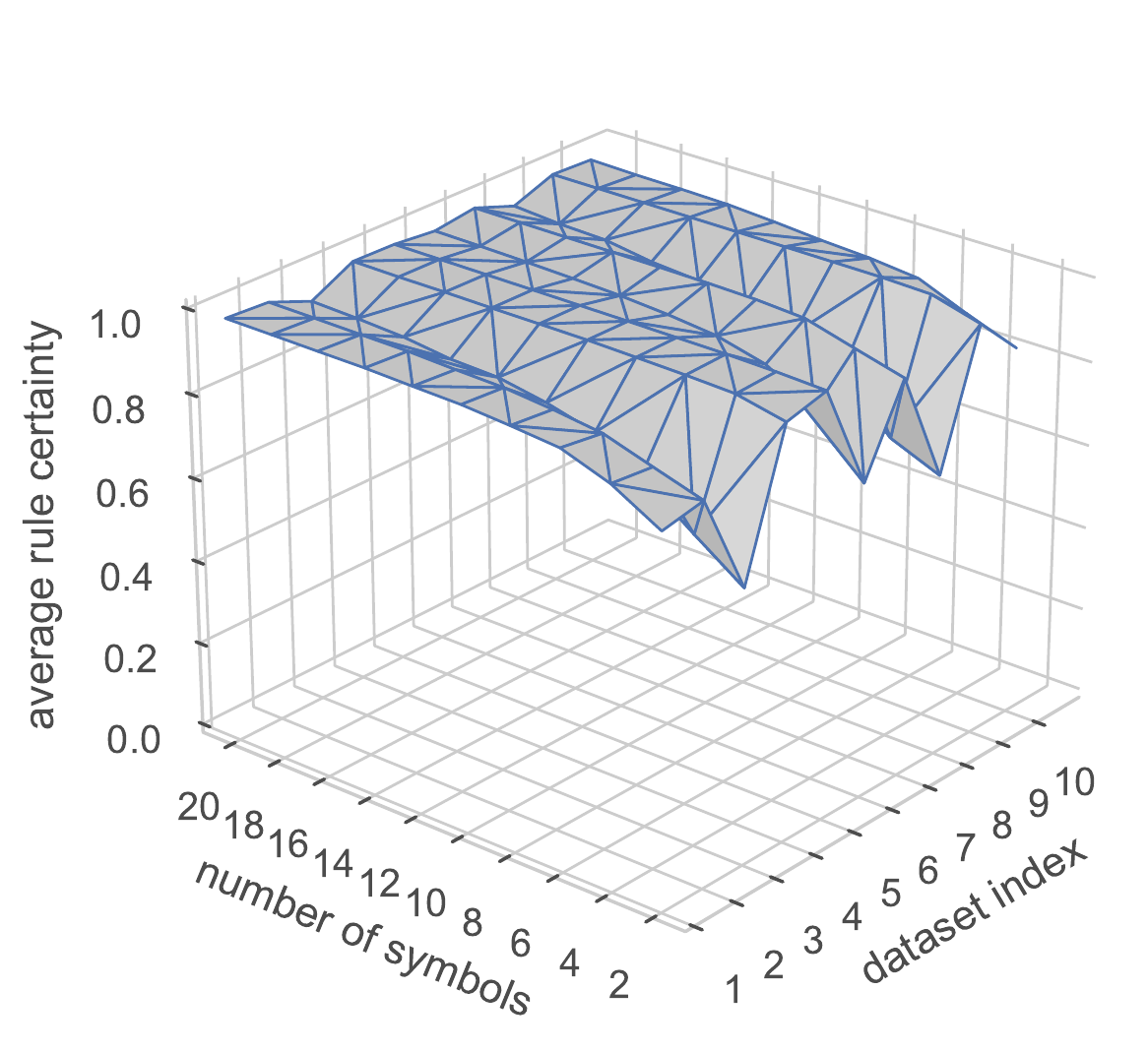}
	\caption{Łukasiewicz and Equation \eqref{eq:distance-local2}}
	\end{subfigure}	
	
	\captionsetup{justification=justified}
	\caption{Average rule confidence for each dataset when varying the number of symbols and the fuzzy implication function. Only G\"odel and Łukasiewicz are visualized. In these simulations, Na\"ive Bayes is the black box.}
\label{fig:nb-terms}
\end{figure*}

\begin{figure*}[!htbp]
\center

    \begin{subfigure}{0.49\textwidth}
	\center
	\includegraphics[width=\textwidth]{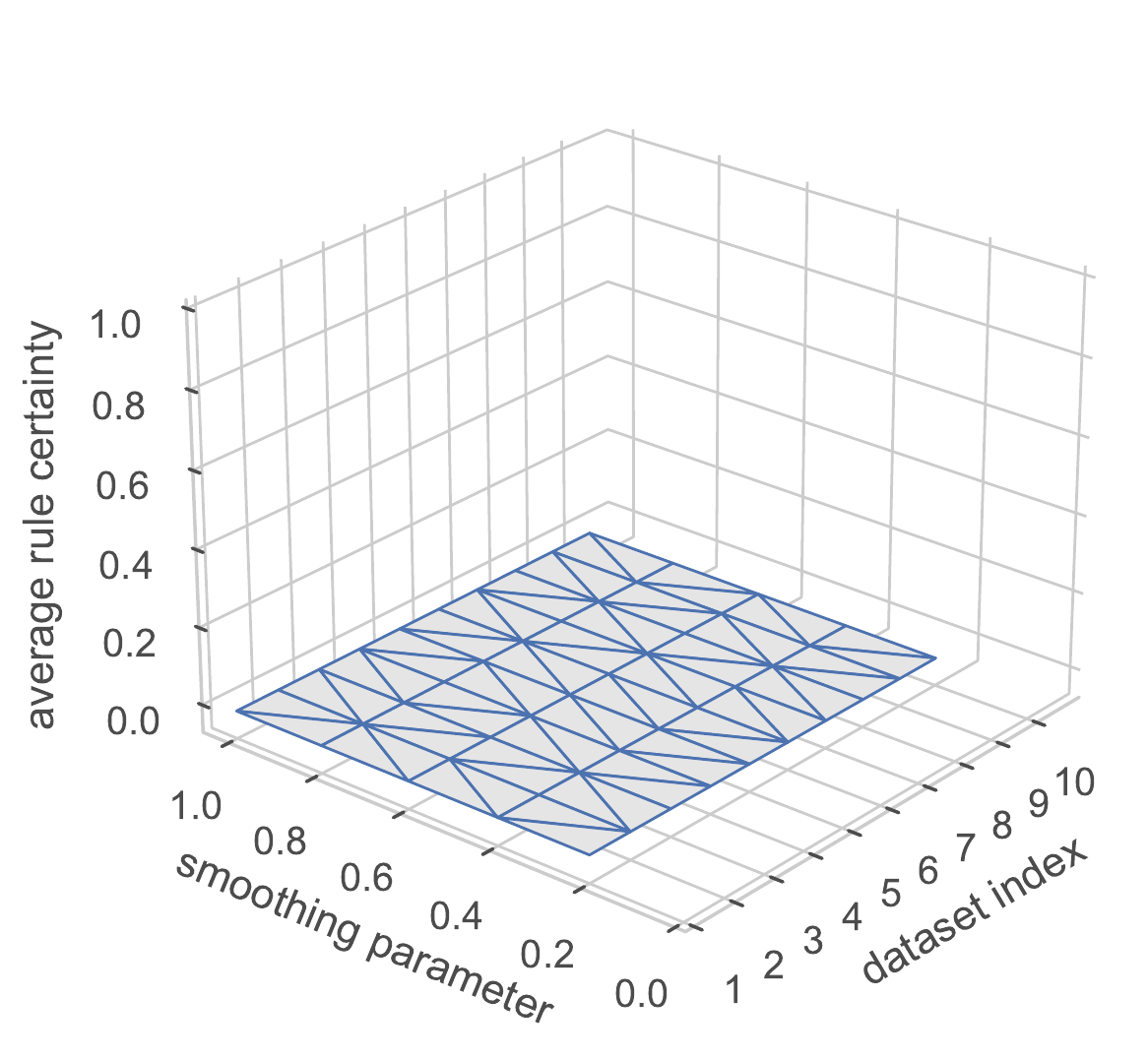}
	\caption{G\"odel and Equation \eqref{eq:distance-local1}}
	\end{subfigure}
	\begin{subfigure}{0.49\textwidth}
	\center
	\includegraphics[width=\textwidth]{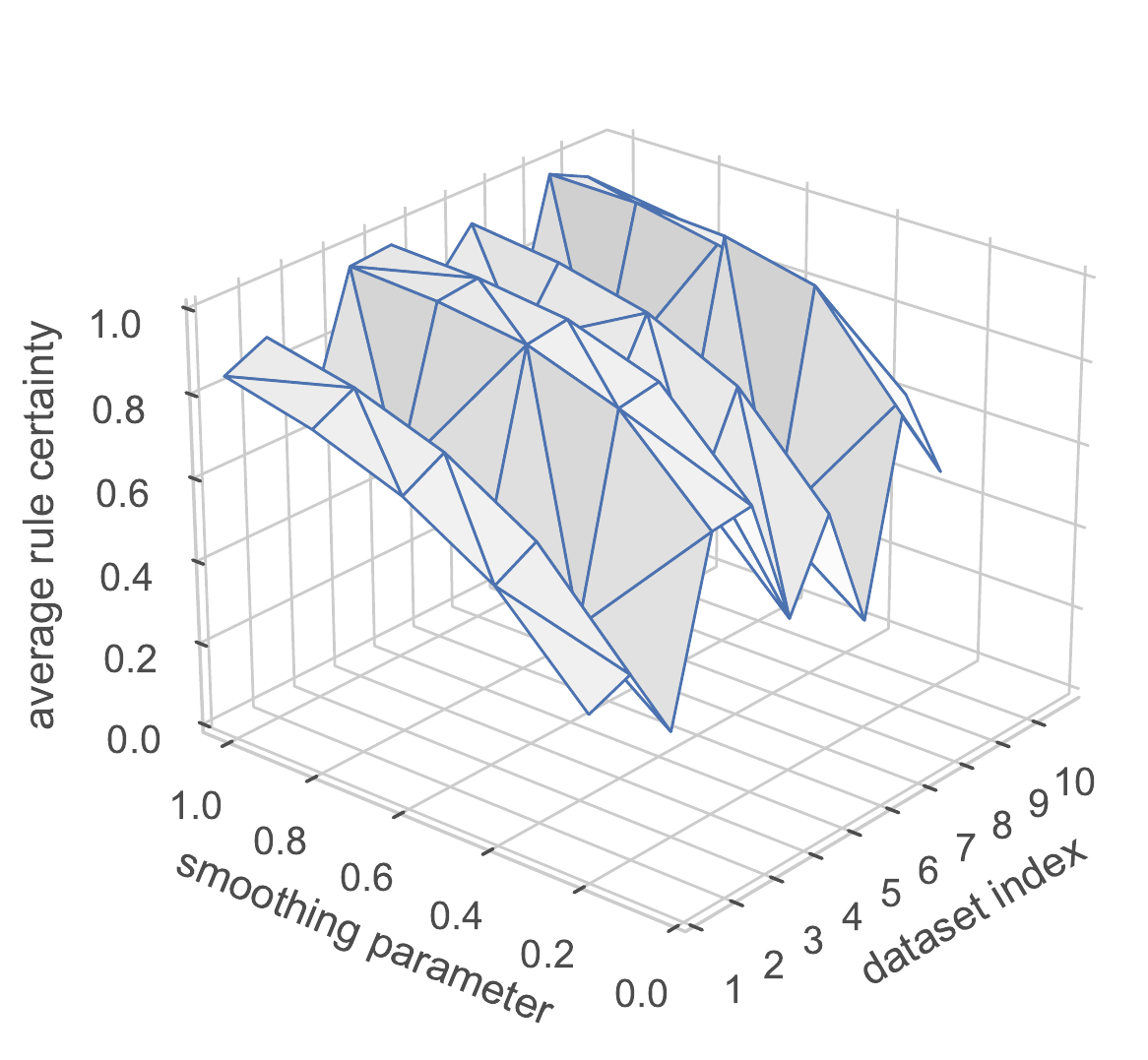}
	\caption{Łukasiewicz and Equation \eqref{eq:distance-local1}}
	\end{subfigure}
	\begin{subfigure}{0.49\textwidth}
	\center
	\includegraphics[width=\textwidth]{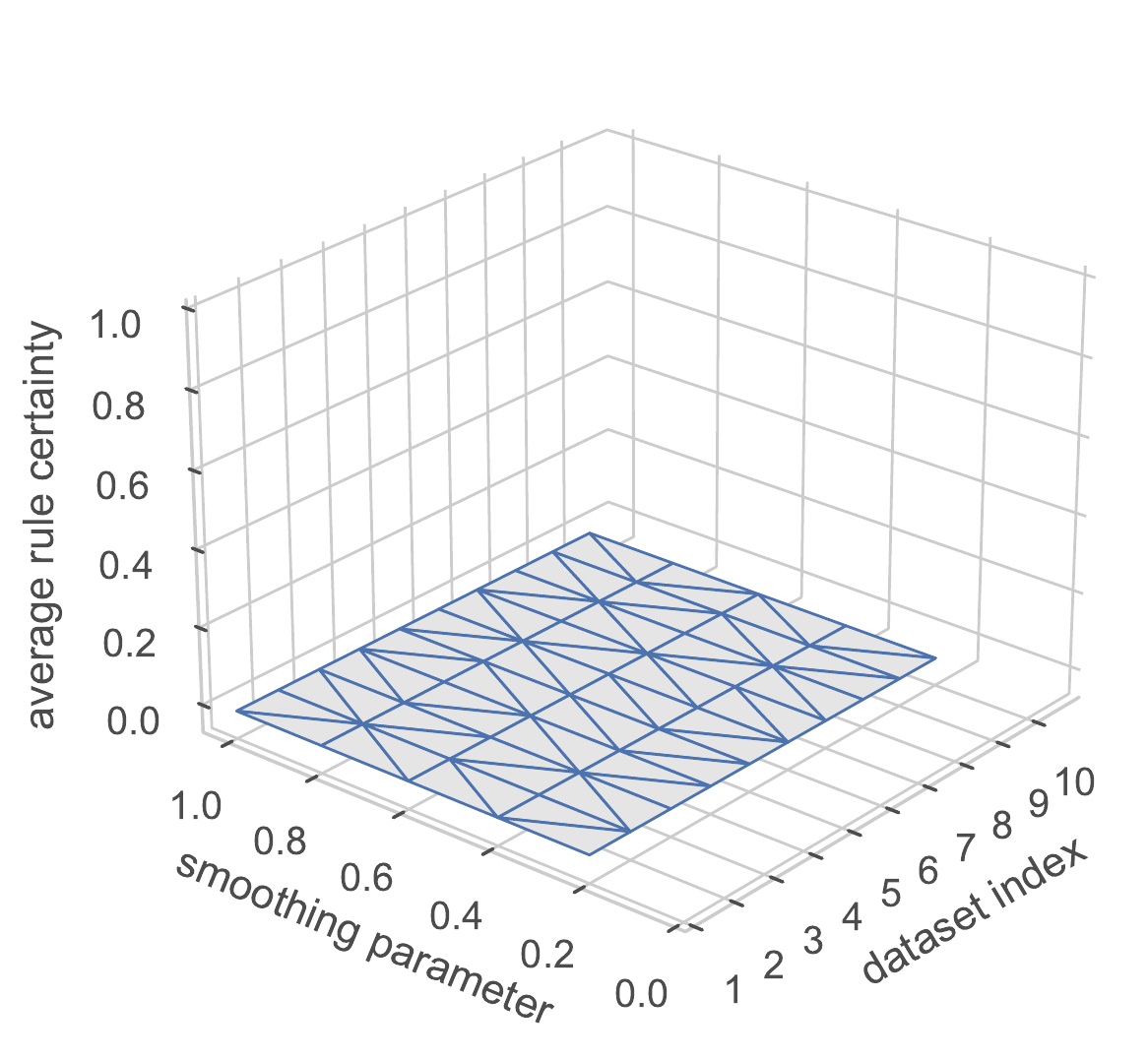}
	\caption{G\"odel and Equation \eqref{eq:distance-local2}}
	\end{subfigure}
	\begin{subfigure}{0.49\textwidth}
	\center
	\includegraphics[width=\textwidth]{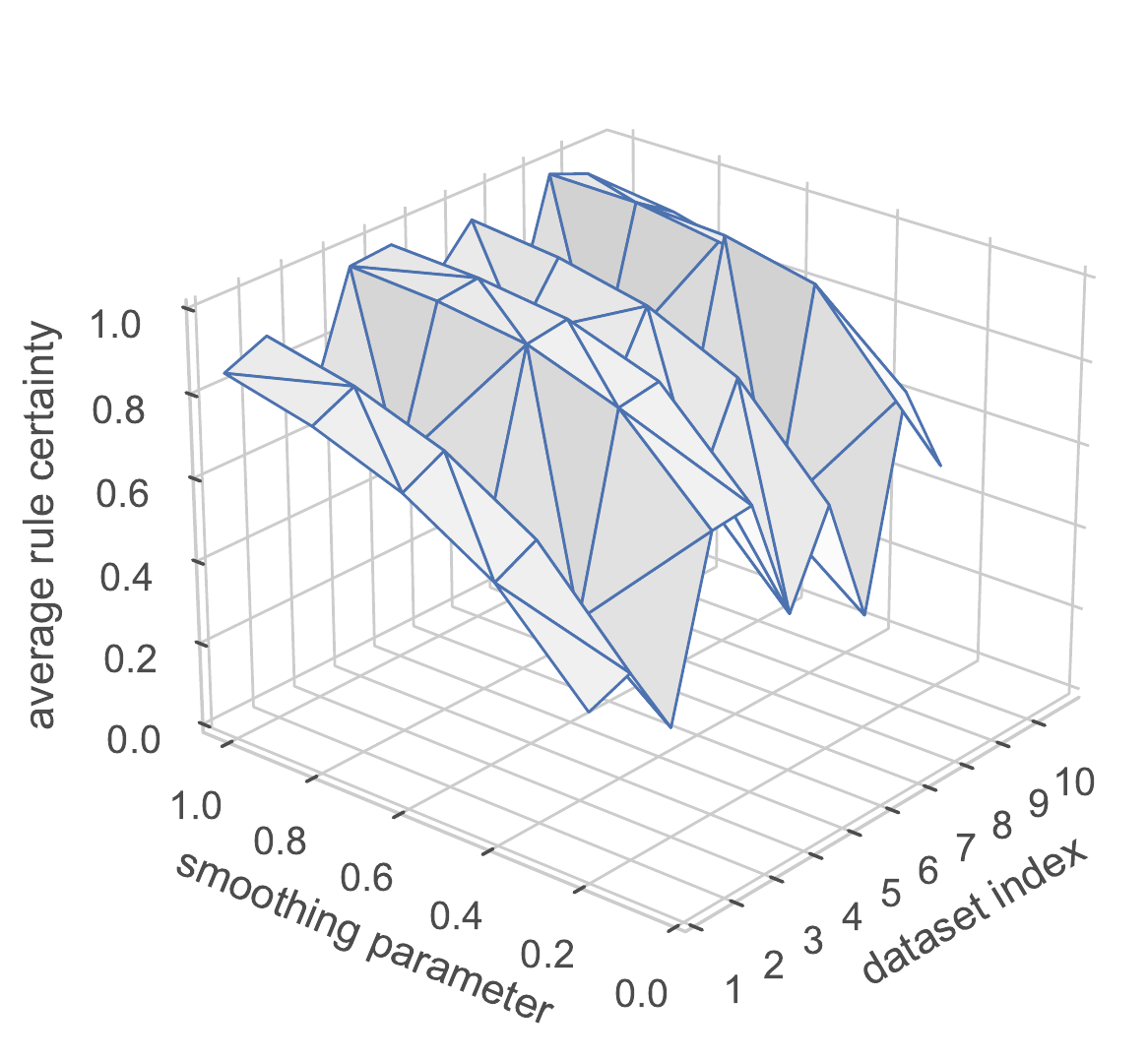}
	\caption{Łukasiewicz and Equation \eqref{eq:distance-local2}}
	\end{subfigure}	
	
	\captionsetup{justification=justified}
	\caption{Average rule confidence for each dataset when varying the smoothing parameter and the fuzzy implication function. Only G\"odel and Łukasiewicz are visualized. In these simulations, Na\"ive Bayes is the black box.}
\label{fig:nb-smoothing}
\end{figure*}

In the subsequent phase, we built a Na\"ive Bayes classifier to conduct the sensitivity analysis. The outcomes are visualized in Figures \ref{fig:nb-terms} and \ref{fig:nb-smoothing}, where we see an analogous behavior. The smoothing parameter is more influential than the number of symbols.

\paragraph{Light Gradient Boosting Machine}

\begin{figure*}[!htbp]
\center
    \begin{subfigure}{0.49\textwidth}
	\center
	\includegraphics[width=\textwidth]{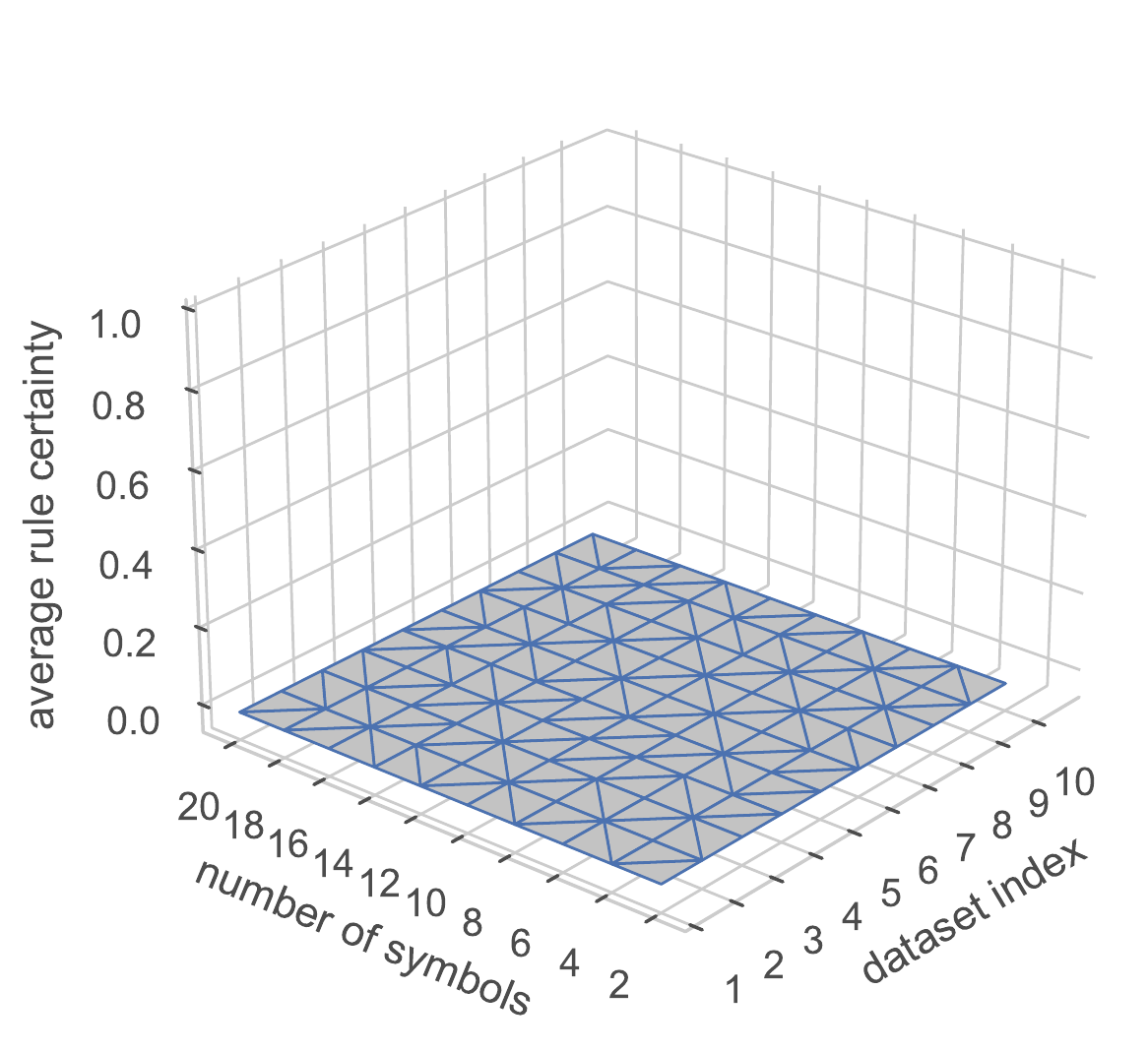}
	\caption{G\"odel and Equation \eqref{eq:distance-local1}}
	\end{subfigure}
	\begin{subfigure}{0.49\textwidth}
	\center
	\includegraphics[width=\textwidth]{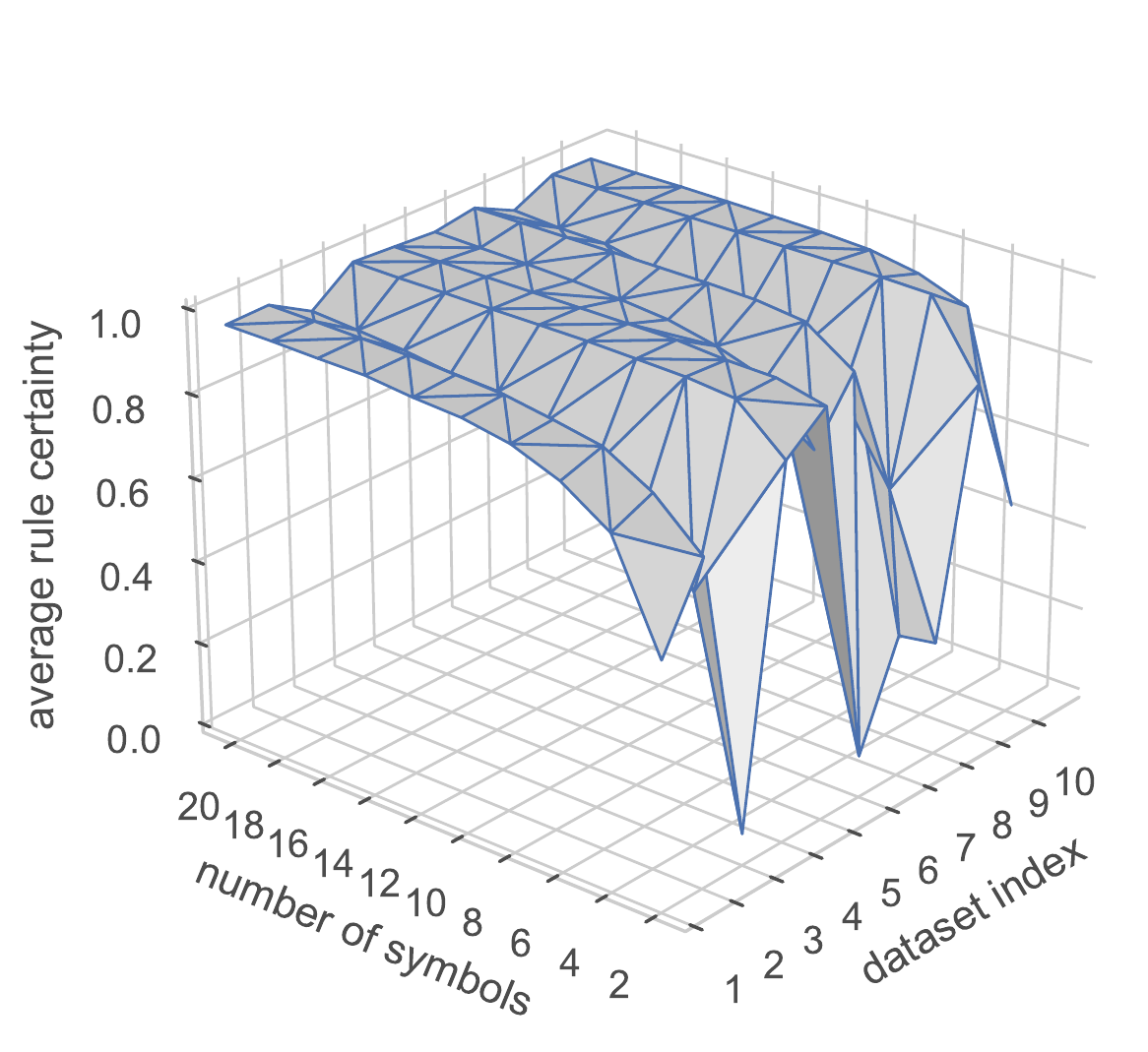}
	\caption{Łukasiewicz and Equation \eqref{eq:distance-local1}}
	\end{subfigure}
	\begin{subfigure}{0.49\textwidth}
	\center
	\includegraphics[width=\textwidth]{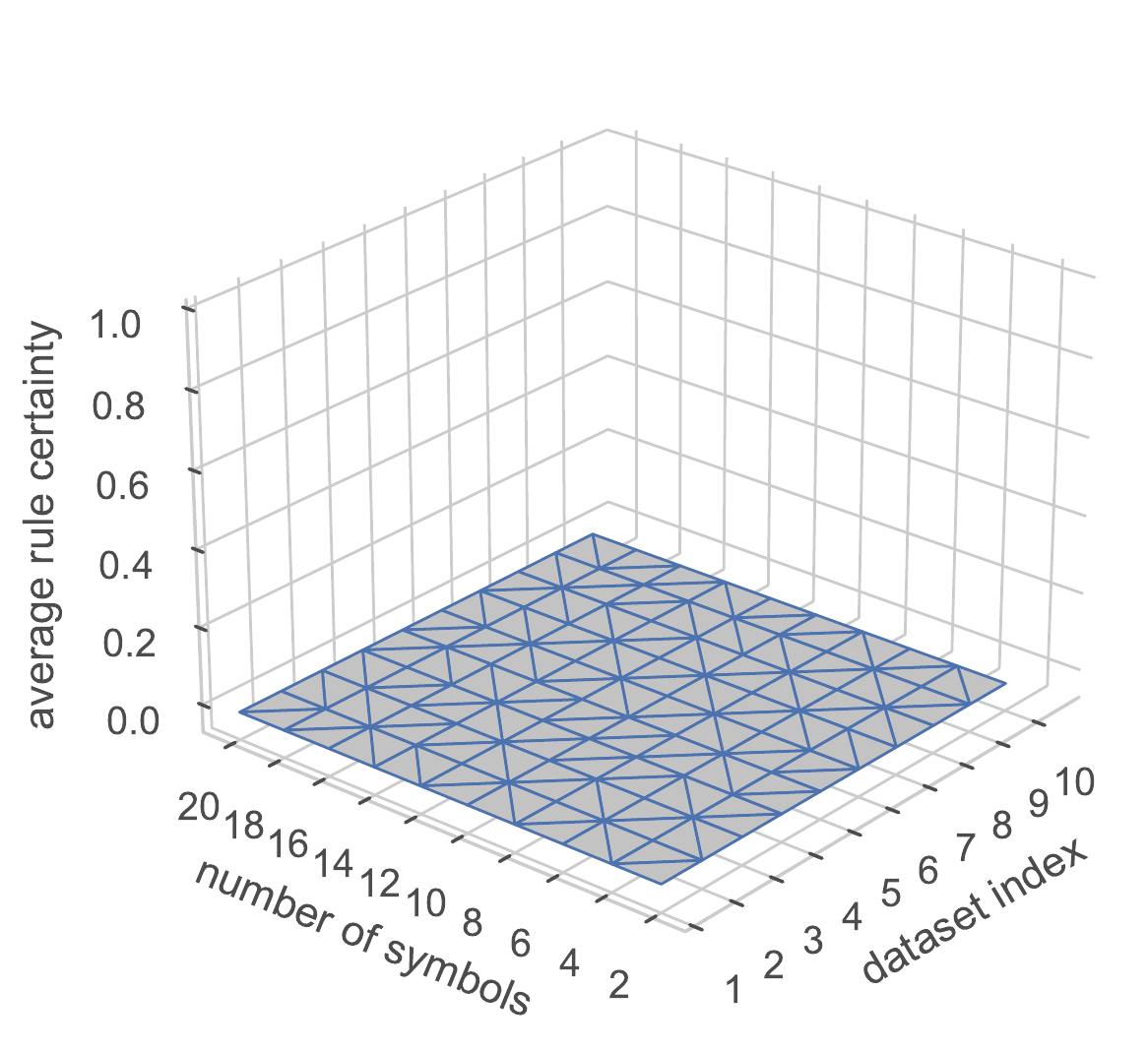}
	\caption{G\"odel and Equation \eqref{eq:distance-local2}}
	\end{subfigure}
	\begin{subfigure}{0.49\textwidth}
	\center
	\includegraphics[width=\textwidth]{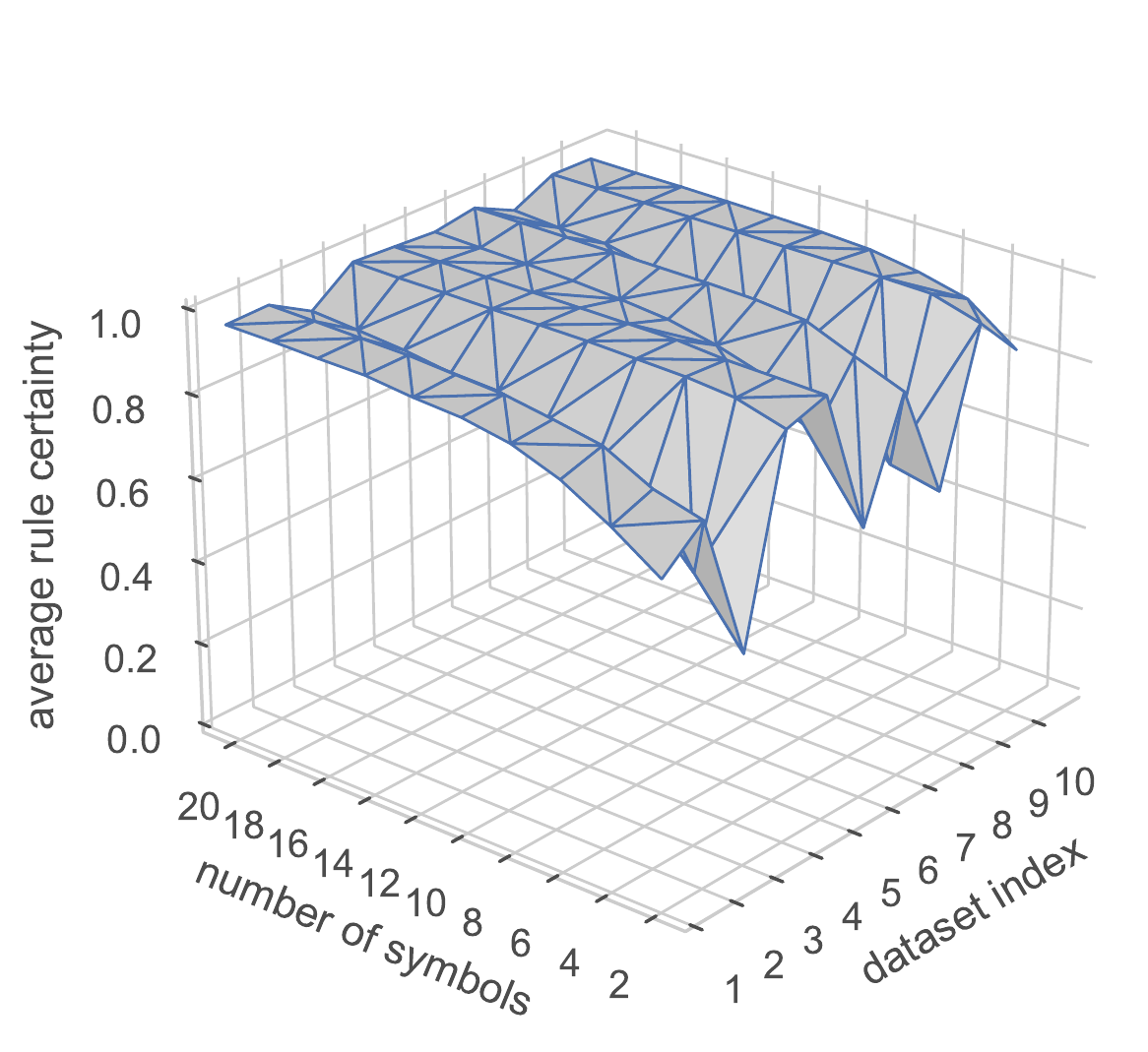}
	\caption{Łukasiewicz and Equation \eqref{eq:distance-local2}}
	\end{subfigure}	
	
	\captionsetup{justification=justified}
	\caption{Average rule confidence for each dataset when varying the number of symbols and the fuzzy implication function. Only G\"odel and Łukasiewicz are visualized. In these simulations, a Light Gradient Boosting Machine is the black box.}
\label{fig:gbm-terms}
\end{figure*}

\begin{figure*}[!htbp]
\center
    \begin{subfigure}{0.49\textwidth}
	\center
	\includegraphics[width=\textwidth]{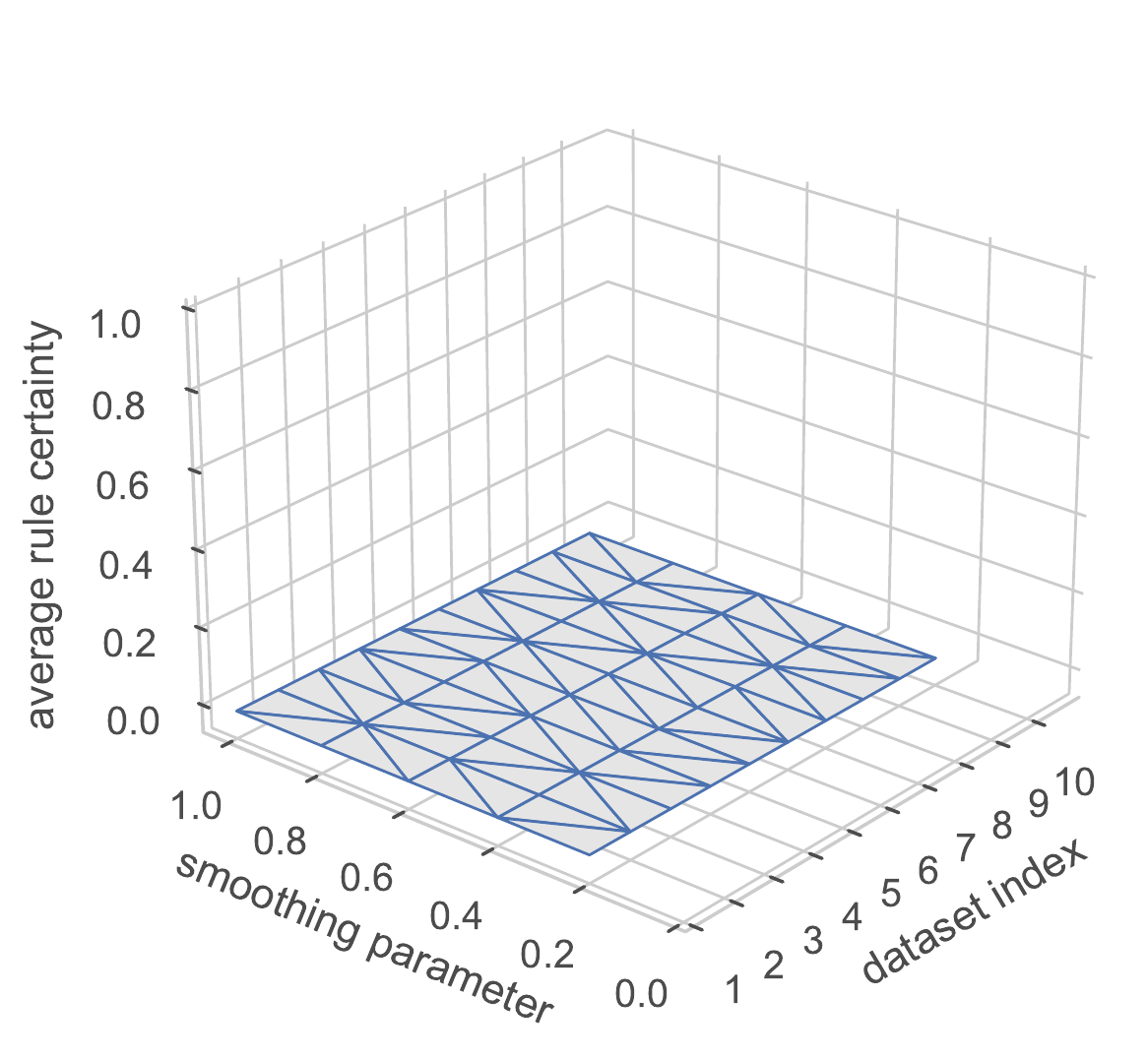}
	\caption{G\"odel and Equation \eqref{eq:distance-local1}}
	\end{subfigure}
	\begin{subfigure}{0.49\textwidth}
	\center
	\includegraphics[width=\textwidth]{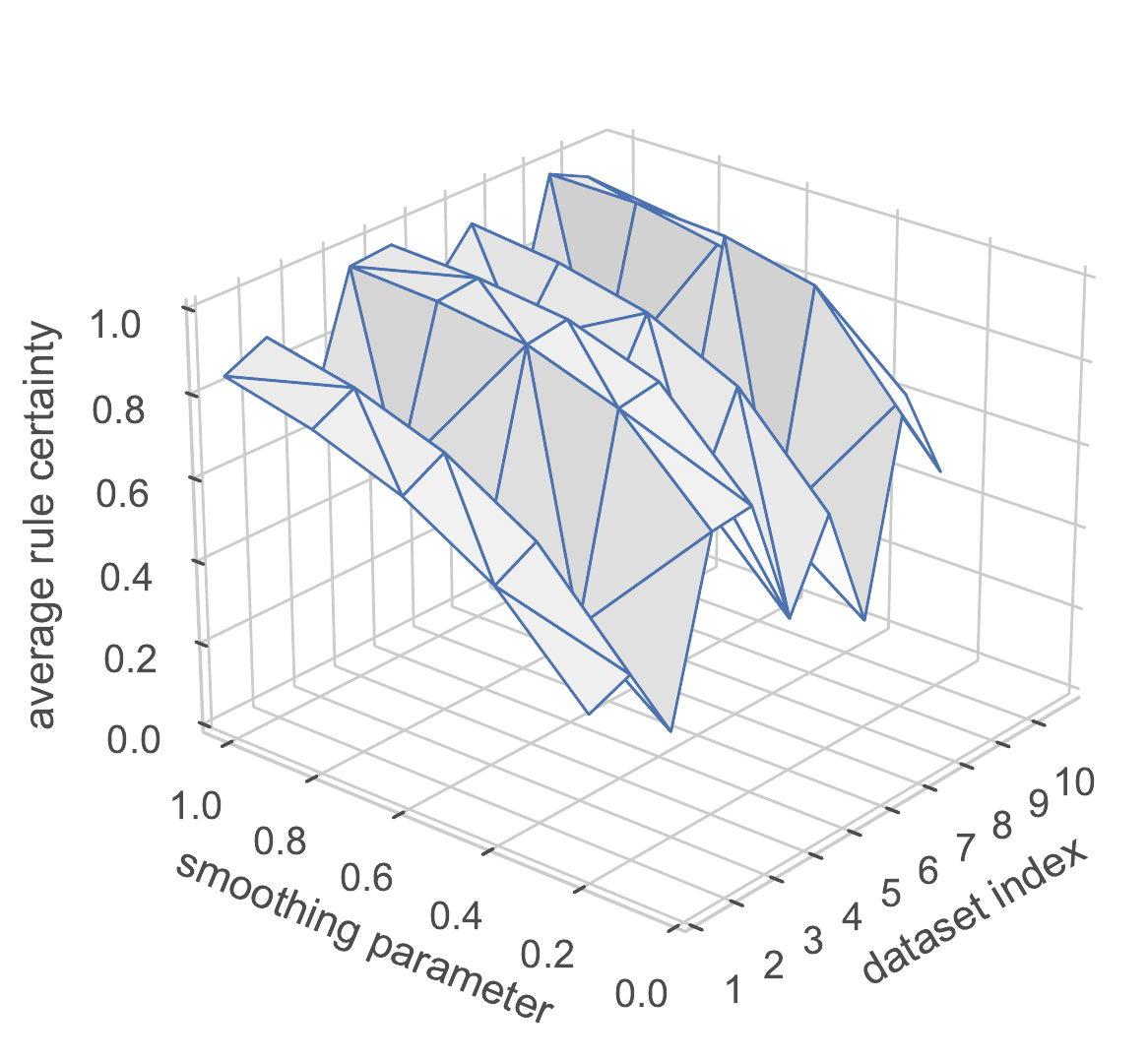}
	\caption{Łukasiewicz and Equation \eqref{eq:distance-local1}}
	\end{subfigure}
	\begin{subfigure}{0.49\textwidth}
	\center
	\includegraphics[width=\textwidth]{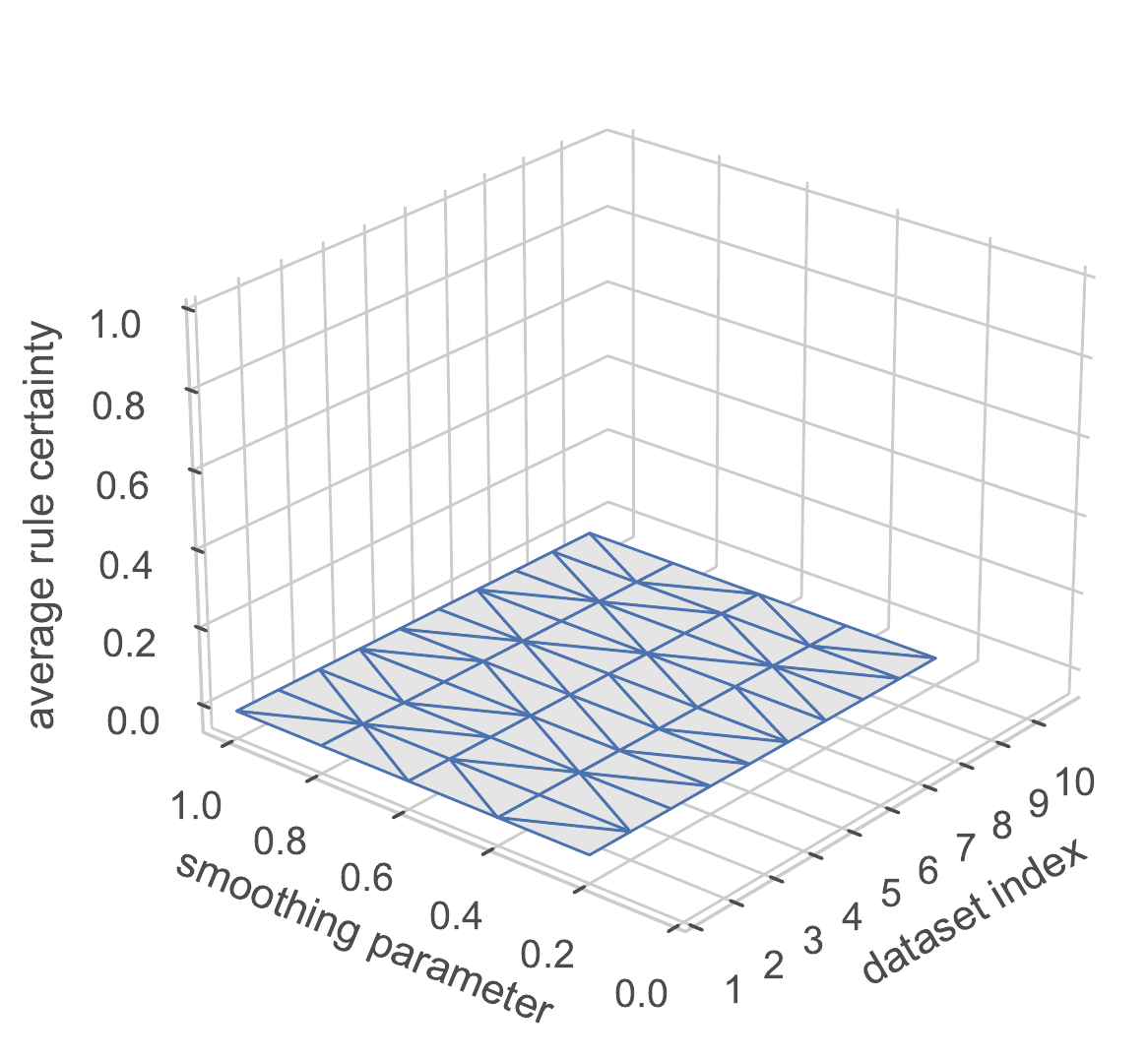}
	\caption{G\"odel and Equation \eqref{eq:distance-local2}}
	\end{subfigure}
	\begin{subfigure}{0.49\textwidth}
	\center
	\includegraphics[width=\textwidth]{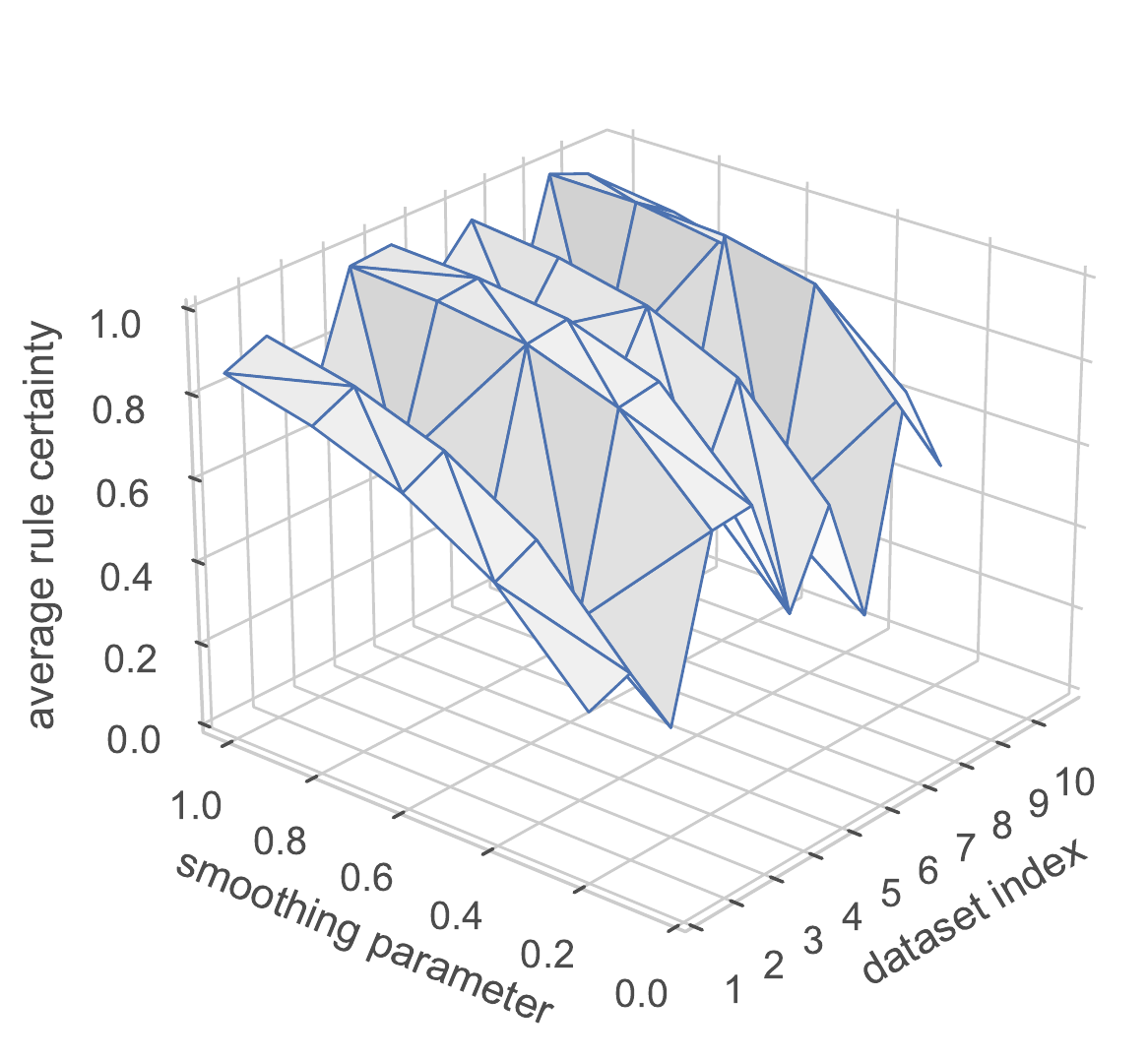}
	\caption{Łukasiewicz and Equation \eqref{eq:distance-local2}}
	\end{subfigure}	
	
	\captionsetup{justification=justified}
	\caption{Average rule confidence for each dataset when varying the smoothing parameter and the fuzzy implication function. Only G\"odel and Łukasiewicz are visualized. In these simulations, a Light Gradient Boosting Machine is the black box.}
\label{fig:gbm-smoothing}
\end{figure*}

Finally, we present the extended sensitivity analysis using a Light Gradient Boosting Machine as the black box. The outcomes of these experiments are illustrated in Figures \ref{fig:gbm-terms} and \ref{fig:gbm-smoothing}. The behavior is consistent for this model as well. The smoothing parameter impacts the average rule confidence more significantly than the number of symbols. We also observe a saturating behavior of these two parameters where the supremum values are recommended.

\paragraph{Discussion} The sensitivity analysis results using different classifiers showed that the explanation module is truly model-agnostic. The results across the different classifiers are comparable, and all lead to the same conclusion that the Łukasiewicz implicator, the proposed distance function, and large values for $\lambda$ lead to the best results. The behavior of the model, quantified with the use of the average rule confidence, proved to be highly similar no matter which classifier model was being explained.

\end{document}